\documentclass[10pt,twocolumn,letterpaper]{article}

\usepackage{ijcb}
\usepackage{times}
\usepackage{epsfig}
\usepackage{graphicx}
\usepackage{amsmath}
\usepackage{amssymb}
\usepackage[breaklinks=true,letterpaper=true,colorlinks,bookmarks=false]{hyperref}
\usepackage{booktabs}
\usepackage{multirow}
\usepackage{multicol}
\usepackage{caption}
\usepackage{rotating}
\usepackage{subcaption}
\usepackage{caption}
\usepackage{bm}
\usepackage{float}
\usepackage{placeins}
\usepackage[table]{xcolor}
\newcommand{\secref}[1]{\S\ref{#1}}
\usepackage[accsupp]{axessibility}

\ijcbfinalcopy % *** Uncomment this line for the final submission

\ifijcbfinal\fi

\def\showAppendix{1}

\begin{document}

%%%%%%%%% TITLE
\title{BiOcularGAN: Bimodal Synthesis and Annotation of Ocular Images\vspace{-3mm}}%using Dual Branch StyleGAN}

\author{Darian Tomašević$^{1,*}$, \  Peter Peer$^{1,\dagger}$, \ Vitomir Štruc$^{2,\ddagger}$\\
$^{1}$\it{Faculty of Computer and Information Science, University of Ljubljana, Slovenia}\\
% Večna pot 113, 1000 Ljubljana, Slovenia\\
$^{2}$\it{Faculty of Electrical Engineering, University of Ljubljana, Slovenia} \\
{\tt\small $^{*}$darian.tomasevic@fri.uni-lj.si, $^{\dagger}$peter.peer@fri.uni-lj.si, $^{\ddagger}$vitomir.struc@fe.uni-lj.si}
\vspace{-4mm}
}
% \author{Darian Tomašević\\
% Faculty of Computer and Information Science, University of Ljubljana\\
% Večna pot 113, 1000 Ljubljana, Slovenia\\
% {\tt\small darian.tomasevic@gmail.com} % 
% \and 
% Peter Peer \\ 
% Faculty of Computer and Information Science, University of Ljubljana\\
% Večna pot 113, 1000 Ljubljana, Slovenia\\
% {\tt\small peter.peer@fri.uni-lj.si}
% \and
% Vitomir Štruc\\
% Faculty of Electrical Engineering, University of Ljubljana \\
% Tržaška cesta 25, 1000 Ljubljana, Slovenia\\
% {\tt\small vitomir.struc@fe.uni-lj.si}
% }
% % \and
% % Second Author\\
% % First line of institution2 address\\
% % {\tt\small secondauthor@i2.org}
% % }

\maketitle
\thispagestyle{empty}

%%%%%%%%% ABSTRACT
\begin{abstract}
% Current state-of-the-art segmentation techniques for ocular images are critically dependent on large-scale annotated datasets, which are labor-intensive to gather and often raise privacy concerns. In this paper, we present a novel framework, called BiOcularGAN, capable of generating synthetic large-scale datasets of photorealistic (visible light and near-infrared) ocular images, together with corresponding segmentation labels to address these issues. At its core, the framework relies on a novel Dual-Branch StyleGAN2 (DB-StyleGAN2) model that facilitates bimodal image generation, and a Semantic Mask Generator (SMG) that produces semantic annotations by exploiting DB-StyleGAN2's feature space. We evaluate BiOcularGAN through extensive experiments across five diverse ocular datasets and analyze the effects of bimodal data generation on image quality and the produced annotations. Our experimental results show that BiOcularGAN is able to produce high-quality matching bimodal images and annotations (with minimal manual intervention) that can be used to train highly competitive (deep) segmentation models (in a privacy aware-manner) that perform well across multiple real-world datasets. The source code for the BiOcularGAN framework has been made publicly available at {\url{ https://github.com/dariant/BiOcularGAN}}.
Current state-of-the-art segmentation techniques for ocular images are critically dependent on large-scale annotated datasets, which are labor-intensive to gather and often raise privacy concerns. In this paper, we present a novel framework, called BiOcularGAN, capable of generating synthetic large-scale datasets of photorealistic (visible light and near-infrared) ocular images, together with corresponding segmentation labels to address these issues. At its core, the framework relies on a novel Dual-Branch StyleGAN2 (DB-StyleGAN2) model that facilitates bimodal image generation, and a Semantic Mask Generator (SMG) component that produces semantic annotations by exploiting latent features of the DB-StyleGAN2 model. We evaluate BiOcularGAN through extensive experiments across five diverse ocular datasets and analyze the effects of bimodal data generation on image quality and the produced annotations. Our experimental results show that BiOcularGAN is able to produce high-quality matching bimodal images and annotations (with minimal manual intervention) that can be used to train highly competitive (deep) segmentation models (in a privacy aware-manner) that perform well across multiple real-world datasets. The source code for the BiOcularGAN framework is publicly available at {\url{ https://github.com/dariant/BiOcularGAN}}.
%\href{https://github.com/dariant/BiOcularGAN}{GitHub}. 
\vspace{-0mm}

% Inspired by the recent advancements in GAN models, we present in t

% This paper presents a novel framework for generating large-scale datasets of ocular images
% Ocular techniques ... require data ... Current ocular generation techniques do not use StyleGAN, which is the new state-of-the-art ... In this paper we propose a new dual-branch version ... and build a new framework BiOcularGAN ... 
%  Lorem ipsum   Lorem ipsumLorem ipsumLorem ipsumLorem ipsum Lorem ipsum  Lorem ipsum Lorem ipsum Lorem ipsum Lorem ipsum Lorem ipsum Lorem ipsum Lorem ipsum Lorem ipsum Lorem ipsum Lorem ipsum Lorem ipsum Lorem ipsum Lorem ipsum Lorem ipsum Lorem ipsum Lorem ipsum Lorem ipsum Lorem ipsum Lorem ipsum Lorem ipsum Lorem ipsum Lorem ipsum Lorem ipsum Lorem ipsum Lorem ipsum Lorem ipsum Lorem ipsum Lorem ipsum Lorem ipsum Lorem ipsum Lorem ipsum Lorem ipsum Lorem ipsum Lorem ipsum Lorem ipsum Lorem ipsum Lorem ipsum Lorem ipsum Lorem ipsum Lorem ipsum Lorem ipsum Lorem ipsum Lorem ipsum Lorem 
\end{abstract}

%%%%%%%%% BODY TEXT
\section{Introduction}\label{Sec:Intro}

% Recent advancements in the field of image-based biometry are based on the use of deep learning methods. To use these methods effectively we require large amounts of training data {\color{red}} \cite{nguyen2017iris}, however, the task of gathering such data is often very time-consuming and labor-intensive. Furthermore, access to gathered datasets is frequently restricted, in order to protect personal data of subjects. These issues are especially applicable for the ocular modality, where the majority of current datasets \cite{vitek2020comprehensive} are rather small and also lack  annotations.  
Modern biometric systems are predominantly based on convolutional neural networks (CNNs) and transformer models, which rely on massive annotated (training) datasets to achieve competitive performance \cite{nguyen2017iris}. While large-scale datasets can today easily be collected from the web for many biometric modalities, such collection procedures often raise \textit{privacy} and \textit{copyright-related concerns} \cite{jasserand2018massive,meden2021privacy}. Additionally, the annotation of such large-scale datasets is today (in most cases) still a manual, labor-intensive, and time-consuming task. 
These points are especially true for datasets dedicated to the segmentation of ocular images (in various imaging domains), where, next to the data collection, the generation of high-quality (multi-class) semantic annotations is known to be a costly endeavor \cite{vitek2020comprehensive, zanlorensi2022ocular}.

% Modern computer vision solutions are predominantly based on convolutional neural networks (CNNs) and transformer models, which rely on massive annotated (training) datasets to achieve competitive performance \cite{nguyen2017iris}. While large-scale datasets can today easily be collected from the web, they often raise privacy and copyright-related concerns \cite{jasserand2018massive,meden2021privacy}. Additionally, the annotation of such datasets is still (in most cases) a manual, labor-intensive and time-consuming task.  especially true This especially applicable for the ocular modality, where the majority of current datasets \cite{vitek2020comprehensive} are not only rather small but also lack human annotations.

Researchers are, therefore, increasingly looking into automatic techniques that allow for the generation of synthetic datasets that require no (or minimal) human intervention during the annotation process \cite{galeev2021learning, li2021semantic, pakhomov2021segmentation, zhang2021datasetgan}. However, several challenges are associated with such an approach: $(i)$~the synthetic (training) samples need to be as close as possible to the expected real-world data to allow for the trained model to perform well during deployment, $(ii)$ the synthesis procedure must allow for the generation of large and diverse datasets that can cater to the data needs of modern deep learning models, and $(iii)$ data annotations need to be produced automatically, without (or with minimal) supervision. To meet these challenges, existing solutions often resort to Generative Adversarial Networks (GANs) \cite{goodfellow2014generative, karras2018progressive} %, of which the most notable is the recently developed StyleGAN model \cite{karras2018progressive}, 
due to their ability to generate highly photorealistic and detailed synthetic data and the fact that the model's internal representations can be exploited to generate semantic segmentation labels alongside the generated images \cite{zhang2021datasetgan}.

%can also be used as a basis for generation of annotated datasets \cite{zhang2021datasetgan}.

%The recent development of Generative Adversarial Networks (GANs) \cite{goodfellow2014generative}... These networks rely on the use of two neural networks, the generator and the discriminator, to generate artificial images.

% Recent advancements in the field of image-based biometry are based on the use of deep learning methods. To use these methods effectively we require large amounts of training data {\color{red}} \cite{nguyen2017iris}, however, the task of gathering such data is often very time-consuming and labor-intensive. Furthermore, access to gathered datasets is frequently restricted, in order to protect personal data of subjects.   

%In the past, deep learning approaches relied on simple data augmentation techniques, in order to artificially generate more training data \cite{rot2018deep, loo2020open}. However, these techniques can generate only a limited amount of additional realistic examples \cite{motamed_2021_gandata}.

\begin{figure}[t!]
    \centering
    %\begin{subfigure}{.99\columnwidth}
    %\includegraphics[width=0.99\linewidth,trim = 0 0 0 0, clip]{figures/biocular_gan/teaser_new.pdf}
    %\caption{Illustration of BiOcularGAN}
    %\label{fig:BiOcluar_sub}
    %\end{subfigure}\vspace{2mm}
    %\begin{subfigure}{.99\columnwidth}
    \includegraphics[width=0.99\linewidth]{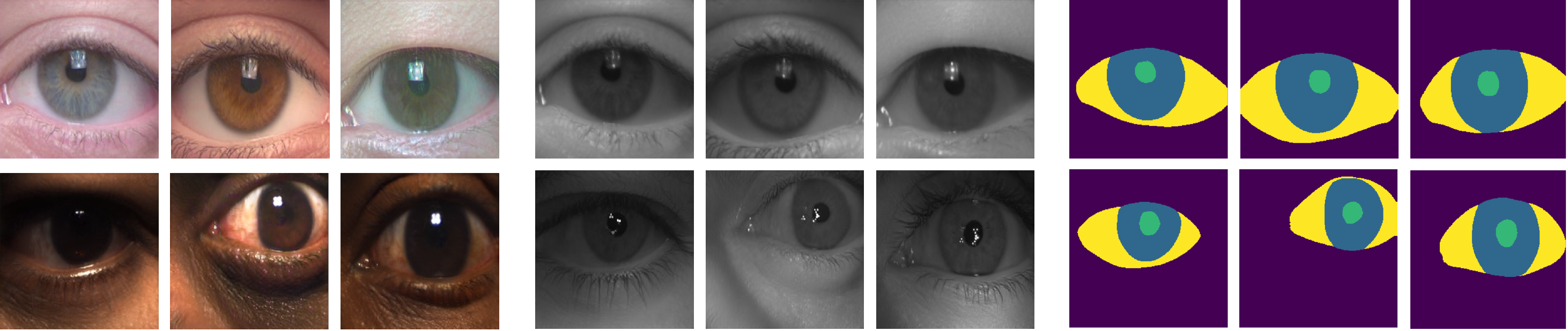}
    %\caption{Generated sample results: VIS, NIR, segmenation masks}
    %\label{fig:BiOcluar_sub_examples}
    %\end{subfigure}
    % \vspace{-2mm}
    \caption{\textbf{Example data generated with BiOcularGAN.} The proposed framework is based on a novel Dual-Branch StyleGAN2 model and can generate (synthetic) per-pixel aligned visible light (VIS) and near-infrared (NIR) ocular images  as well as corresponding segmentation masks. \vspace{-4mm}} 
    \label{fig:teaser}
\end{figure}

Motivated by the needs for large-scale synthetic datasets and the capabilities of recent generative models, we present in this paper a novel data generation framework, called BiOcularGAN, capable of generating aligned photorealistic (bimodal) ocular images in the visible (VIS) and near-infrared (NIR) spectra along with corresponding segmentation masks, as illustrated in Figure~\ref{fig:teaser}. The key components of the framework are $(i)$ a novel \textit{dual-branch} StyleGAN2 (DB-StyleGAN2) model, which extends the capabilities of previous StyleGAN versions to bimodal data synthesis, and $(ii)$ a data annotation procedure, inspired by \cite{zhang2021datasetgan}, that exploits the semantic information encoded by the bimodal synthesis network for segmentation mask generation. %The framework is    
%novel StyleGAN-inspired model, the dual-branch StyleGAN2 (DB-StyleGAN2), which is capable of generating aligned photorealistic (bimodal) ocular images in the visible (VIS) and near-infrared (NIR) spectra. It relies on two images branches along the generator architecture, one for each spectrum, thus allowing shared input information regarding the to-be generated images. In addition, each branch uses its own discriminator, to ensure that images of both spectra remain realistic. Around the proposed model we also build a new framework, BiOcularGAN, capable of generating synthetic datasets of ocular VIS and NIR images, along with corresponding segmentation masks. 
We evaluate the proposed approach in experiments with five diverse datasets and investigate the impact of the bimodal (VIS and NIR) generation process on the quality of the synthesized images. %and compare generated images of both DB-StyleGAN2 and the unimodal StyleGAN2. 
Furthermore, we analyze the ability of BiOcularGAN to generate useful datasets by observing how well current semantic segmentation models, trained on synthetic labeled data, generalize to diverse real-world datasets. In summary, we make the following main contributions: 

%To evaluate the proposed approach we analyse the quality of images generated by DB-StyleGAN2 and compare it with those obtained with the unimodal StyleGAN2. 

%Furthermore we perform in-depth analysis of the 
% {\color{red}} We do what with it? Test it? Showcase that is can be used to do what? .. investigate the effects of the shared NIR / VIS information on the image quality and on the generated masks .... train for semantic segmentation task ... and see how well it generalizes to other datasets ... ... ..

% In this paper, we presented BiOcularGAN, a framework for generating synthetic datasets of  ocular images with corresponding segmentation masks. At the heart of the framework is a novel generative model, i.e., the dual-branch StyleGAN2 (DB-StyleGAN2), capable of generating photorealistic aligned bimodal (VIS and NIR) ocular images. Using the proposed BiOcularGAN framework, we showed that it is possible to generate large and representative synthetic dataset that can be used to train competitive segmentation models that generalize well across a diverse set of ocular images. 

% , aptly named the  dual-branch StyleGAN2 (DB-StyleGAN2)., for generating photo

% In this paper ... we synthesise two images and use two separate Discriminators, in order obtain a bi-modal generator, which simultaneously generates matching RGB and NIR images .. 

% in combination with an updated DatasetGAN model to generate masks ...

%-----------------------------------------
% Contributions
%-----------------------------------------

\begin{itemize}
\vspace{-1mm}
    \item We present BiOcularGAN, a powerful \textit{framework for generating large labeled datasets} of ocular images based on bimodal data representations that can be used to train contemporary segmentation models.\vspace{-1mm}
    \item We design a \textit{novel bimodal generative model}, i.e., the Dual-Branch StyleGAN2 (DB-StyleGAN2), capable of synthesizing visually convincing (aligned) ocular images in both the visible and near-infrared domains.\vspace{-1mm}
    \item We show that using bimodal information as the basis for generating ground truth segmentation masks leads to  improvements in the quality of the generated annotations compared to solutions using only a single modality, e.g., the state-of-the-art DatasetGAN~\cite{zhang2021datasetgan}.
   % \item We make all source code, models, weights and supporting scripts publicly available %to the research community 
%    with the goal of fostering reproducibility via: URL\footnote{URL removed for review.}. 
\end{itemize}
\vspace{-2mm}

%%---------------------------------------------------
% Related work
%----------------------------------------------------
\section{Related work}\label{Sec:RelatedWork}

% Tale sekcija je med pol in 3/4 strani, nič več - pisati strnjeno.
%The proposed BiOcularGAN framework builds on recent advances in image generation.
%In this section, we review work closely related to the proposed BiOcularGAN framework. %We position our contribution with respect to existing research on image and datasets generation as wel as ocular image synthesis.   

\textbf{Image and Dataset Generation.}
% Dejmo povedat nasledenje v dveh delih:
%
% Ogromno napredka v zadnjem času - predvsem zaradi GAN
% Razni modeli za sintezo slik - lastni
% Potem pride StyleGAN in variante - osredotočamo se na hihg-level opise kaj se je konceptualno spreminjalo
%
Image synthesis techniques have experienced rapid development in the past decade, most notably due to the introduction of Generative Adversarial Networks (GANs) \cite{goodfellow2014generative}. Over time, a myriad of improvements and iterations to the GAN model have been proposed, from manipulating latent space distributions \cite{brock2018large} to using multiple discriminator networks \cite{durugkar2015generative}. Despite numerous advancements \cite{karras2018progressive, miyato2018spectral}, some of the inner workings of the generator networks remained poorly understood \cite{bau2019visualizing}. 
\begin{figure*}[t!]
    \centering
    \includegraphics[width=0.95\linewidth,trim = 0 3mm 0 0, clip]{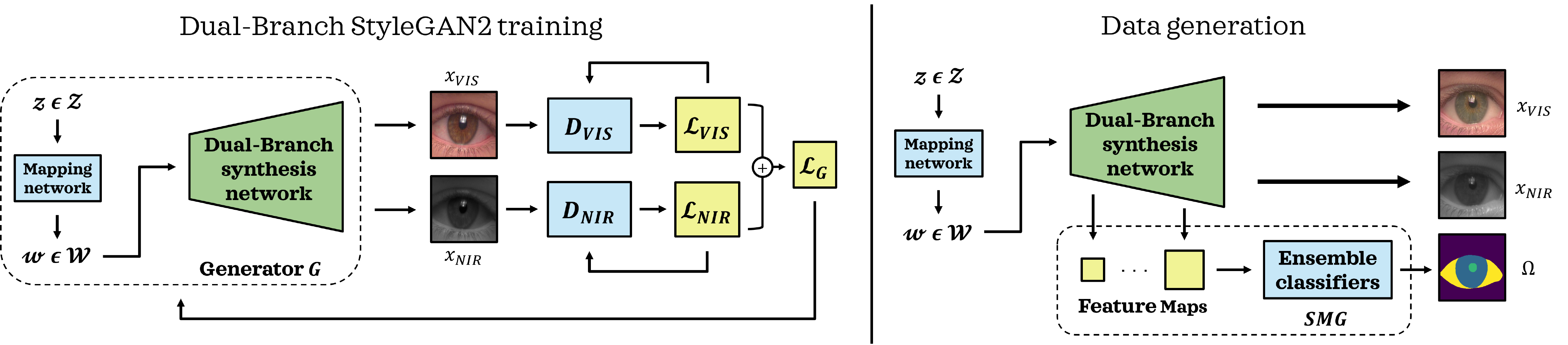}
    \caption{\textbf{High-level overview of the BiOcularGAN framework.}  The proposed Dual-Branch StyleGAN2 simultaneously produces pairs of VIS and NIR images. The model is trained using two separate (VIS and NIR) discriminators, $D_{VIS}$ and $D_{NIR}$ with corresponding losses, $\mathcal{L}_{VIS}, \mathcal{L}_{NIR}$. The combined loss $\mathcal{L}_{G}$ is used to train the generator $G$. The final data generation process  first produces a pair of VIS and NIR images with the DB-synthesis network and then passes the internal feature maps to the Semantic Mask Generator (SMG), which generates the corresponding ground truth segmentation masks. \vspace{-2mm}%, i.e., the segmentation mask.
    }
    \label{fig:BiOcularGAN_pipeline}
\end{figure*}

More recently, a powerful new generation model, called StyleGAN, was proposed by Karras \textit{et al.} in \cite{stylegan_1_karras2019style}. With its high-resolution image synthesis capabilities, the model drastically outperformed other unconditional image generation techniques across a variety of datasets. Since then, the authors further iterated on the model (with StyleGAN2 and StyleGAN3) \cite{stylegan2_karras2020analyzing, stylegan3_karras2021alias} and addressed several of its characteristic artifacts with changes to model architecture and training procedures. Most notably,  Karras \textit{et al.} \cite{stylegan2_ADA_karras2020training} also introduced various image augmentations to the discriminator, thus immensely lowering the amount of training data required to train the StyleGAN2 model.

Several approaches have also been proposed to enable the synthesis of segmentation masks alongside images generated by StyleGAN, either by using separate generator branches  \cite{li2021semantic} or by exploiting the feature space of the generator \cite{pakhomov2021segmentation, zhang2021datasetgan}. The latter approach showcased the ability to generate high-quality datasets of paired images and segmentation masks, with only a few annotated examples, and was aptly named DatasetGAN \cite{zhang2021datasetgan}. In this paper, we build on the outlined advances and present, to the best of our knowledge, \textit{the first StyleGAN2-based model for bimodal data synthesis}. As we demonstrate in the experimental section, the model leads to visually convincing generation results and allows us to synthesize large datasets of matched ocular images in the VIS and NIR imaging domains with corresponding ground truth segmentation masks.

\textbf{Ocular Synthesis.} 
% Pol pride AMPAK sekcija v stilu: Despite the considerable progress in generative models, a limited number of solutions capable of generating photorealistic high-quality coular images have so far presented in the literature. To the best of ou knowledge, ... naštejmo nekaj rešitev za iris in podobne zadeve in podajmo omejitve (niso cele slike, ena modalnost, itd.) Potem pride stavek: Different from these works, we ... poudari naše prednosti in izvirne zadeve -pazi kaj lahko zagovarjamo - v glavnem bimodal. Mogoče bomo morali dat v ločeno sekcijo.
%
% Nedavni: Iris and periocular biometrics for head mounted displays: Segmentation, recognition, and synthetic data generation
%
% Facebook challenge - glej zgornji članek in trace-aj.
%
% Initial GAN-based models included 
% Iris generation ... pretenta security \cite{kohli2017synthetic}
% RaSGAN \cite{yadav2019synthesizing}
Despite the considerable progress in generative models, only a limited number of solutions capable of generating photorealistic high-quality ocular images have so far been presented in literature. 
Shrivastava \textit{et al.} \cite{shrivastava2017learning} presented one of the initial GAN-based models for ocular synthesis, capable of converting pre-rendered ocular images \cite{wood2015rendering} into more realistic ones. Lee \textit{et al.} \cite{lee2018simulated} built on this approach with the use of CycleGAN \cite{zhu2017cyclegan}. However, the resulting images remained rather noisy and often did not match the original gaze direction. 
Concurrently, Kohli \textit{et al.} \cite{kohli2017synthetic} explored convolutional GAN models for iris generation. Despite significant artifacts, they successfully performed presentation attacks on the recognition systems of the time. 
%Later approaches build on higher resolution images \cite{yadav2019synthesizing}, however, they are remain limited to only iris generation.
Based on the need for large datasets, Facebook organized the OpenEDS Synthetic Eye Generation challenge \cite{facebook_openEDS_challenge}.  
Buhler \textit{et al.} \cite{buhler2019content} emerged victorious with their Seg2Eyes model, a mix of StyleGAN \cite{stylegan_1_karras2019style} and GauGAN \cite{park2019semantic}, capable of generating identity-preserving ocular images based on the desired style and input segmentation masks.  
Kaur \textit{et al.} \cite{kaur2020eyegan} introduced the EyeGAN model for the same task, and later upgraded it with a cyclic training mechanism \cite{kaur2021subject} to ensure consistency of gaze direction and style. Boutros \textit{et al.} \cite{boutros2020headmounted} proposed an alternative solution to the problem with a novel D-ID-network solution. 
Nevertheless, the generated images still featured visible artifacts.

Despite significant improvements in ocular synthesis, all current approaches generate images of only a single modality. In addition, they are also mostly focused on identity-preserving image generation and feature mechanisms that can limit the diversity of generated synthetic data. 
Different from these works, we focus in this paper on the generation of diverse and appearance-rich datasets of bimodal VIS and NIR data, along with matching synthetically-generated reference annotations. The bimodal aspect is especially useful from a segmentation aspect, since NIR images often contain important cues that are not present in VIS images, and vice versa. Furthermore, we base our work on insights from state-of-the-art image generation techniques, i.e. StyleGAN2~\cite{stylegan2_ADA_karras2020training, stylegan2_karras2020analyzing}, allowing us to learn highly successful models using a limited amount of training data. 

\section{Methodology}\label{Sec:BiOcularGAN}

The main contribution of this work is the BiOcularGAN framework that allows for photo-realistic generation of bimodal ocular images and the corresponding reference segmentation masks. In this section, we describe BiOcularGAN in detail and elaborate on its main characteristics. %together with a and highlight the differences between it and current state-of-the-art approaches \cite{stylegan2_ADA_karras2020training, zhang2021datasetgan}. 

%%%%%%%%%%%%
\subsection{Overview of the BiOcularGAN framework}
%Quick overview of 

The proposed BiOcularGAN framework, depicted in Figure~\ref{fig:BiOcularGAN_pipeline}, consists of two key components. These being $(i)$ the Dual-Branch StyleGAN2 (DB-StyleGAN2) generative model~(\secref{sec_DBStyle2}), which generates pixel-aligned VIS and NIR ocular images (\secref{sec_DBStyle2_train}), and $(ii)$ the Semantic Mask Generator (SMG)  that produces corresponding semantic segmentation masks (\secref{sec_SMG}). Jointly, these components allow for the  generation of matching photo-realistic bimodal ocular images along with corresponding high-quality annotations and, consequently, for the creation of synthetic large-scale datasets that can be used for training data-hungry deep learning (segmentation) models in a privacy-aware manner, e.g., for semantic segmentation tasks.

Formally, the BiOcularGAN generator $G$ begins with an input latent code $\mathbf{z}\in\mathcal{Z}$ that is first transformed into an intermediate latent representation $\mathbf{w}\in\mathcal{W}$ and then fed to the DB-StyleGAN2 synthesis network $g$, which produces the pixel-aligned VIS and NIR ocular images, $\mathbf{x}_{vis}\in\mathbb{R}^{W\times H\times 3}$ and $\mathbf{x}_{nir}\in\mathbb{R}^{W\times H}$, respectively, i.e.:
\begin{equation}
    \{\mathbf{x}_{vis},\mathbf{x}_{nir}\} =  G(\mathbf{z}) = g(f(\mathbf{z})),
    \label{eq:generetor}
\end{equation}
where the latent-space transformation $\mathbf{w} = f(\mathbf{z})$ is implemented with a mapping network $f$, as shown in Figure \ref{fig:BiOcularGAN_pipeline}. To generate the semantic segmentation masks, the feature maps computed along the different layers of DB-StyleGAN2 are pooled and then fed to the semantic mask generator $S$, similarly to \cite{zhang2021datasetgan}, i.e.:
%\begin{equation}
    $\Omega =  S(\phi_1(\mathbf{z}), \phi_2(\mathbf{z}),\ldots,\phi_k(\mathbf{z}))$,
%\end{equation}
where $\Omega\in\mathbb{R}^{W\times H}$ is the generated segmentation mask, $\phi$ is a mapping implemented within the generator $G$, and $k$ is the number of feature maps used. Thus, given a latent code $\mathbf{z}$, drawn from a normal distribution, BiOcularGAN generates a triplet of the following form: $\{\mathbf{x}_{vis},\mathbf{x}_{nir},\Omega\}$.

%Following the success of the data annotation procedure in \cite{}  
%This
%The entire BiOcularGAN framework  is depicted in Figure \ref{fig:BiOcularGAN_pipeline}.

%%%%%%%%%%%%
\subsection{Dual-Branch StyleGAN2}\label{sec_DBStyle2}

\begin{figure}[t!] 
\begin{center}
\begin{tabular}{c|c}
%\toprule
    %\midrule
    \small DB-synthesis network & \small DB-StyleGAN2 design \\ 
     \includegraphics[width=0.21\textwidth]{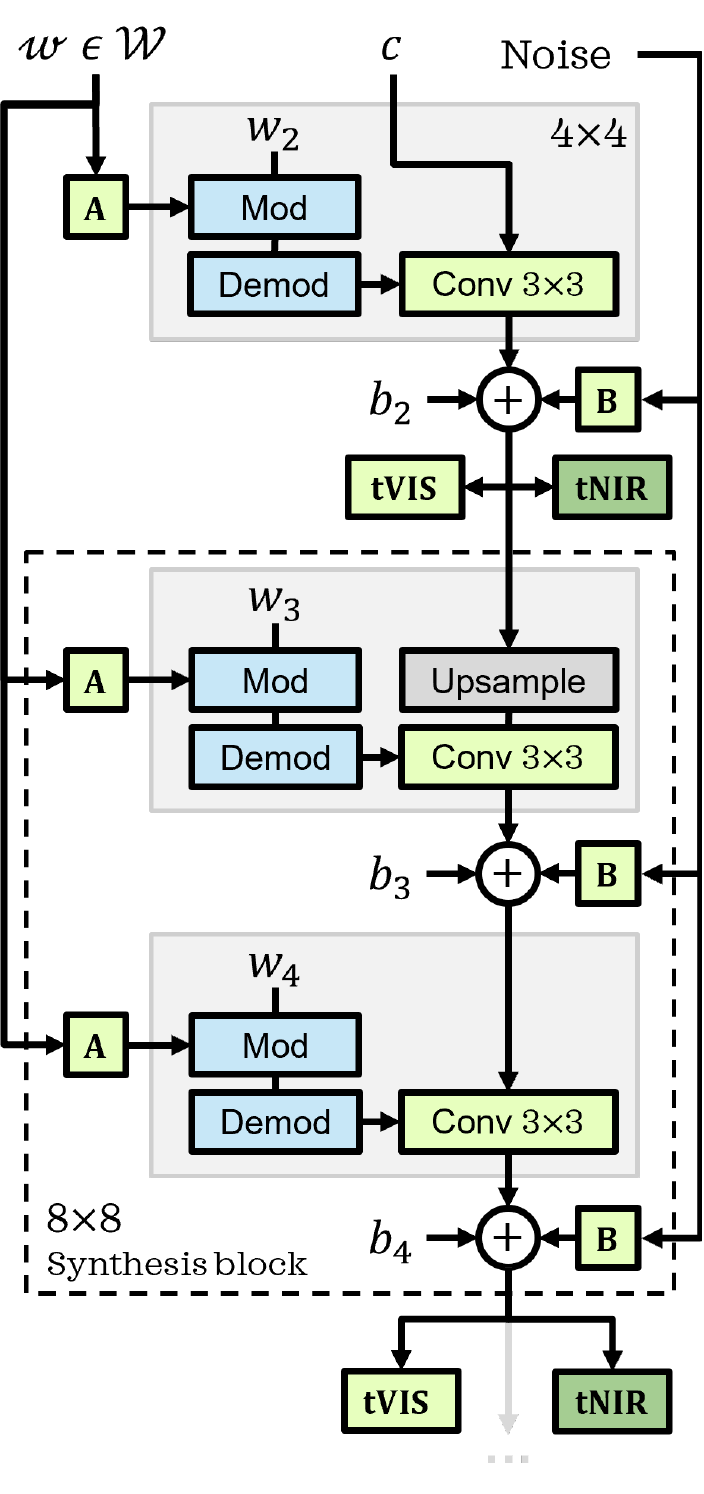} 
    &
    \includegraphics[width=0.22\textwidth]{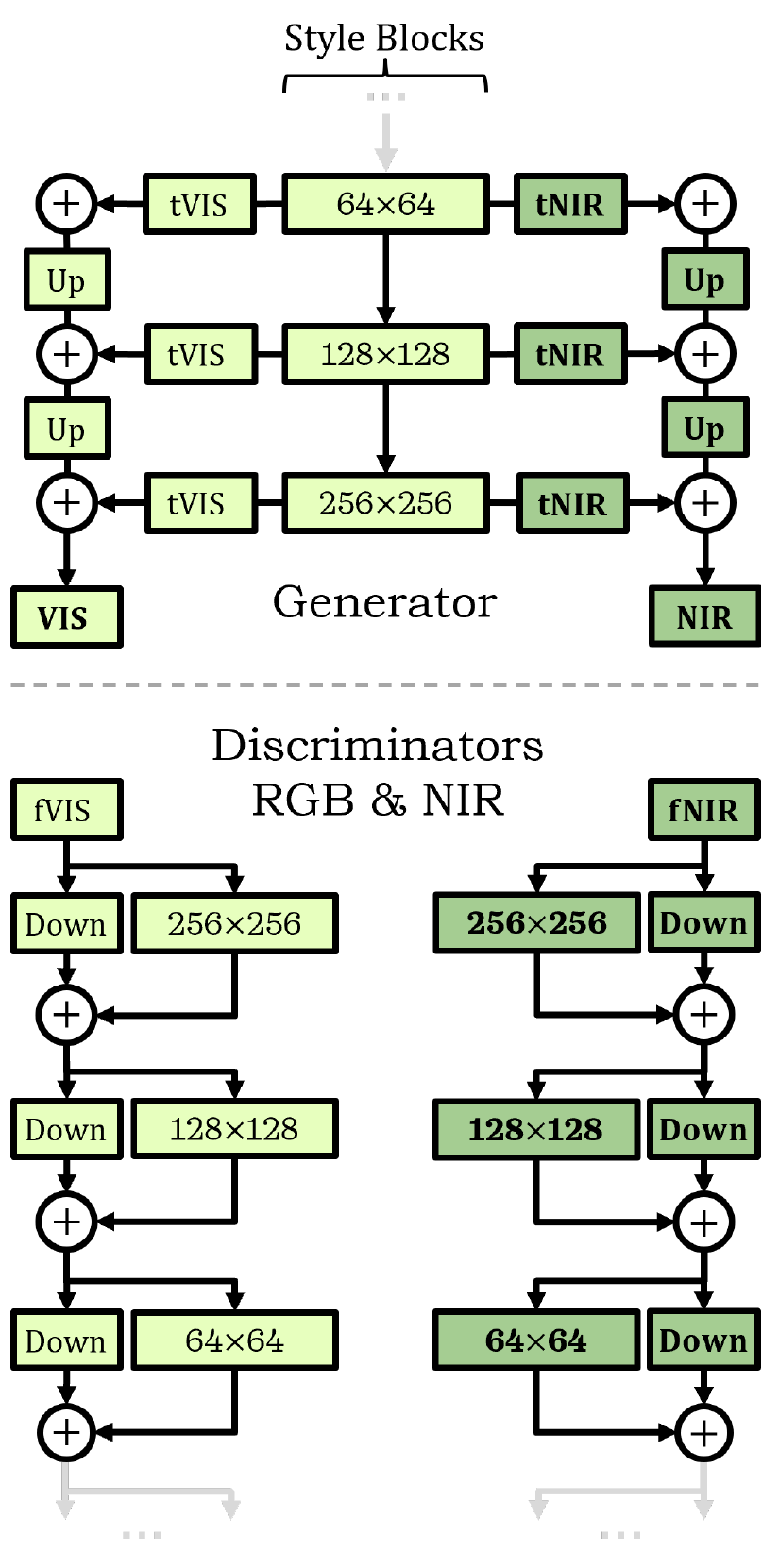}\\
    
\end{tabular}
\end{center}\vspace{-4mm}
\caption{\textbf{Overview of Dual-Branch StyleGAN2.} %The dark green blocks represents new additions to the original StyleGAN2 model. 
Each synthesis block simultaneously generates VIS and NIR images, via the tVIS and tNIR layers, which transform high-dimensional per-pixel data to images (fVIS and fNIR perform the opposite action). 
Outputs are then upsampled and passed to separate discriminators. In the synthesis network, $A$ and $B$ represent the style and noise inputs respectively. Leaky ReLU is applied after each $(+)$ in the generator. %Up and Down are bilinear up and downsampling,respectively.
\vspace{-3mm}
}
\label{fig:db_sg2_changes}
\end{figure}

The key component of BiOcularGAN is the novel Dual-Branch (DB) StyleGAN2 generator that extends the original StyleGAN2 \cite{stylegan2_ADA_karras2020training, stylegan2_karras2020analyzing} for bimodal data generation. As illustrated in Figure \ref{fig:db_sg2_changes}, the generator consists of a mapping network $f$ that follows a fully connected design, similarly to \cite{stylegan2_karras2020analyzing}, as well as a dual-branch synthesis network, and is trained using two discriminators, $D_{VIS}$ and $D_{NIR}$, one for the VIS and one for the NIR images. Details on the generator and discriminators are given below. %layer    DB-StyleGAN2 is made up of four neural networks. The mapping network and the synthesis network, which together form the Generator, as well as two Discriminator networks. 
%All parts are based on recent StyleGAN2 versions , however, they were adapted, as seen in Figure \ref{fig:db_sg2_changes}, to allow for generation of matching bimodal images. 

%%%%%%%%%%%%
\textbf{The Generator ($G$)} is responsible for producing the synthetic (NIR and VIS) ocular images and builds on recent insights and advancements in image generation \cite{stylegan2_ADA_karras2020training, stylegan2_karras2020analyzing}. Similarly to the original StyleGAN2 design, it consists of a succession of \textit{synthesis blocks} that produce images of progressively higher resolution, as shown on the left side of Figure \ref{fig:db_sg2_changes}. These consist of smaller \textit{style blocks} (light gray boxes), which take the intermediate latent representation $\mathbf{w}$, transformed through $k$ learned affine transformations $A$, as the style input. Convolution weights $w_x$ are then modulated based on the style input and later ``demodulated'' \cite{stylegan2_karras2020analyzing} -- a procedure which mimics the effects of instance normalization. Style is thus incorporated into the convolution operation via the processed weights.
% which is modulated based on weights $w_x$ and then passed to a so-called ``demodulation'' operation \cite{stylegan2_karras2020analyzing}, which mimics the effects of normalization. 
% This processed style is then incorporated into the convolutional weights. 
The network starts from a constant input $c$ ($4\times4\times512$). After each convolutional layer, the noise input (from the noise broadcast operation $B$) and bias $b_x$ are applied to the signal, which is then passed through a leaky ReLU activation function. 
% {\color{blue}{ TODO Darian [v stilu: pass them to XX and XX for XX and XX. The computer representations are processed by ... Content codes C, generated with per-pixel noises via the scaling networks B are also added [cite DiagonalGAN].}} 
A unique feature of the proposed DB-StyleGAN2 model that enables bimodal image generation are the dedicated synthesis blocks that contain two output branches, one for generating VIS and the other for generating NIR data at a specific resolution. Here, each branch features a $1 \times 1$ convolution layer, denoted tVIS (``toVIS'') and tNIR (``toNIR'') in  Figure~\ref{fig:db_sg2_changes}. The outputs of these branches are upsampled and merged with the output of the higher-resolution synthesis block to construct the final VIS and NIR images, thus, forming the DB-synthesis network, as seen on the right part of %capable of generating photo-realistic VIS and NIR images (
Figure~\ref{fig:db_sg2_changes}.            
% b is bias, B is noise, A is style code, 
% mod ... modulation
% demod ... demodulation (replaces instance normalization) .. cite stylegan2 "demodulate" , invert the modulation ... adds as normalization  
% leaky ReLU activation functions right after biases 
% modulation 
%as input The   For the purposes of our research, we modify the existing synthesis network of StyleGAN2 \cite{stylegan2_ADA_karras2020training} as follows. We add an additional $1 \times 1$ convolution layer $tNIR$  to each of its Style Blocks, similarly to existing $tRGB$ layers, as shown in Figure  \ref{fig:db_sg2_changes}. The outputs of all $tNIR$ layers are also upsampled and merged to construct the final NIR image, thus forming a dual-branch synthesis network, as depicted in Figure \ref{fig:db_sg2_changes}, capable of generating photo-realistic VIS and NIR images.      

%%%%%%%%%%%%
\textbf{The Discriminators ($D_{VIS}$, $D_{NIS}$)} aim to determine, whether images are real or artificially generated, and help to ensure that the data generated by the DB generator is as close to the training data distribution as possible. %To this end, the feedback is then used during training by the Generator to generate more photo-realistic images. 
For BiOcularGAN, we utilize two discriminators, $D_{VIS}$, $D_{NIS}$, one for each branch of the DB-synthesis network, corresponding to the VIS and NIR image modalities, as shown on the right side of Figure~\ref{fig:db_sg2_changes}. The discriminators take a pair of real (or fake) bimodal images as input and first pass them through $1 \times 1$ convolutional layers denoted as fVIS (``from VIS'') and fNIR (``from NIR''). The processed input is then passed through a ResNet-like \cite{he2016deep} downsampling architecture, with each block consisting of two convolution layers and a separate skip connection. 
%process {\color{blue}{them through TODO ...}} 
The output of each of the discriminators is a binary decision, i.e., real or fake. The two discriminators share the same architecture.

%%%%%%%%%%%%
\subsection{DB-StyleGAN2 training}\label{sec_DBStyle2_train}

Different from StyleGAN2, our dual-branch model produces two semantically similar output images in two distinct imaging domains. The training is, therefore, done with adversarial learning objectives involving two discriminators. Because of the (dual) bimodal output produced by DB-StyleGAN2, the training follows a multi-task learning regime, where the correlations between the two tasks (i.e., VIS and NIR image generation) help to efficiently capture the shared semantic content of the ocular images. Following the unimodal learning strategy used with StyleGAN2~\cite{stylegan2_ADA_karras2020training, stylegan2_karras2020analyzing} and insight from \cite{kwon2021diagonal, mescheder2018training}, we use a non-saturating soft-plus loss $s(x)=\log(1+\exp(x))$ with $R_{1}$ and path length regularization for the learning objectives:

{\begin{footnotesize}
\begin{gather}
    \small
    \mathcal{L}_{{\omega}} =  s(D_{\omega}(\mathbf{x}_{\omega}))+s(-D_{\omega}(\mathbf{y}_{\omega}))+\frac{\gamma}{2}\mathbb{E}\left[||\nabla D_{\omega}(\mathbf{y}_{\omega})||^2\right] \ \text{and} \\
% \end{equation}
% \begin{equation}
    \small
    % TODO
    % \mathcal{L}_{G} =  s(-D_{VIS}(\mathbf{x}_{vis}))+s(-D_{NIR}(\mathbf{x}_{nir})), \ \text{and} \\
    \mathcal{L}_{G} =  \sum_{\omega}{s(-D_{\omega}(\mathbf{x}_{\omega}))} + \gamma_{2}\, \mathbb{E}\Bigl(\Bigl\Vert\sum_{\omega}\nabla(\mathbf{x}_\omega q_\omega)\Bigr\Vert - a \Bigr)^2,
\end{gather}
\end{footnotesize}
}
where $\omega = \{VIS, NIR\}$, the synthetic images $\mathbf{x}$ are produced with Eq.~\eqref{eq:generetor} and $\mathbf{y}$ denotes real images, while $q$ represents an image with normally distributed pixel intensities and $a$ is the norm average.  Regularization parameters are computed using the resolution $r$ and batch size $bs$ via %the following heuristic formula 
$\gamma_{1} = 10^{-4} \frac{2 r^2}{bs}$ and $\gamma_{2} = \ln{2} / (r^2 (\ln{r} - \ln{2})$ \cite{stylegan2_karras2020analyzing}.

%Due to the proposed changes to the StyleGAN2 architecture \cite{stylegan2_karras2020analyzing, stylegan2_ADA_karras2020training} we must also modify the training procedure of the DB-StyleGAN2 model. While the process of computing all loss functions remains the same, we now have two loss functions $\mathcal{L}_{VIS}, \mathcal{L}_{NIR}$ that we need to take into consideration during backpropagation. Firstly, each loss function is used separately to train its corresponding Discriminator model. Next, the two loss functions are summed to create the global loss $\mathcal{L}$, which we then use to train the Generator model. This training procedure enables the generation of pixel-aligned bimodal images, while still retaining ensuring that both VIS and NIR images are as photo-realistic as possible.
%{\color{red}{} How loss is computed.... Loss same as StyleGAn2 and StyleGAN2-ADA} % path regularization 
%Add equations
%\begin{equation}
%    \mathcal{L}_{G} =  - \log D(G(f(z))) ,
%\end{equation}

%%%%%%%%%%%%
\subsection{Semantic Mask Generator (SMG)}\label{sec_SMG}

To generate \textit{ground truth semantic masks} for the bimodal images generated by the DB-StyleGAN2 model, we rely on the semantic information encoded in the feature maps produced along the DB-StyleGAN2 model during the synthesis process. To interpret the encoded information, we use an ensemble of Multi-layer Perceptron (MLP) classifiers, similarly to \cite{zhang2021datasetgan}, which are utilized within our Semantic Mask Generator (SMG) to predict the semantic class label of each pixel in the generated bimodal ocular data. %and are used with a majority voting strategy to minimize the randomness of the training procedure. 

However, different from the procedure of Zhang \textit{et al.}~\cite{zhang2021datasetgan}, we extract feature maps from each Leaky ReLU activation function
%after each of the $k$ convolutional layers 
in the dual-branch synthesis network (in Figure~\ref{fig:db_sg2_changes}), related to a single style and resolution.
% from which the bimodal images of a certain resolution are generated. 
This allows us to capture the semantic information of the bimodal ocular images before they are rendered in a certain imaging domain. %close as possible to the make efficient use of the dual  
%The semantic mask generator requires a small set of manually annotated images with properly represented semantic classes to be able to train the MLPs.   
%However, due to the differences in StyleGAN architectures, we extract feature maps at slightly different stages of the network. More accurately, we obtain our feature maps after each convolution layer in the synthesis network depicted in Figure \ref{fig:db_sg2_changes}.
We then upsample these feature maps to the output resolution and construct a $W \times H \times d$ tensor, from which $d$-dimensional feature vectors\footnote{Here, $d$ denotes the combined length of all extracted feature maps.}
% is the total number of feature maps extracted from the $k$ layers.}
corresponding to each of the $WH$ image pixels can be obtained. Using the obtained high-dimensional feature vectors as input, we train an ensemble of $10$ three-layer MLPs to classify pixels into the semantic classes. Here, manual annotations over an incredibly small set ($<10$)  of generated bimodal images are used as the ground truth for the training procedure. We note that  a majority voting strategy is utilized over the predictions of the MLP ensemble to minimize the randomness of the learning stage. Once trained, the SMG can be used together with the DB-StyleGAN2 model to generate unlimited amounts of pixel-level aligned bimodal ocular images with corresponding semantic ground truth masks. Here, a single forward pass is needed to generate one triplet $\{\mathbf{x}_{vis},\mathbf{x}_{nir},\Omega\}$.   %as the targeted semantic ground truthwithin our SMG  and obtain feature vectors for each pixel, which describe the pixel's style. These vectors are then passed to the MLP classifiers, which predict the class of each generated pixel based on the corresponding feature vectors extracted from the synthesis network. 

\begin{table}[t!] %  use * at the table* for double column
\begin{center}
\resizebox{\columnwidth}{!}{%
\begin{tabular}{lcccccc}
\toprule
    \bf{Dataset} & \bf{\# Images} & \bf{\# IDs} &  \bf{\# Eyes} & \bf{Resolution} & \bf{Modality$^\dagger$} &  \bf{Purpose$^\ddagger$}   \\ 
    \midrule
    PolyU~\cite{nalla2016toward} & $12540$ & $209$ & $518$& $640 \times 480$ &  NIR/VIS &    TR/SV \\
    CrossEyed~\cite{sequeira2016cross, sequeira2017cross}& $3840$ & $120$ & $240$ &$400\times 300$ &NIR/VIS &  TR/SV\\
    \midrule
    SMD~\cite{das2017towards} & $500$ & $25$ & $50$ &$3264 \times 2448$ &VIS &  SE \\
    MOBIUS~\cite{SSBC2020} & $3542$ & $35$ & $70$ & $ 3000 \times 1700$ & VIS &  SE \\
    SBVPI~\cite{vitek2020comprehensive,rot2020deep} & $1858$ & $55$ & $110$ & $ 3000 \times 1700$ & VIS & SE\\ 
\bottomrule
\multicolumn{7}{l}{$^\dagger$NIR -- near-infrared, VIS -- visible light}\\
\multicolumn{7}{l}{$^\ddagger$TR -- training, SV -- synthesis validation, SE -- segmentation experiments}
\vspace{-5mm}
\end{tabular}}
\end{center}
\caption{\textbf{Summary of the experimental dataset.} We train (and validate) all components of BiOcularGAN on the cross-spectral datasets and evaluate segmentation performance on the visible spectrum datasets.\vspace{-4mm}}
\label{tab:dataset_comparison}
\end{table}
% To interpret this information we rely on an ensemble of Multi-layer Perceptron (MLP) classifiers, which predicts the class of each generated image pixel based on the feature maps extracted from the DB-synthesis network. 

%
% However, our approach gets ... instead of after ADA-In layer ... we get it where?

% The mask generation procedure relies on the use of an ensemble of Multi-layer Perceptron (MLP) classifiers, which predicts the class of each generated image pixel based on the feature maps extracted from the DB-synthesis network. 

% use MLP as a style interpreter ... feature maps basically define the style 

% we rely on the semantic information of the feature maps used to produce said images.

% Similar procedure to DatasetGAN  .. except based on StyleGAN2 architecture .. Upsample feature maps of the Synthesis network .... (where are they taken from ... sicne we don't have AdaIN? as in StyleGAN1) ... 

% DatasetGAN as style interpreter
% (similar to DatasetGAN, but based on StyleGAN2 instead of 1) .. get latent vectors at slightly different stages

% based on the idea that if feature maps encode the information of each pixel .. then they must also contain some form of semantic information ... regarding its designated class for example ...

%%---------------------------------------------------
% Experiments
%----------------------------------------------------
\section{Experiments and result}\label{Sec:Experiments}

%  datasets (PolyU cross-spectral \cite{nalla2016toward}, CrossEyed \cite{sequeira2016cross, sequeira2017cross}, SMD \cite{das2017towards}  )

%\iffalse
\begin{figure*}[t!] 
\begin{center}
\resizebox{0.9\textwidth}{!}{%
\begin{tabular}{cc|c|c|c}
%\toprule
    %\midrule
    & \small PolyU samples &\small CrossEyed samples&\small DB-StyleGAN2-P &  \small DB-StyleGAN2-CE\\ 
    \rotatebox{90}{\hspace{10mm} \small VIS images} &
    % \includegraphics[width=0.21\textwidth]{figures/3x3_samples/PolyU_256_ALL_RGB_samples.pdf} &
    % \includegraphics[width=0.21\textwidth]{figures/3x3_samples/CrossEyed_256_ALL_RGB_samples.pdf} &
    % \includegraphics[width=0.21\textwidth]{figures/3x3_samples/GENERATED_DATASETS_PolyU_NIR_RGB_5000_RGB_samples.pdf} &
    % \includegraphics[width=0.21\textwidth]{figures/3x3_samples/CrossEyed_NIR_RGB_train_5000_320_1RGB_samples_new.pdf}\\
    % \hline 
    % & &&&\vspace{-3mm}\\ 
    % \rotatebox{90}{\hspace{10mm} \small NIR images} & 
    % \includegraphics[width=0.21\textwidth]{figures/3x3_samples/PolyU_256_ALL_NIR_samples.pdf} &
    % \includegraphics[width=0.21\textwidth]{figures/3x3_samples/CrossEyed_256_ALL_NIR_samples.pdf} &
    % \includegraphics[width=0.21\textwidth]{figures/3x3_samples/GENERATED_DATASETS_PolyU_NIR_RGB_5000_NIR_samples.pdf} &
    % \includegraphics[width=0.21\textwidth]{figures/3x3_samples/CrossEyed_NIR_RGB_train_5000_320_1NIR_samples_new.pdf}\\ 
    % & \multicolumn{2}{c}{\small Original images}&\multicolumn{2}{c}{\small Synthesized images}\\ 
     \includegraphics[width=0.21\textwidth]{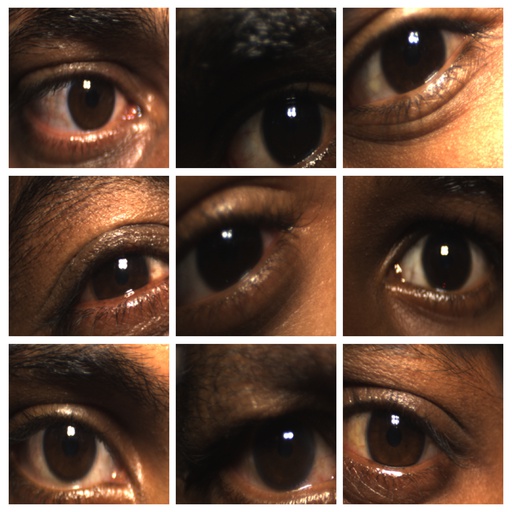} &
    \includegraphics[width=0.21\textwidth]{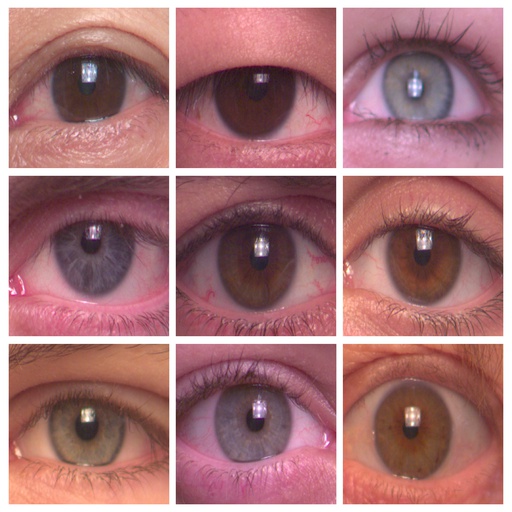} &
    \includegraphics[width=0.21\textwidth]{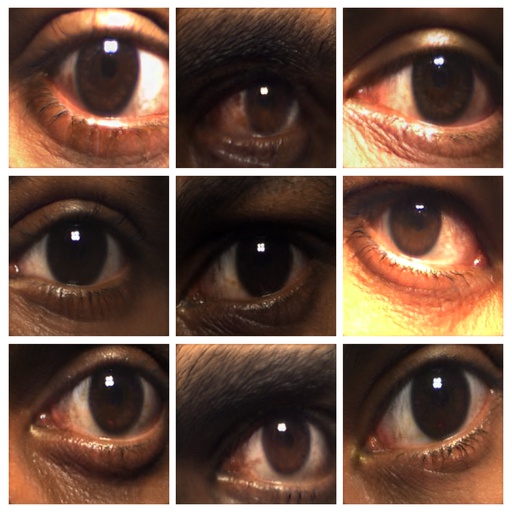} &
    \includegraphics[width=0.21\textwidth]{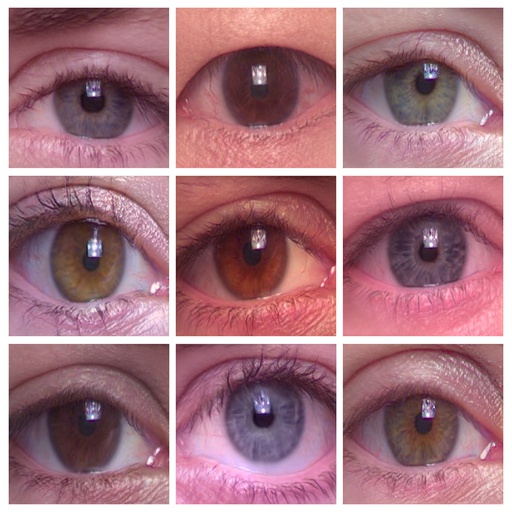}\\
    \hline 
    & &&&\vspace{-3mm}\\ 
    \rotatebox{90}{\hspace{10mm} \small NIR images} & 
    \includegraphics[width=0.21\textwidth]{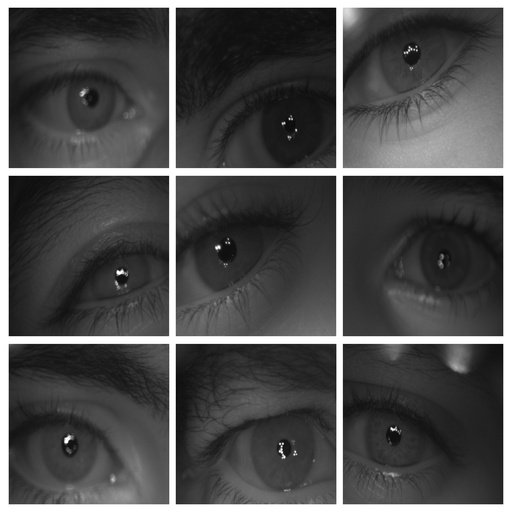} &
    \includegraphics[width=0.21\textwidth]{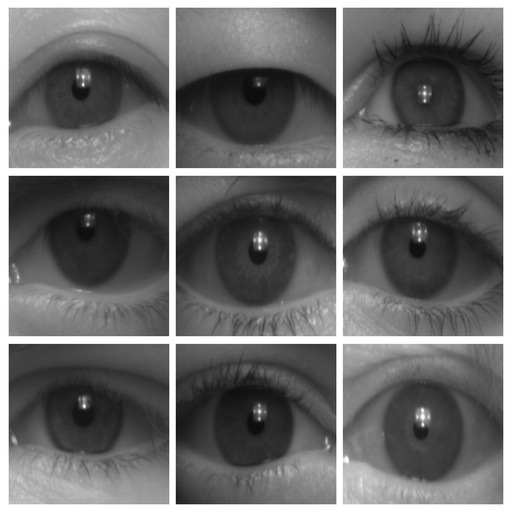} &
    \includegraphics[width=0.21\textwidth]{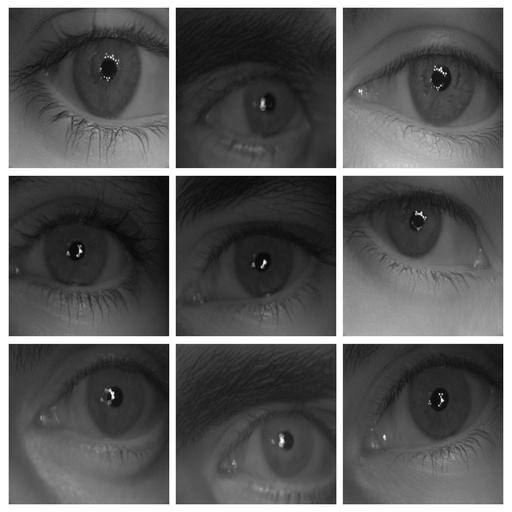} &
    \includegraphics[width=0.21\textwidth]{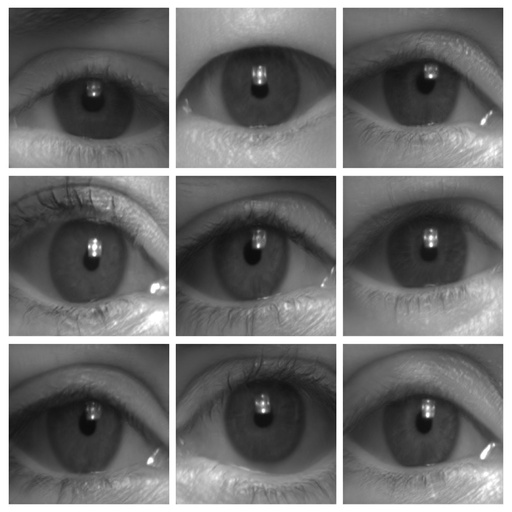}\\ 
    & \multicolumn{2}{c}{\small Original images}&\multicolumn{2}{c}{\small Synthesized images}\\ 
\end{tabular}}
\end{center}\vspace{-6mm}
\caption{\textbf{Visual examples of original and generated ocular images in both domains.} The first two columns show samples from the PolyU and CrossEyed datasets and the last two columns show examples of images generated by the DB-StyleGAN2 models trained on the PolyU (DB-StyleGAN2-P) and CrossEyed (DB-StyleGAN2-CE) datasets.} \vspace{-4mm}
\label{tab:generated examples}
\end{figure*}
%\fi

\subsection{Experimental setup}

\textbf{Datasets.} We use five datasets for training and evaluation of BiOcularGAN, i.e., the PolyU cross-spectral Iris database (PolyU) \cite{nalla2016toward}, CrossEyed \cite{sequeira2016cross, sequeira2017cross}, the Sclera Mobile Dataset (SMD) \cite{das2017towards}, SBVPI~\cite{rot2020deep, vitek2020comprehensive} and MOBIUS \cite{SSBC2020}. The main characteristics of the datasets are summarized in Table \ref{tab:dataset_comparison}, while the key details are provided below:

\begin{itemize}
    \item \textbf{Cross-spectral datasets:} The PolyU and CrossEyed datasets contain ocular images captured in the near-infrared (NIR) and visible light (VIS) spectra. The acquisition procedure for both datasets was performed with custom sensors capable of simultaneous acquisition of the NIR and VIS images. The images in PolyU are aligned with pixel-level correspondences, while the CrossEyed data is loosely aligned, i.e., with small (random) perturbations in scale and position in the NIR-VIS image pairs. For our experiments, the image pairs of both datasets are split into subject disjoint training and evaluation parts in a ratio of $9:1$. The training part is used to learn the DB-StyleGAN2 model and SMG annotation procedure, whereas the (hold-out) evaluation part is reserved for the performance evaluation. %For concrete numbers see  
    \item \textbf{Visible spectrum datasets:}  The SMD, SBVPI and MOBIUS datasets consist of  high-resolution VIS ocular images captured primarily for research into sclera biometrics. All three datasets have manual annotations of some key regions of the ocular images, e.g., the sclera, iris or pupil, and are therefore used to evaluate the performance of the segmentation models trained with the annotated data generated by BiOcularGAN. %Note also that SMD features mostly subjects of Asian origin (similarly to PolyU), whereas the subjects in MOBIUS are predominantly Caucasian (similarly to CrossEyed) The dataset 
\end{itemize}
 % and a few sample images are shown in Figure \ref{tab:generated examples}.

% datasets (PolyU cross-spectral \cite{nalla2016toward}, CrossEyed \cite{sequeira2016cross, sequeira2017cross}, SMD \cite{das2017towards}  )

\textbf{Implementation Details.} All components of BiOcularGAN were implemented in PyTorch and are made publicly available from URL\footnote{\url{https://github.com/dariant/BiOcularGAN}}. The %main component of BiOcularGAN is the 
Dual-Branch StyleGAN2 is implemented based on the %StyleGAN2 model \cite{stylegan2_karras2020analyzing}, more specifically the 
StyleGAN2-ADA variant \cite{stylegan2_ADA_karras2020training}. %Despite changes to the architecture, to enable bimodal image generation, the training parameters and procedures share many similarities with the original implementation.
%As a starting point, we begin our training of 
The main part of the DB-StyleGAN2 is initialized with weights pretrained on the FFHQ dataset (of resolution $256 \times 256$) and then optimized further using the Adam optimizer~\cite{kingma2014adam} with a learning rate of $0.0025$ and a batch size of $16$. For the other hyperparameters, we use the recommended values $\beta_{1} = 0$, $\beta_{2} = 0.99$, and $\epsilon=10^{-8}$ for both, the generator and the two discriminators. We train all models for $2500 \ kimgs$ or until training diverges, due to the low amount of training data. To combat model divergence, we enable data augmentation in the form of horizontal image flipping and additionally employ the adaptive discriminator augmentation procedure proposed in \cite{stylegan2_ADA_karras2020training}.
% Trained for 2500kimgs or until divergence ...  due to the limited amount of data 
% DB-StyleGAN2 was trained with the Adam optimizer ... {\color{red}{}}   
% Pretrained FFHQ pretrained at $256 \times 256$ resolution ... 
% Share many similarities with recommended training options for StyleGAN2 ... 
% Mapping network 8 layers, input 512 dimensional latent (both z and intermediate) .., allowing flips along the $x$ axis, ... resolution 256 ... 
% Adam optimizer ... LR = 0.0025 for both the Generator and Discriminator training ... betas (0, 0.99). .., eps =1e-8 
% r1 gamma = computed from something write it 
% batch size = 16 
% DatasetGAN
For the Semantic Mask Generator (SMG), %we use an ensemble of $10$ MLP classifiers, trained with the manually annotated images to predict the class of each pixel. T
training is performed based on the cross-entropy loss  and the Adam optimizer \cite{kingma2014adam}, with a learning rate of $10^{-3}$. Each MLP classifier is trained on randomly sampled image pixels in batches of $64$. The training is stopped once no improvement is observed in the learning objective over $50$ batches following the third epoch, similarly to \cite{zhang2021datasetgan}. Additional implementation details can be found in the publicly released source code.

% and training is stopped when convergence is achieved ... each classifier is stopped trianing when hasn't improved in 50 batches ... randomly samples images

% to predict the class of each pixel based on the feature maps extracted from the DB-synthesis network.

% Then the DatasetGAN stuff.. how we train classifiers .... on the generated synthetic datasets 
% 10 classifiers 
% 8 hand annotated images by the same annotator 
% Adam optimizer Lr = 0.001 
% CrossEntropy Loss 
% batch size = 64 

% DeepLab and U-Net:
% batch size = 8 
% Adam:  lr = 1e-4 
% epochs 50 ... with Learning rate decreasing by factor 10, with patience 5 ... 
% ReduceLROnPlateau 
% with crossEntropy loss or Binary cross entropy, when only 2 classes ...

% \iffalse
% \begin{itemize}
%     \item \textbf{DB-StyleGAN2} ... based on StyleGAN2 ... Main difference Dual  Branch in Synthesis network to generate both RGB and NIR images (they share latent features) ... uses a separate Discriminator for each image type ... 
%     The Discriminators are based on StyleGAN2 - ADA ... augmentations used due to the limited amount of training data
%     {\color{red}{TODO}}  
    
%     \item \textbf{Mask Generator} ... based on the DatasetGAN approach ... We use an ensemble of 10 classifiers to classify each pixel in an image based on its latent vectors .. Main improvement: manual annotation can now be performed on either the RGB or NIR image, since they match .. Can make annotations easier and more accurate  ... 8 images per approach, by the same annotator
    
% \end{itemize}
% \fi

\textbf{Experimental Hardware.} All experiments are conducted on a Desktop PC with an Intel i9-10900KF CPU with $64$ GB of RAM and an Nvidia $3090$ GPU with $24$ GB of video RAM. Using this hardware, we trained two DB-StyleGAN2 models, one on PolyU and one on CrossEyed, denoted as \textbf{DB-StyleGAN2-P} and \textbf{DB-StyleGAN2-CE} hereafter. Once converged, the models are able to generate visually convincing bimodal ocular images of $256 \times 256$ pixels in size, as demonstrated in the following sections.   %  The training procedure takes approximately $24$ hours for DB-StyleGAN2-P and $20$ hours for DB-StyleGAN2-CE to complete, starting from a model pretrained on the Flickr-Faces-HQ (FFHQ) dataset \cite{stylegan_1_karras2019style}. Once trained, DB-StyleGAN2 is able to generate a single pair of bi-modal ocular images of $256 \times 256$ pixels in size in $14$ ms on average (computed over $1000$ test images). The training of the data annotation procedure took around $13$ minutes on PolyU and $11$ on CrossEyed and required only $8$ annotated images per dataset. At run-time, a single segmentation mask is produced in $77.8$ ms. Additional implementation details can be found in the publicly released source code.     

%\subsection{Performance criteria}

\subsection{Synthesis evaluation}

In the first set of experiments, we explore the capabilities of the trained DB-StyleGAN2 models.

\textbf{Visual Evaluation.} Figure~\ref{tab:generated examples} shows a selection of (real) VIS and NIR images from the PolyU and CrossEyed datasets, as well as a few examples generated by the two trained DB-StyleGAN2 models. As can be seen, both models are capable of generating high-quality and visually convincing images that well match the visual characteristics of the training data in the visual as well as near-infrared domain. The trained models  are able to synthesize crisp image details, such as individual eyelashes, eyebrows, skin textures and even reproduce the specular reflections present in the training samples. Due to the dual-branch design of the DB-StyleGAN2 model, these fine image details are also consistent across the bimodal image pairs.  

\begin{figure}[t]
%\iffalse
%\resizebox{\columnwidth}{!}{%
%    \begin{tabular}{ccccc}
%    \vspace{2mm}
%    \Huge PolyU  & \Huge DB-StyleGAN2-P && \Huge CrossEyed & \Huge DB-StyleGAN2-CE \\ %\vspace{2mm}
%    \includegraphics[width = 0.7\columnwidth]{figures/YCbCr_comparison/Luma_PolyU_256_ALL_031_L_8_real.pdf}&
%    \includegraphics[width = 0.7\columnwidth]{figures/YCbCr_comparison/Luma_PolyU_NIR_RGB_5000_2998.pdf}&&
%    \includegraphics[width = 0.7\columnwidth]{figures/YCbCr_comparison/Luma_CrossEyed_256_ALL_u121_vw_004_real.pdf}&
%    \includegraphics[width = 0.7\columnwidth]{figures/YCbCr_comparison/Luma_CrossEyed_NIR_RGB_5000_522.pdf}\vspace{1mm}
%    \\ \vspace{2mm}  
%\end{tabular}\vspace{-2mm}}
%\fi

\begin{flushleft}
\footnotesize \hspace{6mm} PolyU \hspace{6mm} DB-StyleGAN2-P \hspace{3mm} CrossEyed \hspace{1mm} DB-StyleGAN2-CE\vspace{-3mm}
\end{flushleft}
\centering
\includegraphics[width = 0.99\columnwidth]{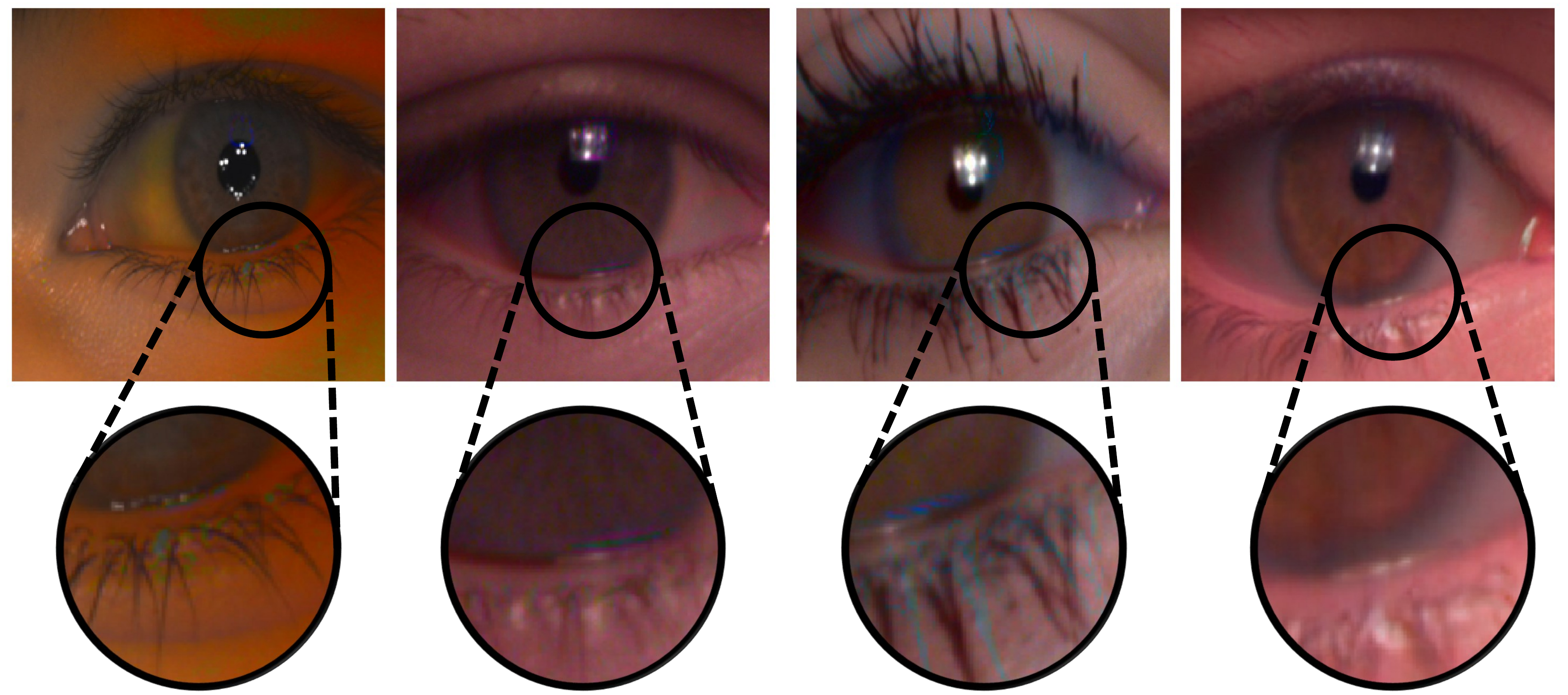}
    \vspace{-1mm}
    \caption{\textbf{Illustration of NIR-VIS alignment.} Shown are composite images, where the luma channel in the VIS image (in the YCbCr space) was replaced by the NIR image. %Note that the trained DB-StyleGAN2 models generate per-pixel aligned ocular image pairs. % {\color{red}Zamenjava zadnjo sliko z novim modelom?} 
    \vspace{-5mm}
    } %The left image in each pair represents the synthesized image and the right an actual image from the training data. 
   % Note that the models learn to generate novel data instances that share important semantic characteristics with the training images. The Mean Squared Error (MSE) for each pair is reported below the images.}
    \label{fig:Alignment}
\end{figure}
%\iffalse
\begin{figure}[t]
%\resizebox{\columnwidth}{!}{%
\centering
%\begin{tabular}{ccc}
\resizebox{0.8\columnwidth}{!}{%
\hspace{9mm} \small DB-StyleGAN2-P/PolyU   \hspace{11mm} \small DB-StyleGAN2-CE/CrossEyed
}%\vspace{2mm}
\newline
    %\end{tabular}}\\ 
    \resizebox{0.8\columnwidth}{!}{%
    \begin{tabular}{cc}
    \vspace{2mm}
    \includegraphics{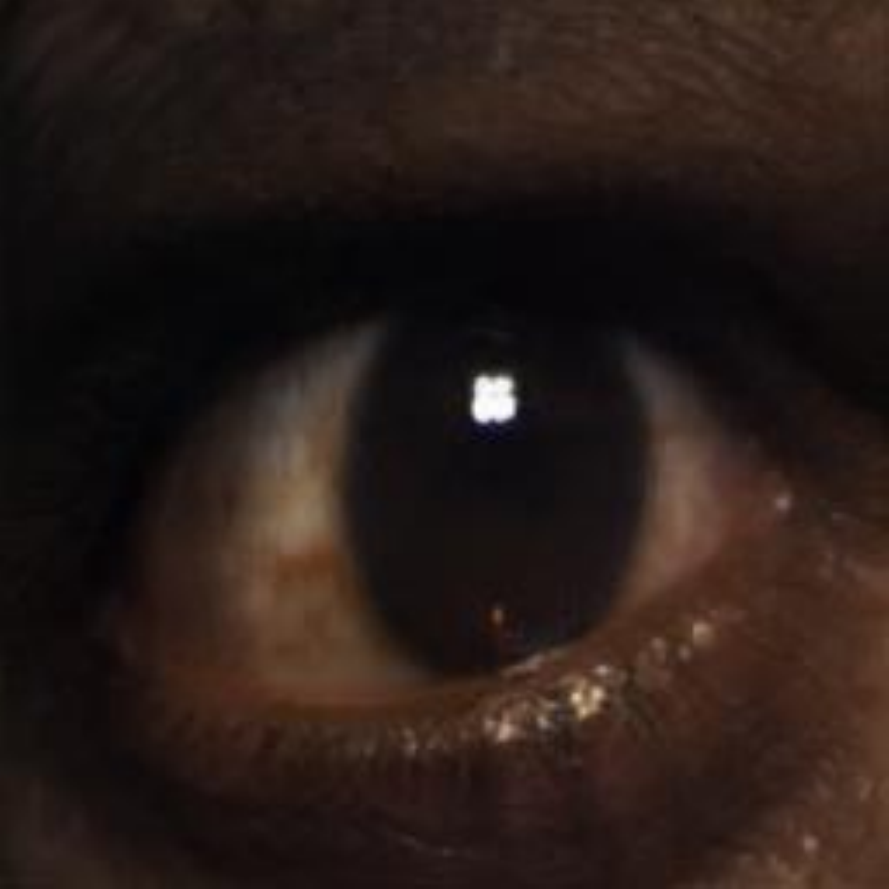}
    \includegraphics{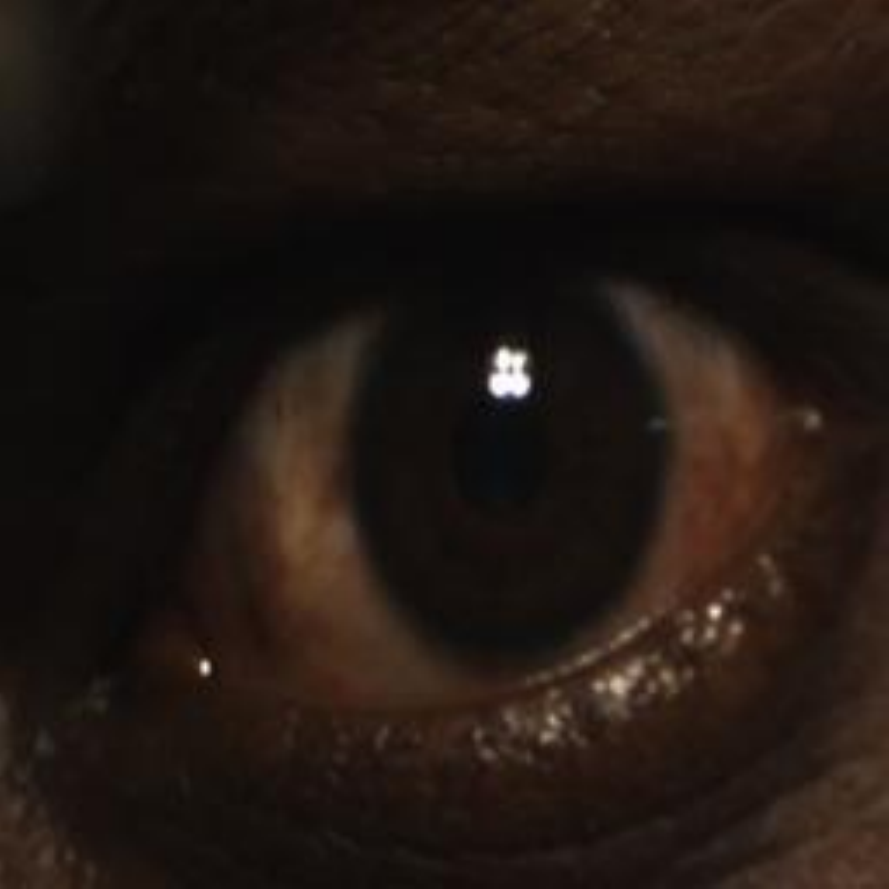}&
    \includegraphics{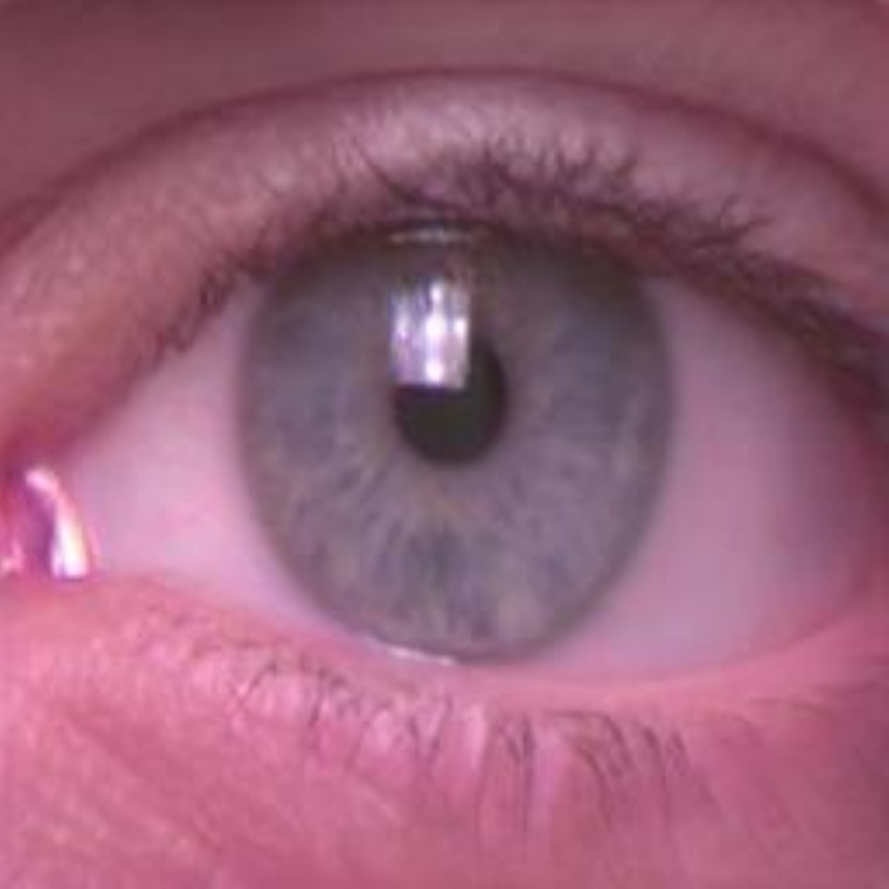}
    \includegraphics{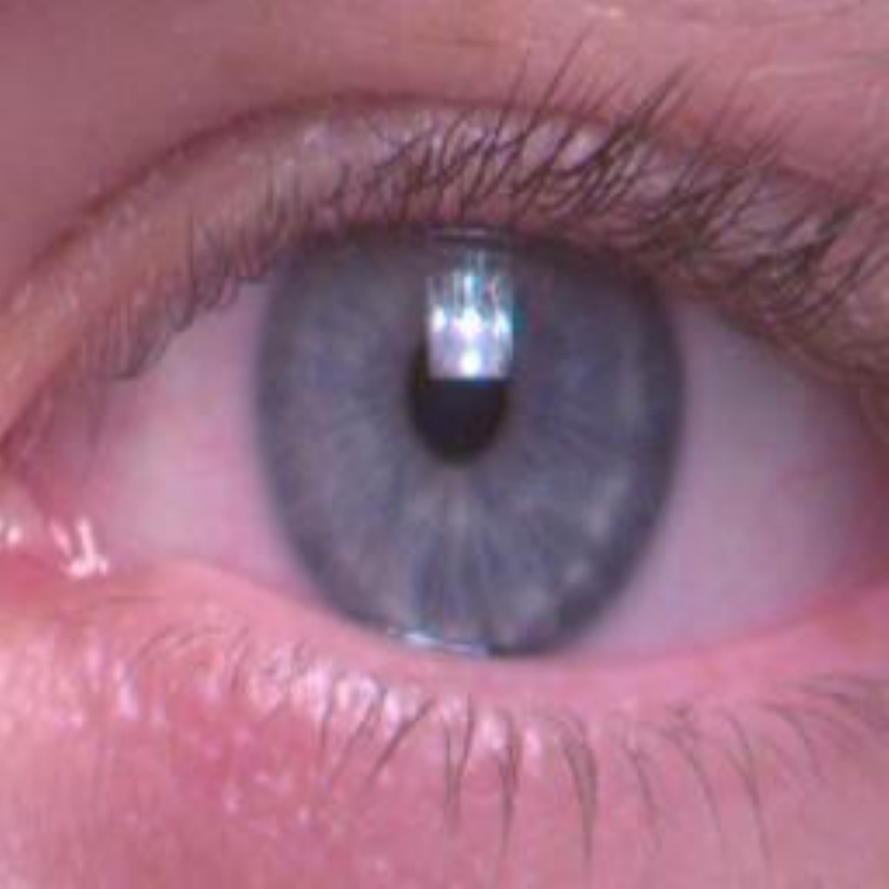}\vspace{1mm}
    \\ \vspace{2mm}  
    %\huge MSE = $1072.584$ &  \huge MSE = $1452.520$\\ %\vspace{1mm}
    \includegraphics{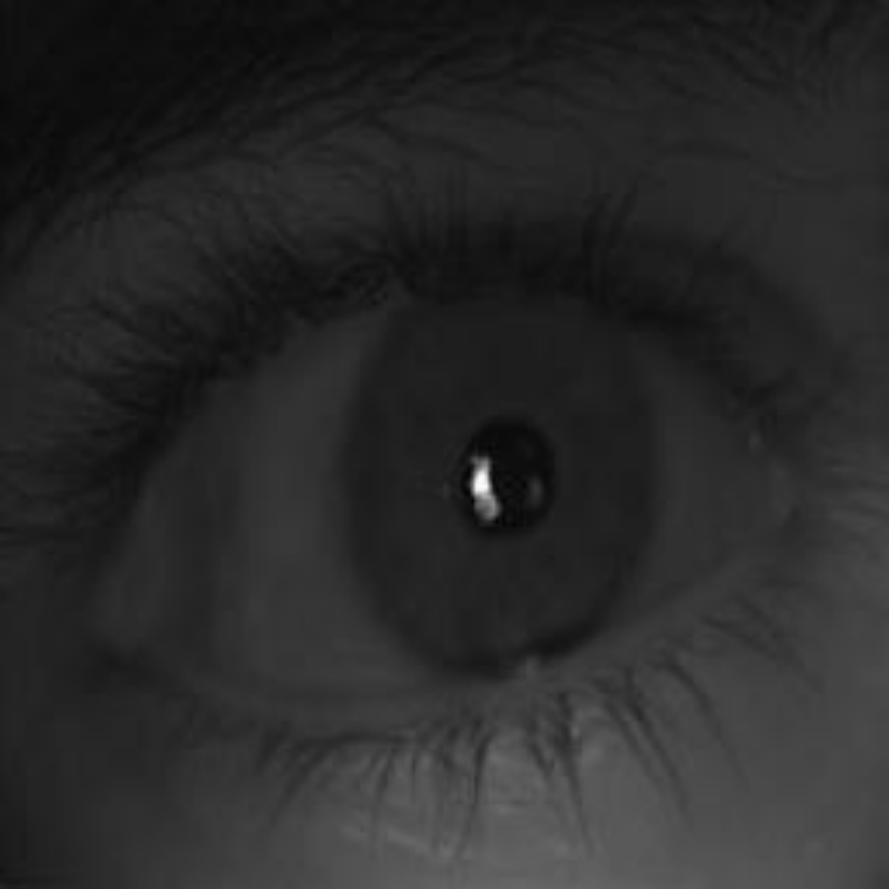}
    \includegraphics{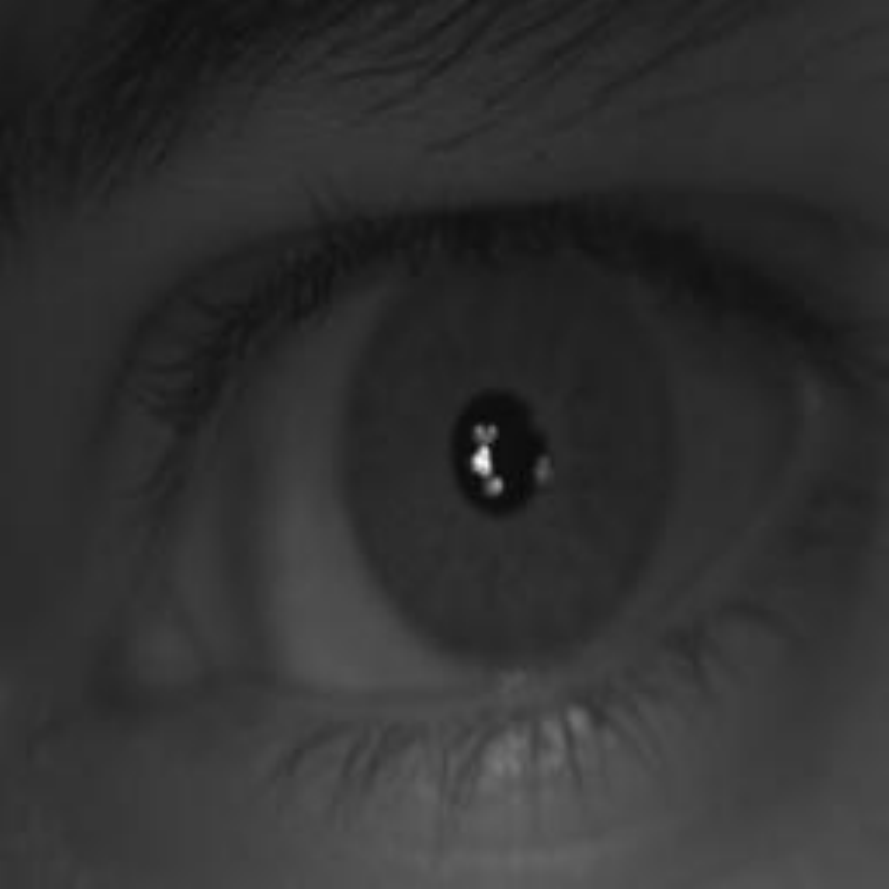}&
    \includegraphics{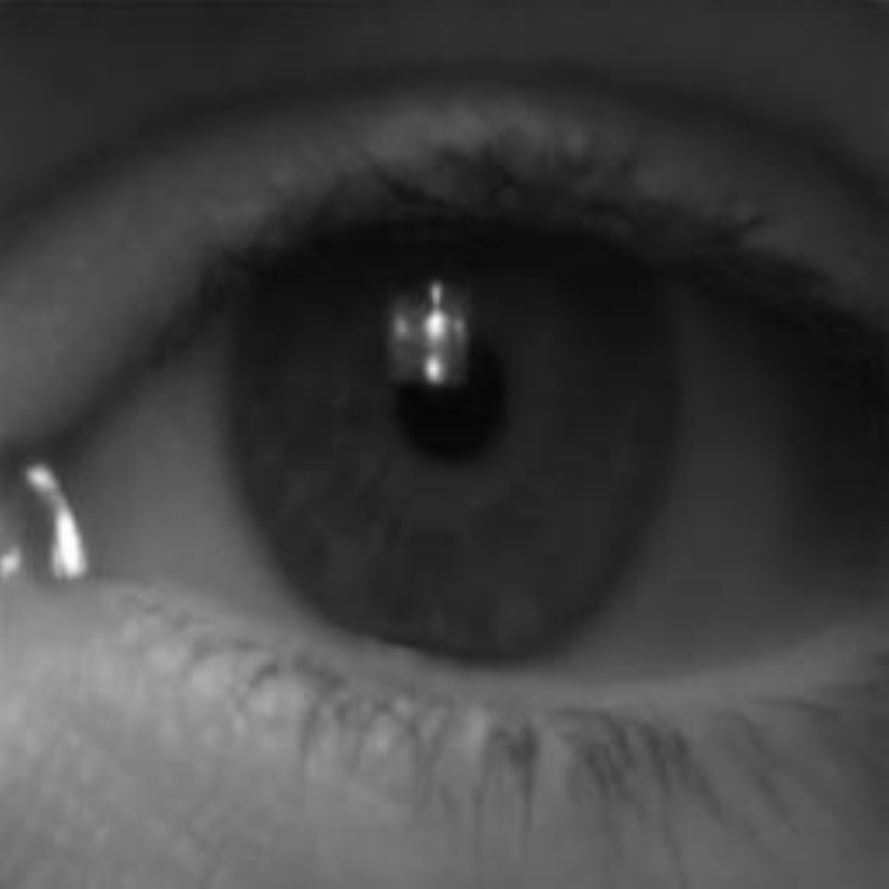} 
    \includegraphics{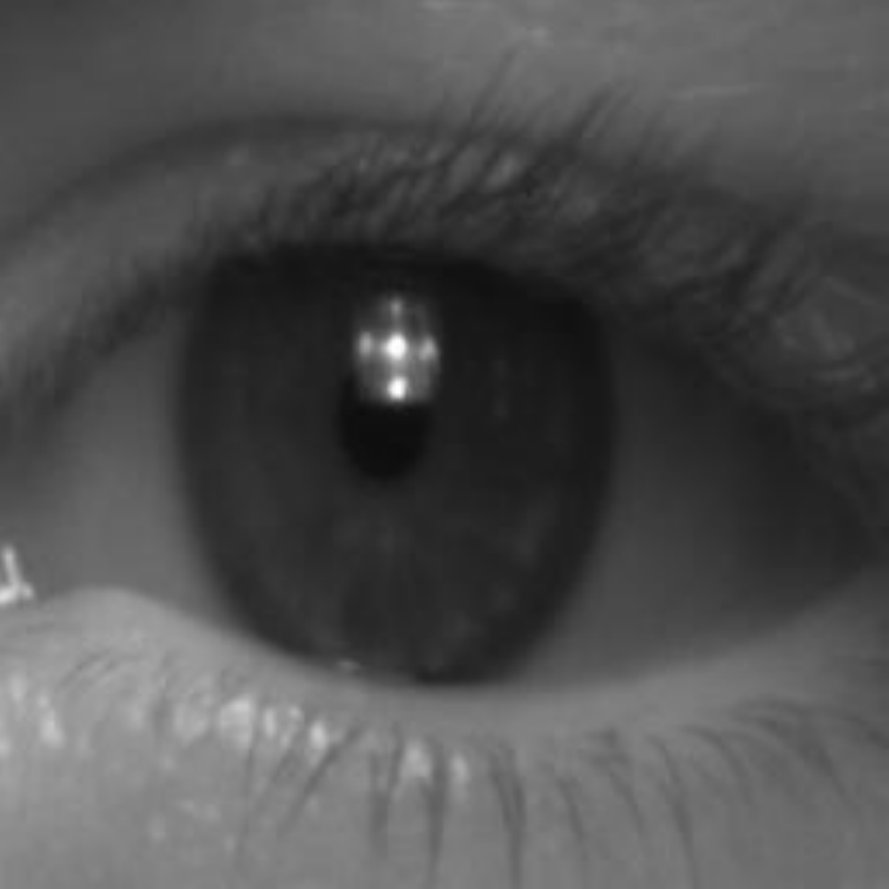}\vspace{1mm}
    % \\ \huge MSE =  $399.77$ & \huge MSE = $627.054$\\ 
    \end{tabular}\vspace{-2mm}}
    \vspace{-2mm}
    \caption{\textbf{Generated sample images (left) and nearest samples (right) from the training set.} %The left image in each pair represents the synthesized image and the right an actual image from the training data. 
    Note that the models learn to generate novel data instances that share important semantic characteristics with the training images.\vspace{-4mm}} %The Mean Squared Error (MSE) for each pair is reported below the images.}
    \label{fig:HQexamples}
\end{figure}
%\fi

\textbf{VIS-NIR Pair Alignment.} While DB-StyleGAN2-P was trained on the per-pixel aligned data from PolyU, the training of DB-StyleGAN2-CE was performed with the loosely aligned images from CrossEyed. Nonetheless, both models produce well-aligned NIR and VIS images due to the shared style blocks in the StyleGAN2 model that capture the semantics of the ocular images, while the two branches %serve as renderers that 
generate the final output images within the specific imaging domains. To visualize the alignment of the original and synthesized image pairs, we generate composite images, where the RGB data from the VIS samples is first transformed into the YCbCr color space and the luma (Y) component is then replaced by the NIR channel. This composition changes the overall color characteristics of the images, as most clearly seen by the PolyU examples in Figure \ref{fig:Alignment}, which now exhibit eye-color and skin-tone changes, % now appear to originate from Caucasian subjects, 
but also highlights the misalignment between the two image domains in the form of color artifacts.

As can be observed, % from the presented examples, 
there is little color artifacts in the original PolyU data and corresponding composite image generated with the DB-StyleGAN2-P model, suggesting that the VIS and NIR data are well aligned in both cases. The only artifacts present are due to differences in specular reflections. Conversely, there is obvious misalignment  in the CrossEyed data, as evidenced by the eyelash-shaped color artifacts. Nevertheless, the trained DB-StyleGAN2-CE model still generates well aligned bimodal ocular images. The loose alignment of the training data has no adverse effect on the alignment of the synthesized images. % generated by the learned model.  

% Observation: in the CrossEyed dataset some images are quite misaligned, but not in our artificial datasets (can add examples)

\textbf{Image Diversity.} Next, we qualitatively analyze the  diversity of the  images generated by the trained DB-StyleGAN2 models. % beyond the training examples. In other words
Specifically, we are interested in the variations of ocular images the models are able to produce with respect to the data seen during training. To this end, we show in Figure \ref{fig:HQexamples} a randomly generated image pair produced by the DB-StyleGAN2-P and DB-StyleGAN2-CE models (left column of each presented example) as well as the most similar VIS-NIR pair from the training data -- where the similarity is measured in terms of Mean Squared Error (MSE) between the VIS images. Several interesting observations can be made from the presented examples, i.e.: $(i)$ the generated images share obvious similarities with the training data in terms of visual appearance, $(ii)$ the models generate distinct data samples that differ from the training examples in terms of gaze direction, eye shape and color (for VIS), eyelash arrangement, eyelid appearance, pupil size, skin and iris texture, and other factors, and $(iii)$ despite appearing similar in the VIS domain at first glance, considerable differences are present in the NIR domain in the presented examples, suggesting that the combined (bimodal) ocular images generated by the models are distinct.

%\iffalse
\begin{figure}[t!] 
\begin{center}
\resizebox{\columnwidth}{!}{%
\begin{tabular}{cc|c|c|c}
%\toprule
    %\midrule
    & \large StyleGAN2 (PolyU) & \large StyleGAN2 (CrossEyed) & \large DB-StyleGAN2-P &   \large DB-StyleGAN2-CE\\ 
    \rotatebox{90}{\hspace{8mm} \Large VIS images} &
    % \includegraphics[width=0.23\textwidth]%,trim = 0 85mm 85mm 0,clip]
    %     {figures/samples_SOTA_comparison/GENERATED_DATASETS_PolyU_RGB_5000_samples.pdf} &
    % \includegraphics[width=0.23\textwidth]{figures/samples_SOTA_comparison/GENERATED_DATASETS_CrossEyed_RGB_5000_samples.pdf} &
    % \includegraphics[width=0.23\textwidth]{figures/samples_SOTA_comparison/GENERATED_DATASETS_PolyU_NIR_RGB_5000_RGB_samples.pdf} &
    % \includegraphics[width=0.23\textwidth]{figures/samples_SOTA_comparison/GENERATED_DATASETS_CrossEyed_NIR_RGB_5000_RGB_samples.pdf}\\
    % \hline 
    % & &&&\vspace{-3mm}\\ 
    % \rotatebox{90}{\hspace{6mm} \Large NIR images} & 
    % \includegraphics[width=0.23\textwidth]{figures/samples_SOTA_comparison/GENERATED_DATASETS_PolyU_NIR_5000_samples.pdf} &
    % \includegraphics[width=0.23\textwidth]{figures/samples_SOTA_comparison/GENERATED_DATASETS_CrossEyed_NIR_5000_samples.pdf} &
    % \includegraphics[width=0.23\textwidth]{figures/samples_SOTA_comparison/GENERATED_DATASETS_PolyU_NIR_RGB_5000_NIR_samples.pdf} &
    % \includegraphics[width=0.23\textwidth]{figures/samples_SOTA_comparison/GENERATED_DATASETS_CrossEyed_NIR_RGB_5000_NIR_samples.pdf}\\ 
    
    \includegraphics[width=0.23\textwidth]%,trim = 0 85mm 85mm 0,clip]
        {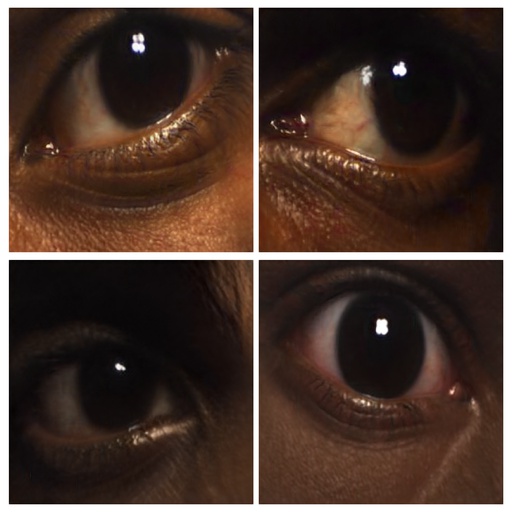} &
    \includegraphics[width=0.23\textwidth]{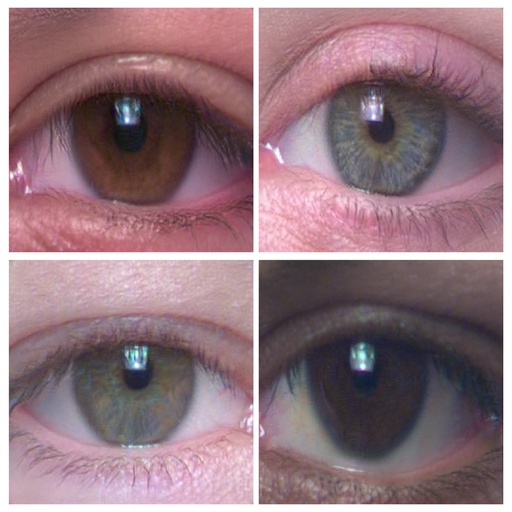} &
    \includegraphics[width=0.23\textwidth]{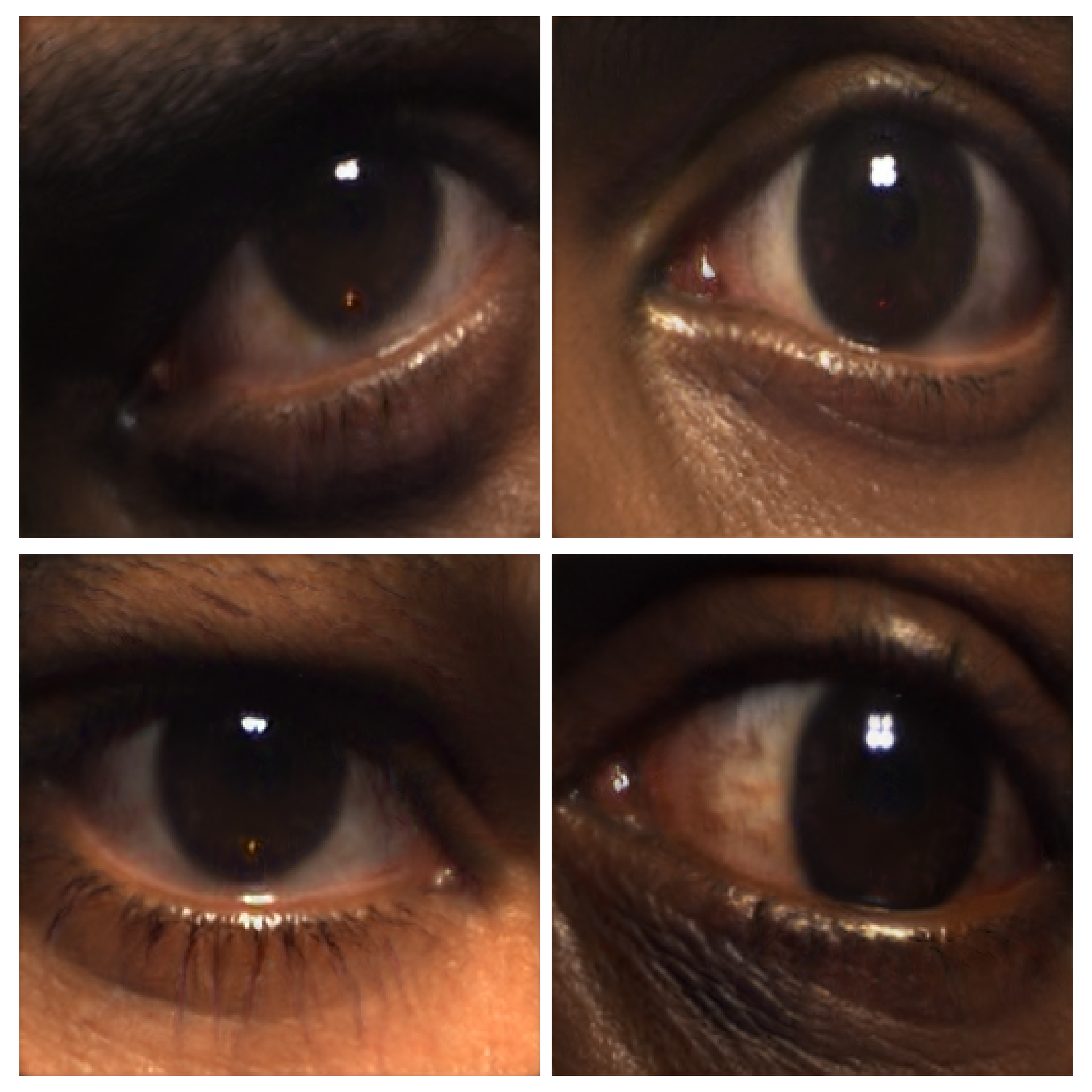} &
    \includegraphics[width=0.23\textwidth]{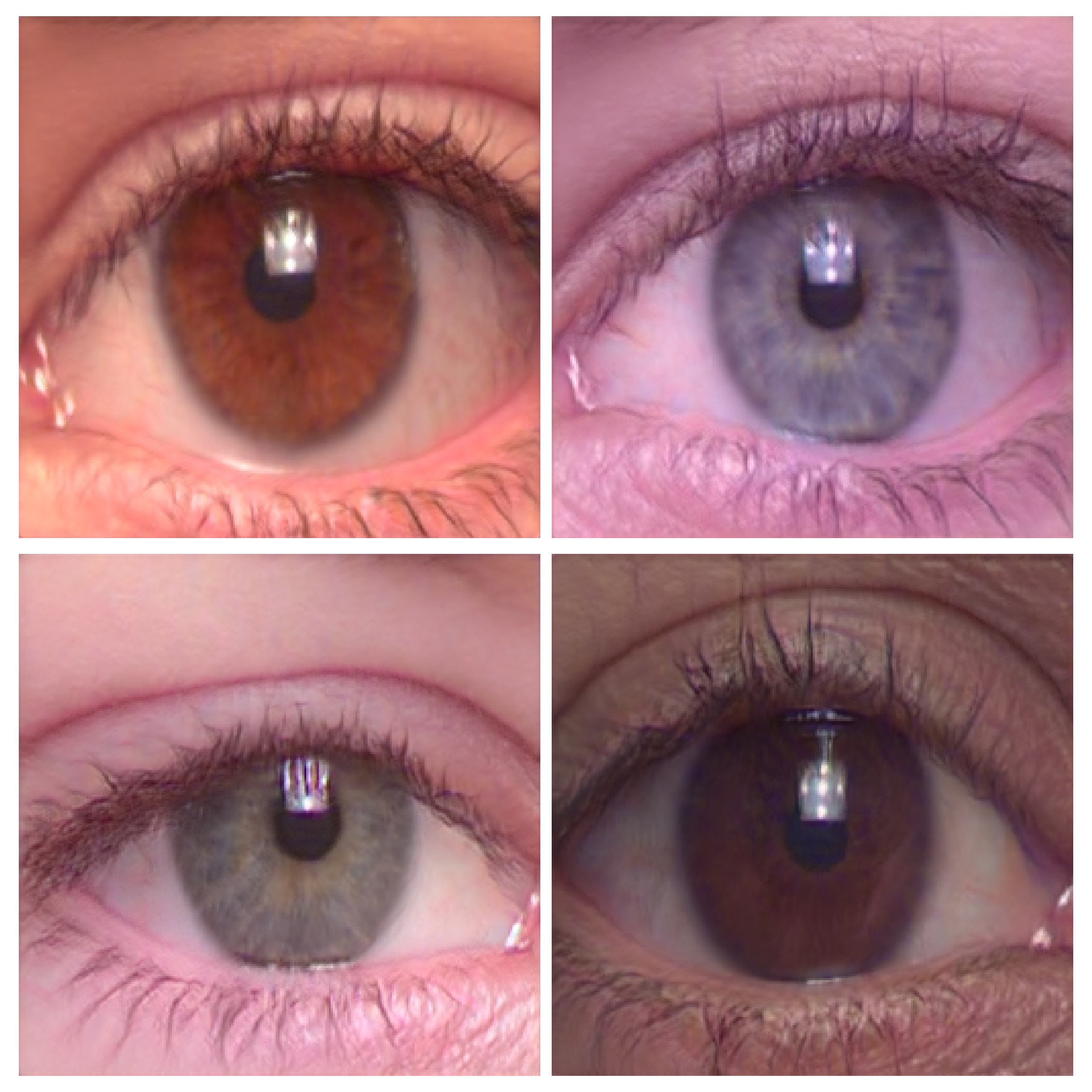}\\
    \hline 
    & &&&\vspace{-3mm}\\ 
    \rotatebox{90}{\hspace{6mm} \Large NIR images} & 
    \includegraphics[width=0.23\textwidth]{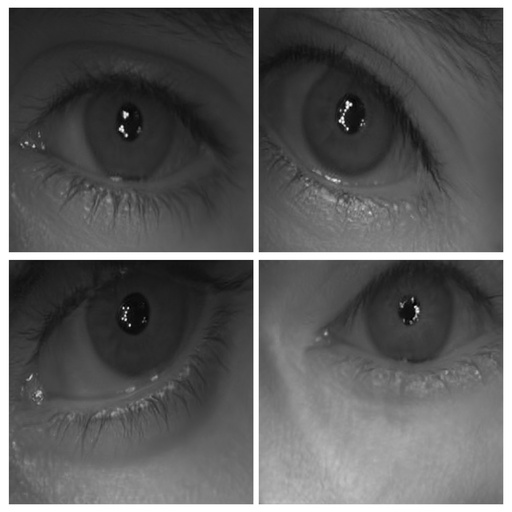} &
    \includegraphics[width=0.23\textwidth]{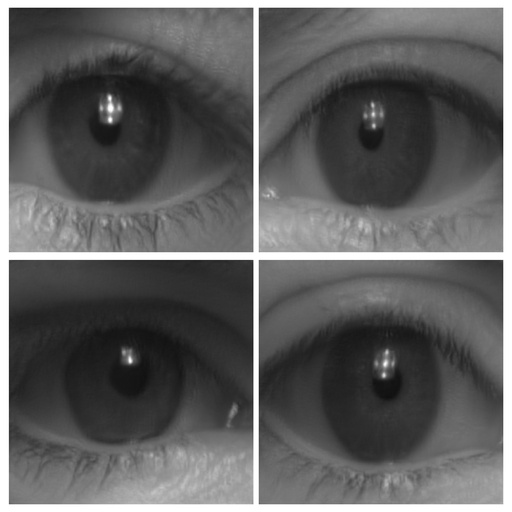} &
    \includegraphics[width=0.23\textwidth]{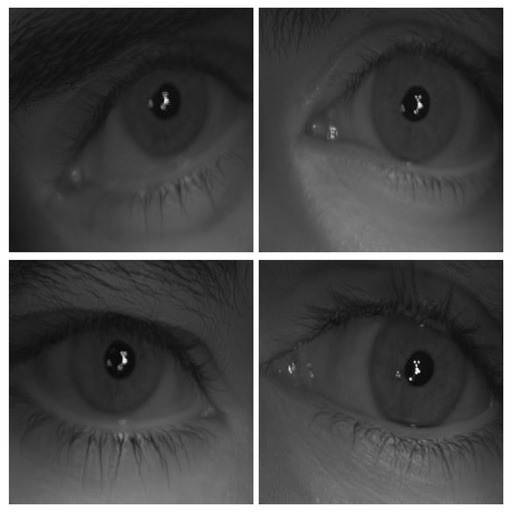} &
    \includegraphics[width=0.23\textwidth]{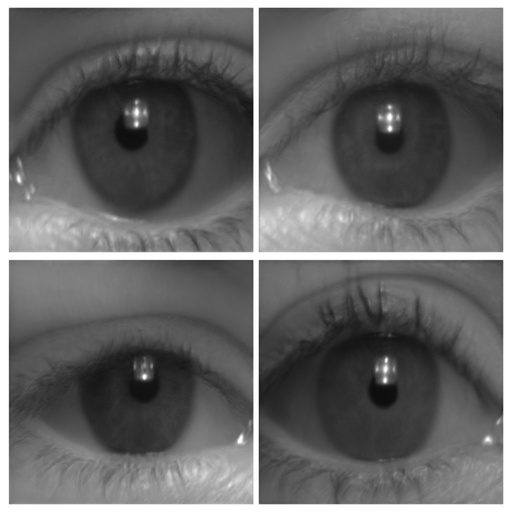}\\

    % \includegraphics[width=0.23\textwidth]%,trim = 0 85mm 85mm 0,clip]
    %     {figures/samples_SOTA_comparison/NEW/PolyU_SG2_RGB_samples_NEW.pdf} &
    % \includegraphics[width=0.23\textwidth]{figures/samples_SOTA_comparison/NEW/CrossEyed_SG2_RGB_samples_NEW.pdf} &
    % \includegraphics[width=0.23\textwidth]{figures/samples_SOTA_comparison/NEW/PolyU_DB_SG2_RGB_samples_NEW.pdf} &
    % \includegraphics[width=0.23\textwidth]{figures/samples_SOTA_comparison/NEW/CrossEyed_DB_SG2_RGB_samples_NEW.pdf}\\
    % \hline 
    % & &&&\vspace{-3mm}\\ 
    % \rotatebox{90}{\hspace{6mm} \Large NIR images} & 
    % \includegraphics[width=0.23\textwidth]{figures/samples_SOTA_comparison/NEW/PolyU_SG2_NIR_samples_NEW.pdf} &
    % \includegraphics[width=0.23\textwidth]{figures/samples_SOTA_comparison/NEW/CrossEyed_SG2_NIR_samples_NEW.pdf} &
    % \includegraphics[width=0.23\textwidth]{figures/samples_SOTA_comparison/NEW/PolyU_DB_SG2_NIR_samples_NEW.pdf} &
    % \includegraphics[width=0.23\textwidth]{figures/samples_SOTA_comparison/NEW/CrossEyed_DB_SG2_NIR_samples_NEW.pdf}\\ 
    %& \multicolumn{2}{c}{ Original images}&\multicolumn{2}{c}{ Synthesized images}\\ 
\end{tabular}}
\end{center}\vspace{-5mm}
\caption{\textbf{State-of-the-art comparison and ablation results.} The figure shows visual examples of  images synthesized with the standard unimodal StyleGAN2 and the proposed bimodal DB-StyleGAN2.
\vspace{-1mm}}
\label{tab:ablation_results}
\end{figure}

%\fi
\begin{table}[t!] %  use * at the table* for double column
% \fontsize{10}{10}\selectfont
% \scriptsize
%\begin{center}
\centering
\resizebox{0.95\columnwidth}{!}{%
\begin{tabular}{ccccc}
\toprule
    \bf{Data} & \bf{Model} & \bf{Domain}& \bf{LPIPS $\downarrow$ (T)$^\dagger$} & \bf{LPIPS $\downarrow$ (H)$^\dagger$}  \\ 
    \midrule 
    \multirow{4}{*}{\rotatebox{90}{PolyU}} 
        & StyleGAN2 & VIS &  $0.561 \pm 0.084$ & $0.561 \pm 0.083$ \\
        & StyleGAN2 & NIR  & $0.491 \pm 0.066$ & $0.492 \pm 0.068$ \\
        \cline{2-5}
        & \multirow{2}{*}{DB-StyleGAN2} & VIS & $0.559 \pm 0.082$ & $0.561 \pm 0.085$\\
        & & NIR & $0.504 \pm 0.064$ & $0.504 \pm 0.064$\\
    \midrule
    
    \multirow{4}{*}{\rotatebox{90}{CrossEyed}}
        & StyleGAN2 & VIS &  $0.476 \pm 0.064$ & $0.473 \pm 0.064$\\
        & StyleGAN2 & NIR  & $0.415 \pm 0.060$ & $0.422 \pm 0.063$ \\
         \cline{2-5}
         & \multirow{2}{*}{DB-StyleGAN2} & VIS  & {$0.453 \pm 0.068$} & {$0.456 \pm 0.063$}  \\ 
         & & NIR & {$0.391 \pm 0.063$} & {$0.392 \pm 0.058$} \\
    % \midrule 
    % \midrule 
    % \bf{Dataset} & \bf{Data} & \bf{LPIPS Real} & \bf{LPIPS generated}  \\
    % \midrule 
    % \multirow{3}{*}{Cross-Eyed}
    %     & RGB &  $0.24261 \pm 0.00289$ & $0.44954 \pm 0.00095$\\
    %     & NIR  & $0.22145 \pm 0.00299$ & $...$ \\
    %     & RGB \& NIR  & $...$ & $(0.41134 \pm 0.00125, )$ \\
\bottomrule \vspace{-2mm}\\
\multicolumn{5}{l}{(T) -- training set; (H) -- hold-out validation set; ($\downarrow$) -- lower is better} \\ 
\end{tabular}}
%\end{center}
\vspace{-1mm}
\caption{\textbf{Comparison of the computed LPIPS scores.} The scores are computed between $5000$ generated images and $(i)$ the training (T) or $(ii)$ hold-out validation set (H).\vspace{-5mm}}
\label{tab:lpips_comparison}
\end{table}

\textbf{State-of-the-Art Comparison and Ablations.} We compare the DB-StyleGAN2 models to the standard (unimodal) StyleGAN2 model from \cite{stylegan_1_karras2019style}. We note that StyleGAN2 represents a state-of-the-art model for image generation and while StyleGAN version $3$ (StyleGAN3) was also introduced recently \cite{stylegan3_karras2021alias}, it only offers superior performance (in terms of texture consistency) when generating sequences of images (or videos) but does not ensure improvements in the quality of the generated images. We train four StyleGAN2 versions for the comparisons using the NIR and VIS images from the two training datasets, i.e., PolyU and CrossEyed. The experiments presented in this section serve a dual purpose: $(i)$ they \textit{compare} the image generation capabilities of DB-StyleGAN2 to a state-of-the-art competitor, and $(ii)$~they \textit{ablate} parts of the DB-StyleGAN2 models to show the effect of bimodal image synthesis.    

\begin{figure}[t]
\centering
\resizebox{0.7\columnwidth}{!}{%
\begin{tabular}{cc}
    %\vspace{2mm}
    \huge PolyU VIS &  \huge CrossEyed VIS \\ 
    \includegraphics[width=1.0\columnwidth]{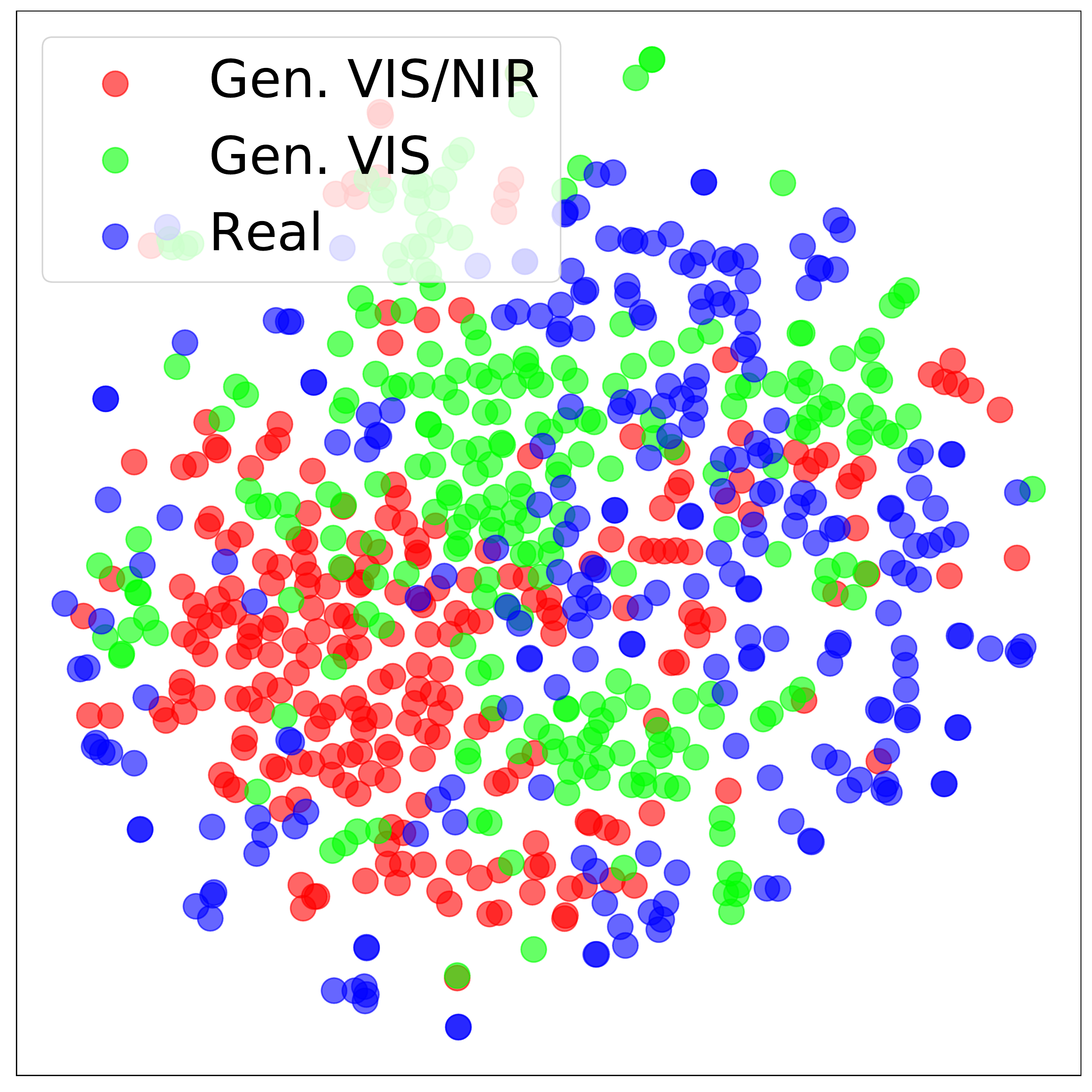}  &
    \includegraphics[width=1.0\columnwidth]{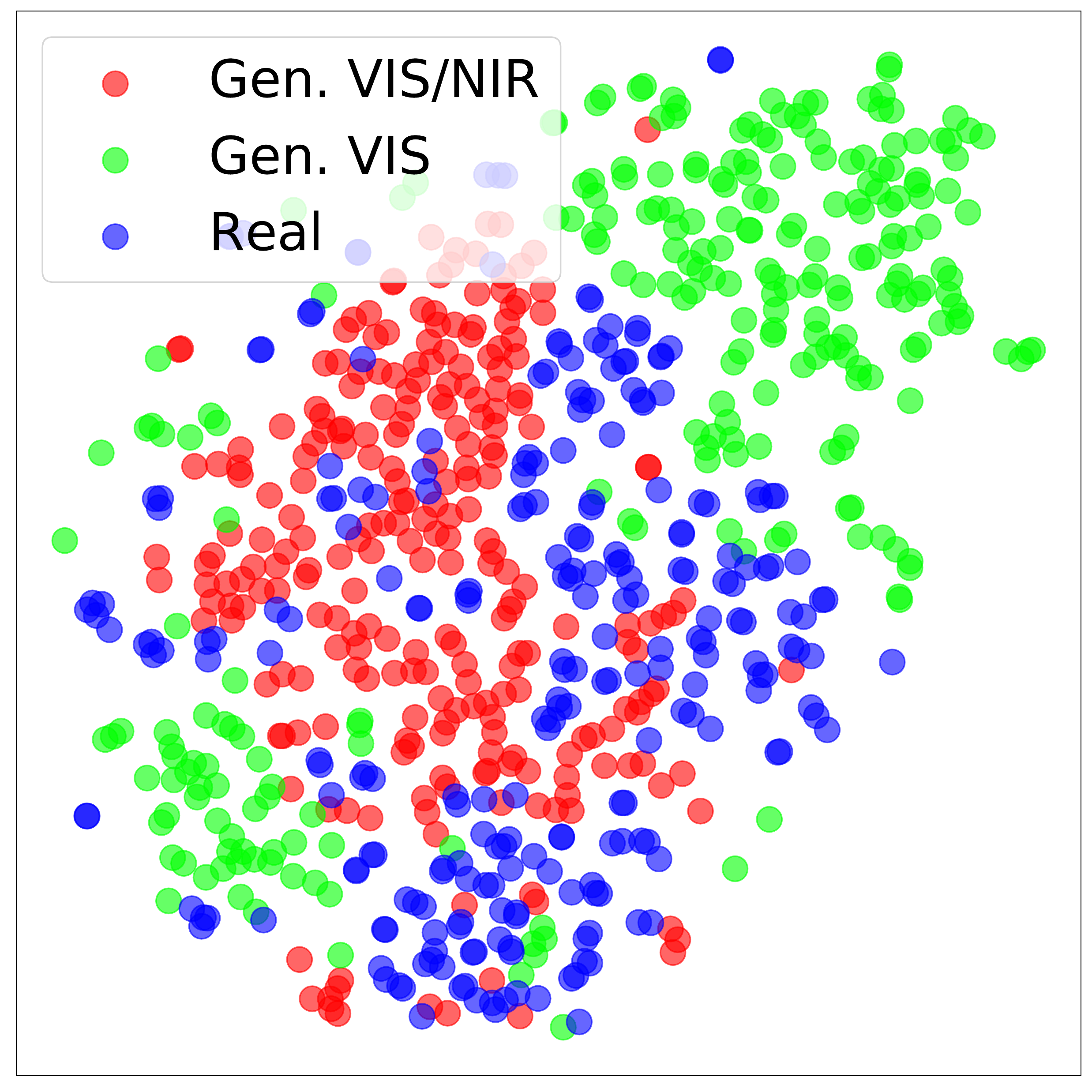} 
    \\
    \includegraphics[width=1.0\columnwidth]{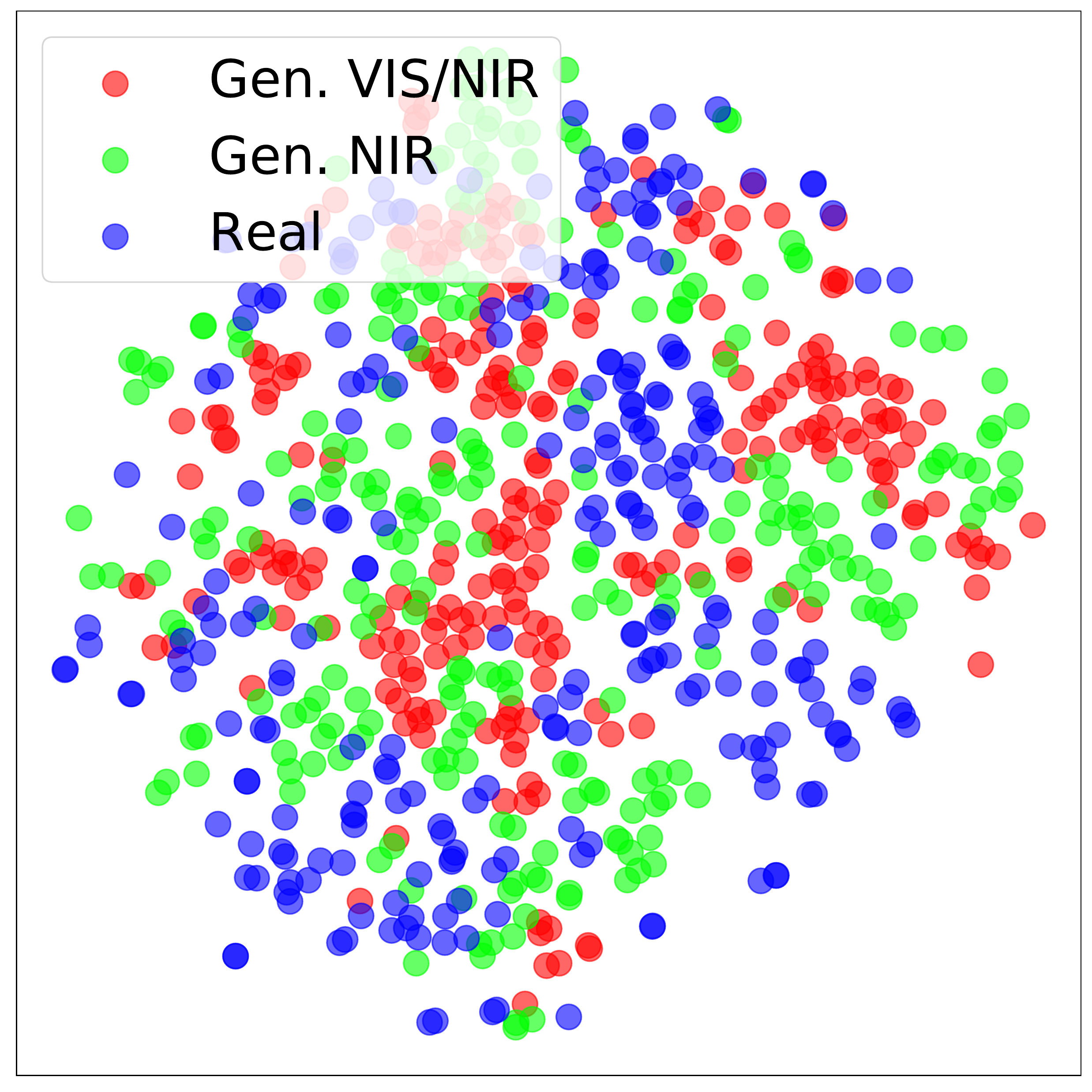} &
    \includegraphics[width=1.0\columnwidth]{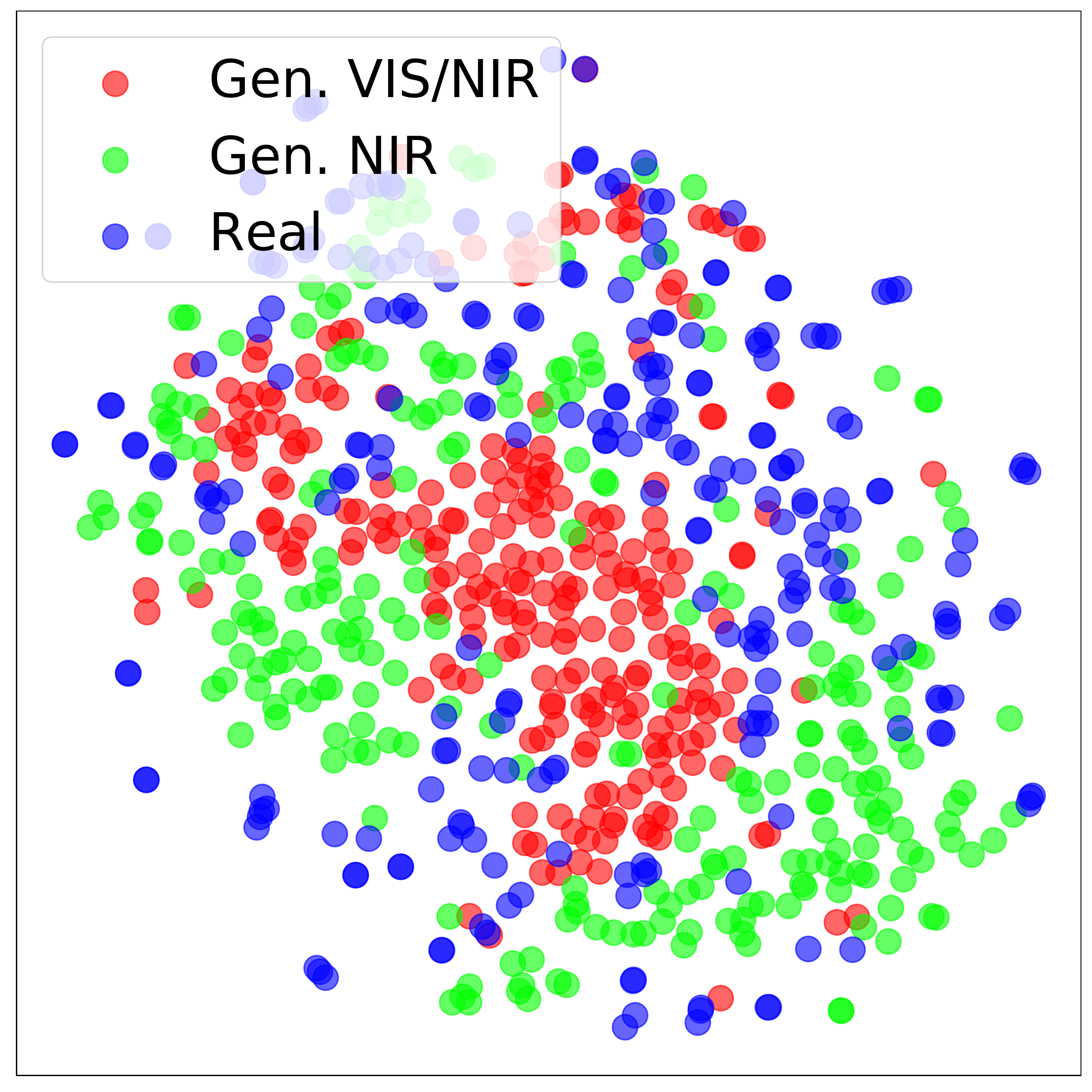}\vspace{1mm}
    \\ 
    \huge PolyU NIR &  \huge CrossEyed NIR \\
    %\vspace{2mm}  
\end{tabular}\vspace{-4mm}}
    % \vspace{-1mm}
    \caption{\textbf{$t$-SNE plots (in 2D) generated for the differently synthesized images}. Both types of models (unimodal and bimodal) produce synthetic data that corresponds well to the training data distribution. \vspace{-5mm}%{\color{red}{}} Interesting observation: when comparing images generated based on RGB versus RGB and NIR images (in the CrossEyed set).. the RGB and NIR images are closer to the real distribution %The left image in each pair represents the synthesized image and the right an actual image from the training data. 
    %Note that the models learn to generate novel data instances that share important semantic characteristics with the training images. The Mean Squared Error (MSE) is also reported for each pair.
    }
    \label{fig:tSNE_plots}
\end{figure}

Figure~\ref{tab:ablation_results} shows a visual comparison between the four unimodal StyleGAN2 models and the proposed DB-StyleGAN2. % that synthesizes bimodal image samples through a single generation step.
Note that all models generate images of comparable visual quality for both datasets in the VIS and NIR domains. However, the DB-StyleGAN2 models are able to synthesize the bimodal images through a single generation step, whereas separate models need to be trained for the off-the-shelf StyleGAN2 generators. Because the generation process is based on latent space sampling, it is also challenging to produce matching samples in both domains using the unimodal models, whereas this is handled seamlessly in DB-StyleGAN2 through the dual-branch design. In Table \ref{tab:lpips_comparison} we show a comparison of the Learned Perceptual Image Patch Similarities (LPIPS) \cite{zhang2018unreasonable} between $5000$ randomly generated images and the training (T) and hold-out validation (H) data from each dataset. As can be seen, on PolyU all models perform similarly (within the standard deviations), whereas our bimodal design has a slight advantage on CrossEyed, suggesting that the generated images are somewhat closer to the real data on average. 
%{\color{blue}{ - komentar vizualnih rezultatov, ko bodo noter. Glede na FID skore in t-SNE plote, se bi lahko pri unimodalnih modelih pojavljali kakšni outlyerji med generiranimi slikami, ki jih bi bilo fino pokazati. S tem bi lahko naredili en point in trdili, da so naši rezultati bližje pravim.}}

To get further insight into the synthesis capabilities of DB-StyleGAN2, we use $t$-distributed Stochastic Neighbor Embedding ($t$-SNE) \cite{van2008visualizing} and visualize the distribution of features extracted from different types of images in Figure~\ref{fig:tSNE_plots}. 
For this purpose, we select a ResNet-$101$ model pretrained on ImageNet (from PyTorch) as a feature extractor %\footnote{Available at: {\scriptsize \url{https://gist.github.com/flyyufelix}}} 
and use the $2048$-dimensional output of the penultimate model layer as the feature representation of the ocular images \cite{he2016deep}. 
We generate $250$ test images for the analysis by randomly sampling the latent space of the two DB-StyleGAN2  and all four unimodal StyleGAN2 models. As can be seen, % from the presented results, 
the distributions corresponding to the generated images overlap reasonably well with the distributions of the original images (marked \textit{Real}) for both types of models. However, %we observe that 
 in certain cases the unimodal models generate less overlap with the training-data distribution than the bimodal models -- see results for CrossEyed VIS for example. %it is worth noting that the bimodal synthesis procedure appears to result in data samples that correspond better to the true data distribution, as evidenced by the larger overlap. To quantify this observation, we report the Kullback–Leibler (KL) divergence between the distributions in the 2D t-SNE data in Table XX. Note that the images generated by the DB-StyleGAN2 indeed result in lower KL divergence scores than the images generated by the unimodal StyleGAN2 models.   

\begin{table}[t!] 
\begin{center}
\resizebox{0.99\columnwidth}{!}{%
\begin{tabular}{lcccc}
\toprule
    \multirow{2}{*}{\bf{Model}} & \multicolumn{2}{c}{\bf{Training time  [hours]$^\dagger$}} && \multirow{2}{*}{\bf{Run-time [ms]}} \\ \cline{2-3} 
    & PolyU & CrossEyed &&   \\
    \midrule
    % DB-StyleGAN2 (bimodal) & $147.88 \pm 4.78$ & $151.3 \pm 4.48$  & &  $13.994 \pm 0.068$ \\
    % StyleGAN2 (unimodal) & $80.64 \pm 2.59$ &   $ 82.13 \pm 2.85 $  && $11.232 \pm 0.071$  \\% $43.23 \pm 1.76$  && $x \pm x$  \\
    
    %DB-StyleGAN2 (bimodal) & $40.17 \pm 0.07$ & $40.13 \pm 0.07$  & &  $13.994 \pm 0.068$ \\
    %StyleGAN2 (unimodal) & $22.87 \pm 0.03$ &   $22.97 \pm 0.05$  && $11.232 \pm 0.071$  \\
    DB-StyleGAN2  & $\sim 20h$  & $\sim 24h$  && $13.994 \pm 0.068$ \\
    StyleGAN2 & $\sim 18h$ & $\sim 18h$ && $11.232 \pm 0.071$ \\
   % \midrule
%    \multirow{2}{*}{\bf{Model}} & \multicolumn{2}{c}{\bf{Total kimgs (time in hours)$^\dagger$} (+ evaluation)} &&  \\ \cline{2-3} 
 %   & PolyU & CrossedEye &&   \\
  %  \midrule
  %  DB-StyleGAN2 (bimodal) & 1600 (20h)  & 2080 (24h)  \\
   % StyleGAN2 (unimodal) & 2500 (18h) & 2500 (18h) \\
\bottomrule
\multicolumn{5}{l}{$^\dagger$Approximate estimate\vspace{-4mm}}
\end{tabular}
}
\end{center}
% \vspace{-2mm}
\caption{\textbf{Training and run-time requirements.} The bimodal DB-StyleGAN2 model takes longer to train than the unimodal StyleGAN2, but is able to match the run-time performance of its unimodal counterpart. %\vspace{-1mm}%{\color{red}Note: while DB-StyleGAN2 takes longer it also has to process twice as many images and it also converges in significantly fewer iterations - kimgs (for example 1600 instead of 2500)}
}
\label{tab:time}
\end{table}

%%%%%%%%%%%%%%%%%%%%%%%%%%%%%%
\begin{table*}[t!] %  use * at the table* for double column
\resizebox{\textwidth}{!}{%
\begin{tabular}{lc|ccc|ccc|ccc}
\toprule
    \multirow{3}{*}{\bf{Data generated by}}& \multirow{3}{*}{\bf{Seg. Model}} &  \multicolumn{9}{c}{\bf{Trained on CrossEyed}} \\ \cline{3-11}
    
    & & \multicolumn{3}{c}{\bf{SMD}$^{\dagger,\ddagger}$} & \multicolumn{3}{c}{\bf{MOBIUS$^{\dagger}$}} & \multicolumn{3}{c}{\bf{SBVPI$^{\dagger}$}}  \\ \cline{3-11}
    %\midrule
     &  & \bf{IoU $\uparrow$} & \bf{$F_{1}$ $\uparrow$} & \bf{Pixel error $\downarrow$ [\%]} & \bf{IoU $\uparrow$} & \bf{$F_{1}$ $\uparrow$} & \bf{Pixel error $\downarrow$ [\%]} & \bf{IoU $\uparrow$} & \bf{$F_{1}$ $\uparrow$} & \bf{Pixel error $\downarrow$ [\%]} \\ 
    \midrule
    DatasetGAN \cite{zhang2021datasetgan} & \multirow{2}{*}{DeepLab-V3}  &  
        $0.601 \pm 0.097$  &  $0.703 \pm 0.101$  &  $0.123 \pm 0.046$ & 
        $0.554 \pm 0.185$  &  $0.652 \pm 0.181$  &  $0.148 \pm 0.136$  & 
        $0.832 \pm 0.052$  &  $0.902 \pm 0.038$  &  $0.046 \pm 0.019$ \\
    
   BiOcularGAN (ours) &  &  
        \bm{$0.658 \pm 0.084$}  &  \bm{$0.756 \pm 0.085$}  &  \bm{$0.082 \pm 0.033$}  & 
        \bm{$0.587 \pm 0.117$}  &  \bm{$0.683 \pm 0.120$}  &  \bm{$0.095 \pm 0.041$} &
        \bm{$0.834 \pm 0.049$} &  \bm{$0.902 \pm 0.038$}  &  \bm{$0.037 \pm 0.012$} \\
        
    \midrule
    DatasetGAN\cite{zhang2021datasetgan} & \multirow{2}{*}{U-Net}  &
        $0.655 \pm 0.083$  &  $0.754 \pm 0.085$  &  $0.085 \pm 0.032$ & 
        $0.541 \pm 0.141$  &  $0.635 \pm 0.155$  &  $0.098 \pm 0.049$ &
        $0.809 \pm 0.052$  &  $0.885 \pm 0.041$  &  $0.045 \pm 0.012$ \\
    BiOcularGAN (ours) &  &
        \bm{$0.722 \pm 0.070$}  &  \bm{$0.812 \pm 0.066$}  &  \bm{$0.048 \pm 0.021$} &
        \bm{$0.551 \pm 0.133$}  &  \bm{$0.638 \pm 0.142$}  &  \bm{$0.086 \pm 0.047$} &
        \bm{$0.839 \pm 0.045$}  &  \bm{$0.906 \pm 0.035$}  &  \bm{$0.035 \pm 0.011$} \\
    \midrule
    \midrule
%\iffalse    
    \multirow{3}{*}{\bf{Data generated by}} &  \multirow{3}{*}{\bf{Seg. Model}}&  \multicolumn{9}{c}{\bf{Trained on PolyU}} \\\cline{3-11}
    & & \multicolumn{3}{c}{\bf{SMD$^{\dagger}$}} & \multicolumn{3}{c}{\bf{MOBIUS$^{\dagger,\ddagger}$}} & \multicolumn{3}{c}{\bf{SBVPI$^{\dagger,\ddagger}$}}  \\\cline{3-11}
    %\midrule
    & & \bf{IoU $\uparrow$} & \bf{$F_{1}$ $\uparrow$} & \bf{Pixel error $\downarrow$ [\%]} & \bf{IoU $\uparrow$} & \bf{$F_{1}$ $\uparrow$} & \bf{Pixel error $\downarrow$ [\%]} & \bf{IoU $\uparrow$} & \bf{$F_{1}$ $\uparrow$} & \bf{Pixel error $\downarrow$ [\%]} \\ 
    \midrule
    DatasetGAN \cite{zhang2021datasetgan} & \multirow{2}{*}{DeepLab-V3}  &  
        $0.728 \pm 0.084$  &  $0.818 \pm 0.077$  &  $0.058 \pm 0.027$ & 
        $0.607 \pm 0.154$  &  $0.701 \pm 0.151$  &  $0.103 \pm 0.122$ &
        $0.808 \pm 0.062$  &  $0.884 \pm 0.047$  &  $0.047 \pm 0.019$ \\
    %& NIR  & $..$ & $...$ \\
   BiOcularGAN (ours) &  &  
        \bm{$0.787 \pm 0.056$}  &  \bm{$0.867 \pm 0.045$}  &  \bm{$0.036 \pm 0.016$} & 
       \bm{ $0.638 \pm 0.167$}  &  \bm{$0.725 \pm 0.175$}  &  \bm{$0.065 \pm 0.048$} &
        \bm{$0.834 \pm 0.046$}  &  \bm{$0.903 \pm 0.037$}  &  \bm{$0.035 \pm 0.009$} \\
        
    \midrule
   DatasetGAN \cite{zhang2021datasetgan}  & \multirow{2}{*}{U-Net} &
        $0.679 \pm 0.089$  &  $0.771 \pm 0.093$  &  $0.064 \pm 0.028$ &
        $0.519 \pm 0.137$  &  $0.605 \pm 0.150$  &  $0.092 \pm 0.053$ & 
        $0.757 \pm 0.058$  &  $0.848 \pm 0.047$  &  $0.064 \pm 0.024$ \\
   BiOcularGAN (ours) &  &
        \bm{$0.772 \pm 0.081$}  &  \bm{$0.853 \pm 0.070$}  &  \bm{$0.041 \pm 0.025$} &
        \bm{$0.584 \pm 0.173$}  &  \bm{$0.674 \pm 0.187$}  &  \bm{$0.082 \pm 0.051$} & 
        \bm{$0.818 \pm 0.052$}  &  \bm{$0.891 \pm 0.041$}  &  \bm{$0.040 \pm 0.015$} \\
\bottomrule
\multicolumn{11}{l}{$^\dagger$Cross-dataset experiments; $^\ddagger$Cross-ethnicity experiments; $^\uparrow$ Higher is better; $^\downarrow$ Lower is better }
\end{tabular}
}
\vspace{-2mm}
\caption{\textbf{Cross-dataset segmentation performance comparison of models trained on artificially generated datasets.\label{tab:state_of_art_comparison}} 
The segmentation models trained on $5000$ images generated by BiOcularGAN outperform the ones trained on $5000$ images generated by DatasetGAN across all datasets and performance measures ($IoU$ and $F_1$ scores along with pixel errors).\vspace{-4mm} %The arrows $\uparrow$ and $\downarrow$ indicate whether higher or lower values correspond to better performance. %{\color{blue} Komentar: kar označujemo kot DatasetGAN je v bistvu DatasetGAN nadgrajen na StyleGAN2 (čeprav ni tok važno)}
}
% .. new idea: show results on validation, SMD (sclera), SBVPI (sclera or multi), MOBIUS (multi) ... The main problem is probably different gaze directions of testing data ... Training images generated with StyleGAN (trained on PolyU dataset) and masks generated with DatasetGAN.. F1 score measures close to average performance and the IoU score measures closer to the worst case performance.. Total error represents the \% of misclassified pixels in the image. .. Note for SBVPI and SMD (where only sclera annotations exist), we simply convert other predictions (iris, pupil) to background and compute metrics over the two classes ... OBSERVATION: RGB \& NIR outperforms on most datasets (except CrossEyed on entire SBVPI and MOBIUS) }
\label{tab:state_of_the_art_comparison}
\end{table*}

\iffalse
\begin{table}[t!] %  use * at the table* for double column
% \fontsize{10}{10}\selectfont
% \scriptsize
\begin{center}
\resizebox{0.99\columnwidth}{!}{%
\begin{tabular}{lccccc}
\toprule
    \multirow{2}{*}{\bf{Model}} & \multicolumn{2}{c}{\bf{PolyU}} && \multicolumn{2}{c}{\bf{CrossEyed}}  \\ \cline{2-3} \cline{5-6}
    & RGB & NIR && RGB & NIR \\
    \midrule
    DB-StyleGAN2 (bimodal) & 3.352 & 3.131  &&  \bf{3.696} & 4.120 \\
    StyleGAN2 (unimodal) & \bf{3.311} & \bf{2.942}    &&  4.402  & \bf{3.865} \\
\bottomrule
\end{tabular}
}
\end{center}
\caption{\textbf{Analysis of 2D t-SNE distributions.} The table shows KL divergence scores computed from the data distributions from Figure \ref{fig:tSNE_plots} between real images and the synthetic images generated either with StyleGAN2 or DB-StyleGAN2.}
\label{tab:KL divergence}
\end{table}
\fi

\textbf{Real-world Time Requirements.} In Table \ref{tab:time} we summarize the training and run-time requirements of the  DB-StyleGAN2 model on both datasets in comparison to the unimodal StyleGAN2 versions using our experimental hardware. Here, run-time is estimated over $1000$ random samples. Note that training of the bimodal models takes longer, as twice the amount of data needs to processed compared to the unimodal models. However, because of the significantly better convergence, the training time is increased only around 11\% with the PolyU data and by 33\% on CrossEyed. At run-time, we observe comparable results, around $14$ms for the bimodal and $11$ms for the unimodal models. However, we note again that for generating ocular images in the NIR and VIS domain, the unimodal StyleGAN2 models need to be run twice.

% \iffalse
% \begin{table}[t!] %  use * at the table* for double column
% % \fontsize{10}{10}\selectfont
% % \scriptsize
% \begin{center}
% \resizebox{0.99\columnwidth}{!}{%
% \begin{tabular}{cccc}
% \toprule
%     \bf{Dataset} & \bf{Data} & \bf{FID T.} & \bf{FID V.}  \\ 
%     \midrule
%     \multirow{3}{*}{PolyU CS} %& RGB &  $18.839$ & $...$ \\
%     % & NIR  & $19.684$  & $...$ \\
%     % & RGB \& NIR  & $(23.921, 19.537)$ & $...$ \\
%     & RGB &  $28.185$ & $43.943$ \\
%     & NIR  & $26.061$  & $39.366$ \\
%     & RGB \& NIR & $(29.798, 33.863) $ & $(45.276, 46.701)$ \\
%     \midrule
%     \multirow{3}{*}{Cross-Eyed} %& RGB &  $34.227$ & $...$\\
%     % & NIR  & $34.738$  & $...$ \\
%     % & RGB \& NIR  & $(35.351, 21.478)$ & $..$ \\
%     & RGB &  $36.160$ & $71.217$\\
%     & NIR  & $33.277$  & $60.712$ \\
%     & RGB \& NIR  & $(38.258, 34.129)$ & $(71.362,  58.883)$ \\
    
% \bottomrule
% \end{tabular}}
% \end{center}
% \caption{{\color{red}{TODO}} image quality results of models trained on the PolyU and Cross-Eyed datasets. FID computed between the generated images and the training, as well as validation set respectively.}
% \label{tab:FID_comparison}
% \end{table}
% \fi

% NIR without RGB mean normalization
% Entire CrossEyed without Resize!
%\iffalse
\begin{figure}[t!] 
\begin{center}
\resizebox{0.8\columnwidth}{!}{%
\begin{tabular}{cc|c|c}
%\toprule
    %\midrule
    & \Large SMD & \Large MOBIUS &  \Large SBVPI \\ 
    \rotatebox{90}{\hspace{6mm} \Large DatasetGAN} & 
    \includegraphics[width=0.23\textwidth]%,trim = 0 85mm 85mm 0,clip]
    {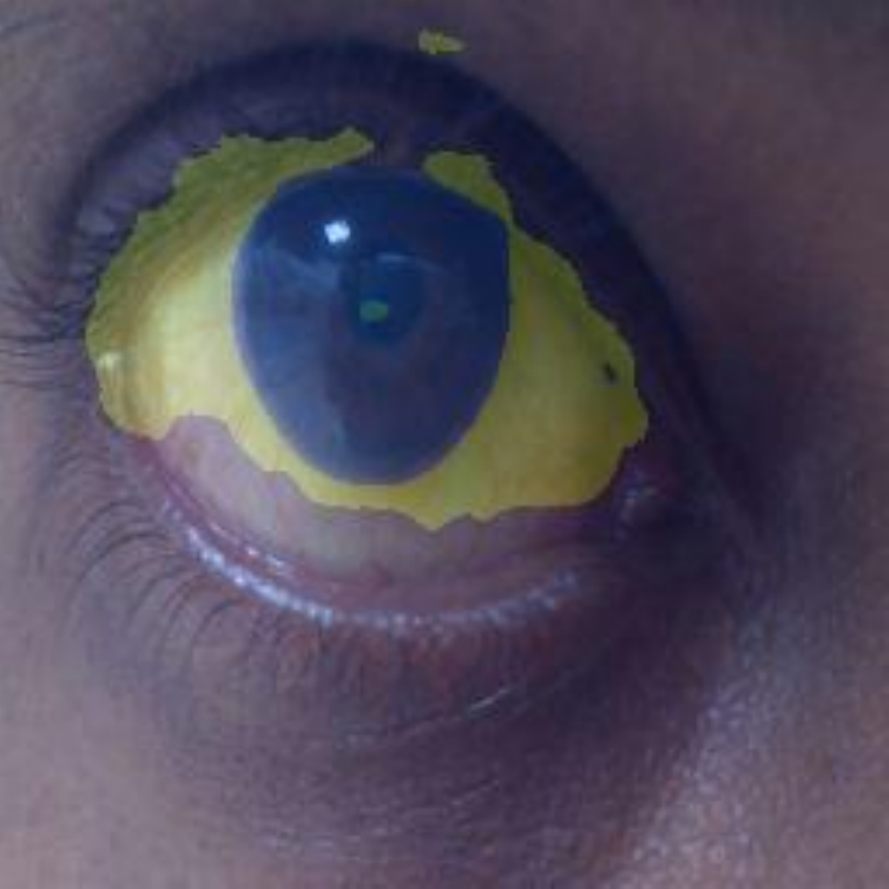} &
    \includegraphics[width=0.23\textwidth]{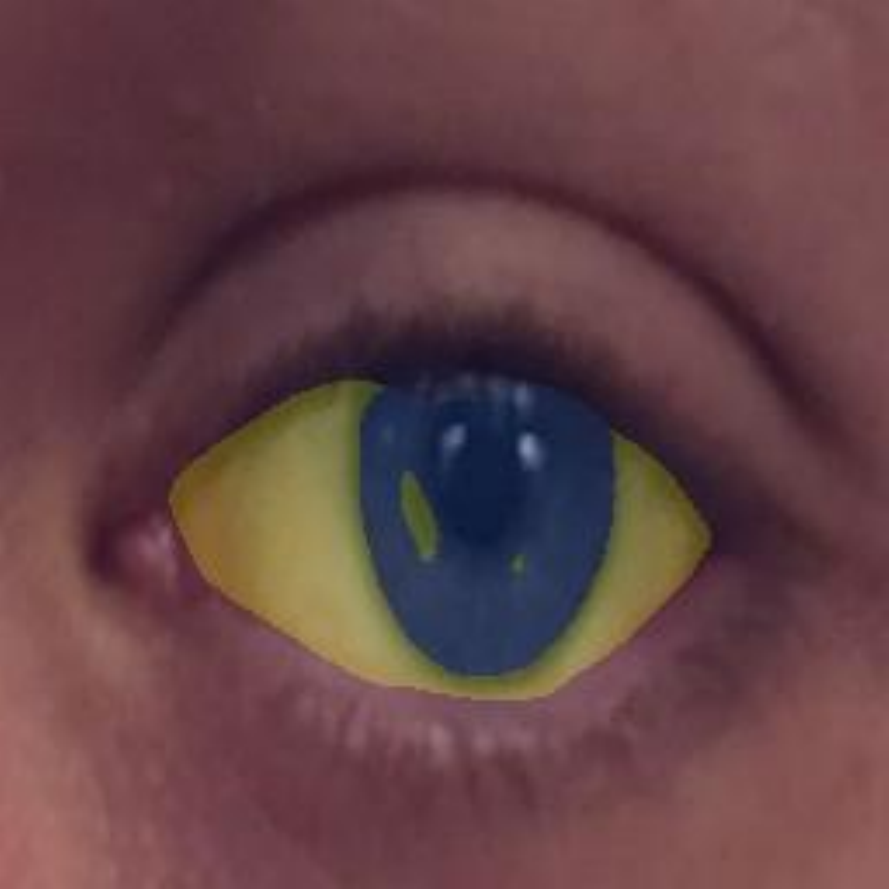} &
    \includegraphics[width=0.23\textwidth]{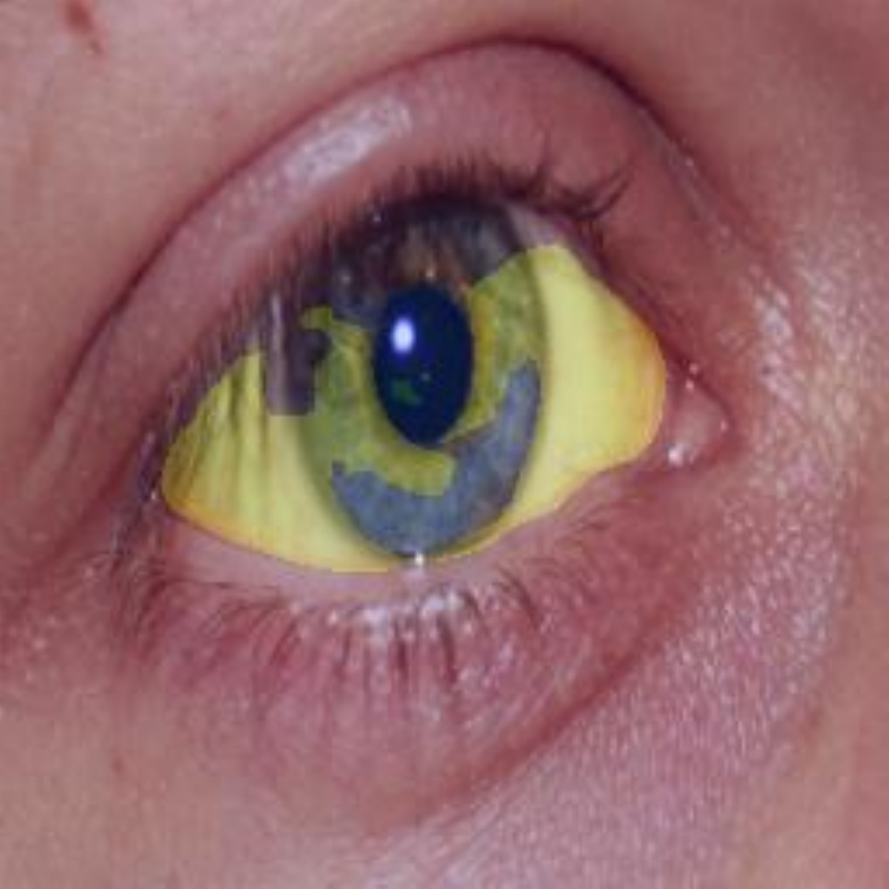} 
    \\
    \hline 
    & &&\vspace{-3mm}\\
    \rotatebox{90}{\hspace{3mm} \Large BiOcularGAN} & 
    \includegraphics[width=0.23\textwidth]%,trim = 0 85mm 85mm 0,clip]
    {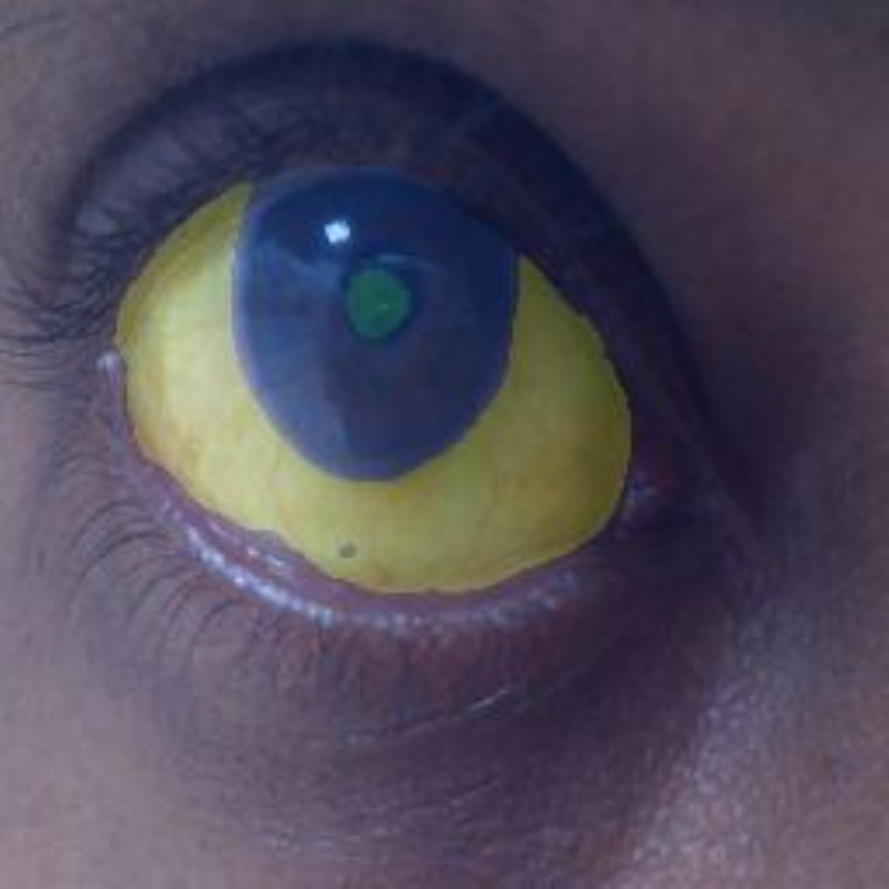} &
    \includegraphics[width=0.23\textwidth]{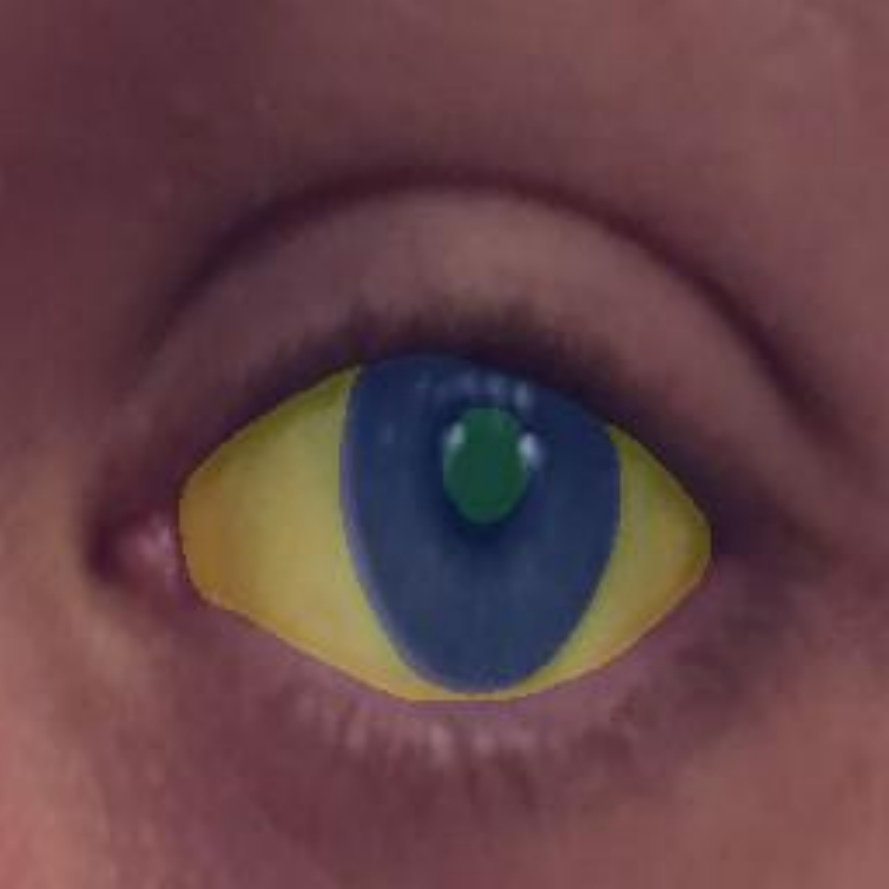} &
    \includegraphics[width=0.23\textwidth]{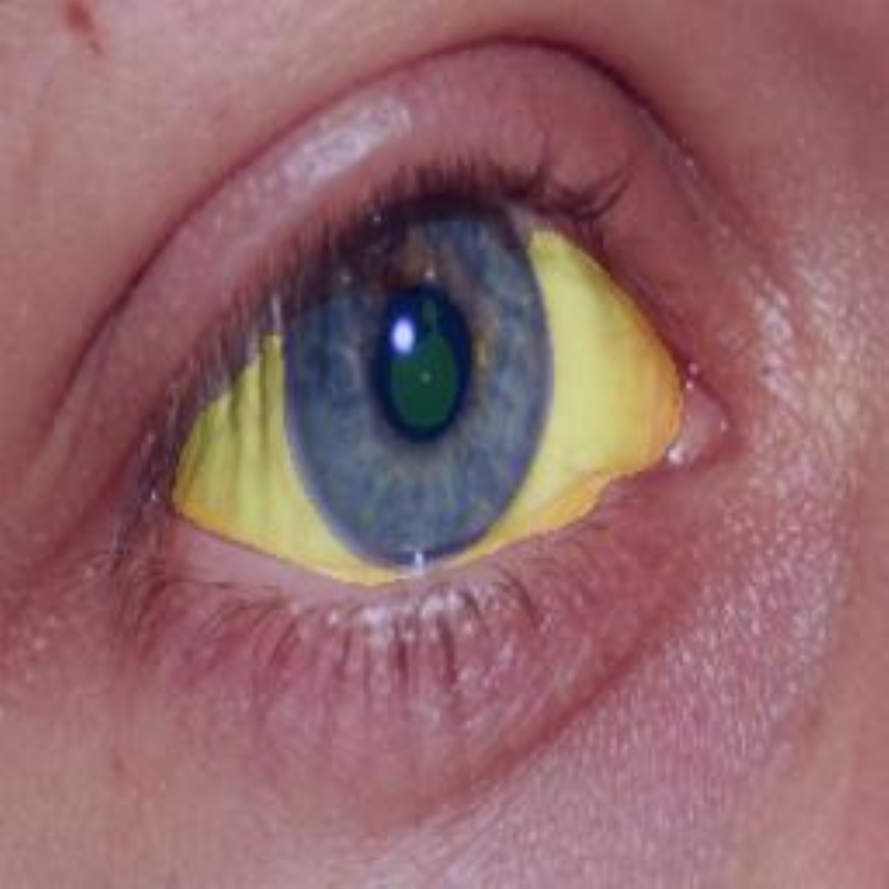} 
    %\\
    %\hline 
    %& &&\vspace{-3mm}\\
    %\rotatebox{90}{\hspace{10mm} \large Ground truth} & 
    %\includegraphics[width=0.23\textwidth]{figures/segmentation_mask_comparison/example_GT_merge_PolyU_NIR_RGB__SMD_256_64.pdf} &
    %\includegraphics[width=0.23\textwidth]{figures/segmentation_mask_comparison/example_GT_merge_PolyU_NIR_RGB__MOBIUS_256_center_3.pdf} &
    %\includegraphics[width=0.23\textwidth]{figures/segmentation_mask_comparison/example_GT_merge_PolyU_NIR_RGB__SBVPI_sclera_center_56.pdf}
    %\\ 
    
    %& \multicolumn{2}{c}{ Original images}&\multicolumn{2}{c}{ Synthesized images}\\ 
\end{tabular}}
\end{center}\vspace{-4mm}
\caption{\textbf{Sample segmentation results.} The results were generated with two U-Net models, trained on artificial data generated by the DatasetGAN and BiOcularGAN frameworks, learned with the DB-StyleGAN2-P model. \vspace{-4mm}%Note that BiOcularGAN better captures the semantic content of the images and therefore 
%leads to %higher quality training data and consequently 
%better segmentation results. % {\color{blue}  Observation: U-Net on BiOcularGAN has more rounded and detailed masks... U-Net on DatasetGAN has issues with the blue iris in SBVPI and the pupil in MOBIUS. Probable reason: the GAN was trained on PolyU (only has brown irises and its pupils are often hard to distinguish). .... Sem imel razlago o latent vektorjih in informaciji, ki jo imajo na podlagi VIS in NIR slik, vendar po pregledu training primerov ni ravno smiselna...
%Thus the latent vectors probably did not include enough information regarding this split, when trained on only VIS images. However, in NIR images these regions can be easily separated (distinguished), thus it stands to reason that the DB-StyleGAN2 could encode more semantic information into its latent vectors than the normal StyleGAN2 (since the DB version uses shared latent maps to generate VIS and NIR images). This in turn helps the ensemble MLP classifiers generate better training examples. 
%}
%\color{red}{ Mogoče moramo primerjati artificial training podatke}
}
\label{tab:semantic_mask_predictions}
\end{figure}

%%---------------------------------------------------
% Semantic masks generation
%----------------------------------------------------
\subsection{Bimodal data annotation and segmentation}

In the second set of experiments, we explore the advantages that bimodal information brings to the ground truth segmentation-mask generation process. To this end, we manually annotate $8$ ocular images generated by each of the two DB-StyleGAN models using $4$ target segmentation classes, i.e., the pupil, the iris, the sclera and the background. We use the NIR images as the basis for the manual annotation procedure (due to better contrast, distinct borders, etc.), but due to the alignment of the artificial bimodal images, these segmentation masks are also applicable to the VIS data. Using the generated annotations, we then train the mask generation procedure and synthesize a training dataset of $5000$ pairs of VIS and NIR images with corresponding reference segmentation masks (and $500$ for validation). 
Finally, we train a DeepLab-V3 \cite{florian2017rethinking} and U-Net \cite{ronneberger2015u} segmentation model using the synthetic datasets. Public implementations are used to foster reproducibility\footnote{U-Net: {\scriptsize \url{https://github.com/milesial/Pytorch-UNet}} \newline Deeplab-V3: {\scriptsize \url{https://github.com/jnkl314/DeepLabV3FineTuning}}}.
To test the performance of the trained models, we use the (frontal gaze) VIS images from SMD, MOBIUS and SBVPI. Thus, segmentation performance with VIS images is used as a proxy for the quality of the generated segmentation masks. 

\textbf{State-of-the-art Comparison.} In Table \ref{tab:state_of_art_comparison}, we report the results of the segmentation experiments in terms of the Intersection-Over-Union (IoU), $F_1$ score and overall Pixel error following established methodology \cite{rot2018deep,SSBC2020} and compare the performance ensured by the data generated by our BiOcularGAN to that produced by the unimodal DatasetGAN procedure from \cite{zhang2021datasetgan}. Here, the DatasetGAN approach is learned from the unimodal StyleGAN2 model trained on VIS images, and with $8$ manually annotated images. 

Interestingly, the segmentation models trained with the artificial dataset generated by the proposed BiOcularGAN framework clearly outperform the models trained with DatasetGAN on all three test datasets and across all three performance measures. This suggests that the joint bimodal supervision used to train the DB-StyleGAN model helps to better capture the semantic information of the images in the model layers and consequently leads to higher quality training data. This observation is further supported by the sample results in Figure \ref{tab:semantic_mask_predictions}, where we again see better segmentation performance following the use of the BiOcularGAN framework for data generation.  Here, the examples were produced with U-Net and the BiOcularGAN and DatasetGAN frameworks trained using the PolyU data.     

%{\color{blue}{ komentar, ko bodo vsi rezultati noter. Observation, PolyU might perform better on entire MOBIUS just because of larger differences in gaze direction than CrossEyed}}

%In  we show a few qualitative segmentation results.

%{\color{blue}{ (Figure \ref{tab:state_of_art_comparison_10k}) dodaj par vizualnih primerov rezultatov segmentacij z našimi in datasetsgan naučenimi modeli.}}

\begin{figure}[!ht] 
\begin{center}
\resizebox{\columnwidth}{!}{%
\begin{tabular}{ccc}
%\toprule
    %\midrule
    & \huge Training data & \huge Testing data (SMD/MOBIUS/SBVPI)   \\
   % & &&\vspace{-3mm}\\
    \rotatebox{90}{\hspace{7mm} \Large VIS images} &
    \includegraphics[width=0.2\textwidth]%,trim = 0 85mm 85mm 0,clip]
    {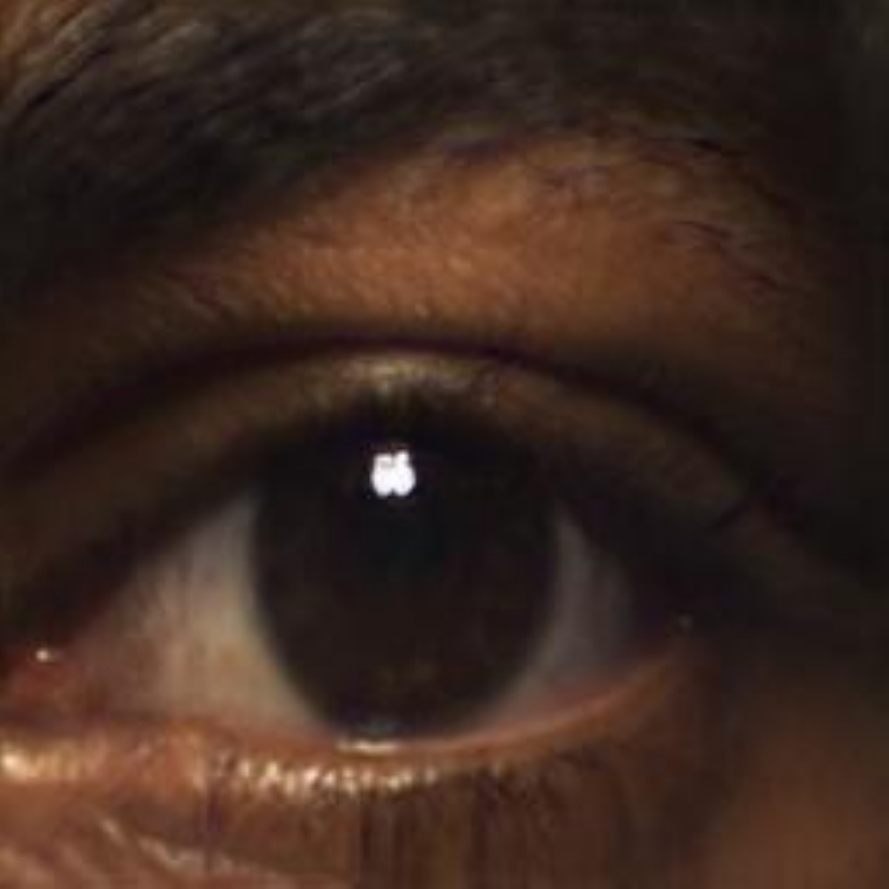} 
    \includegraphics[width=0.2\textwidth]{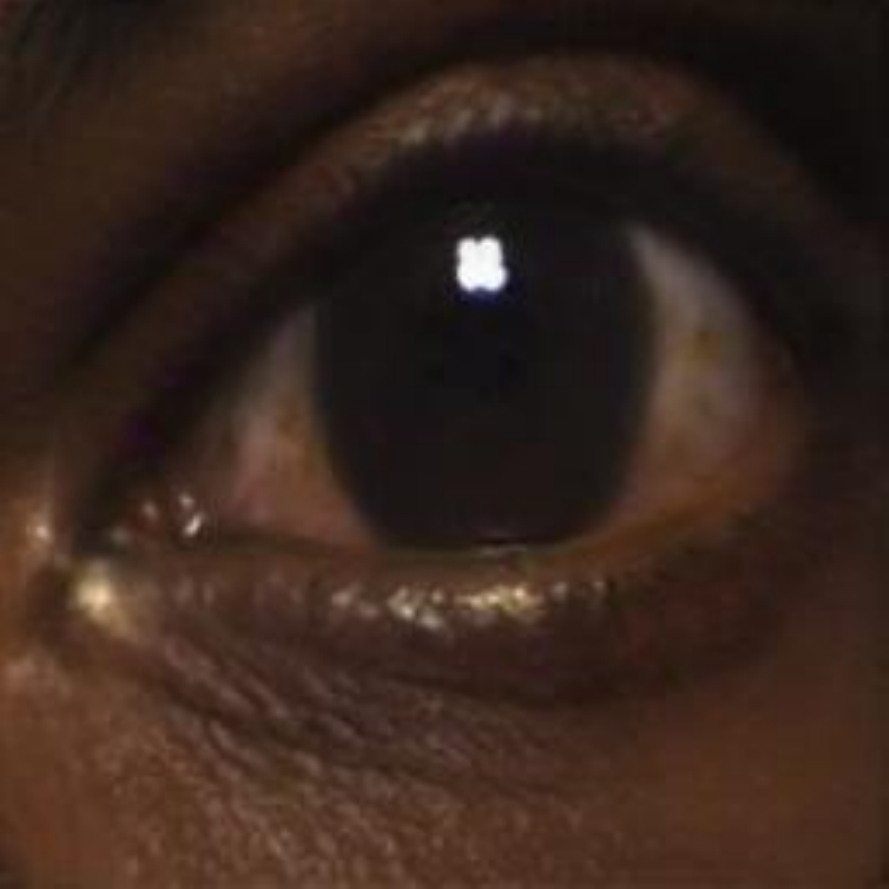}
    \includegraphics[width=0.2\textwidth]{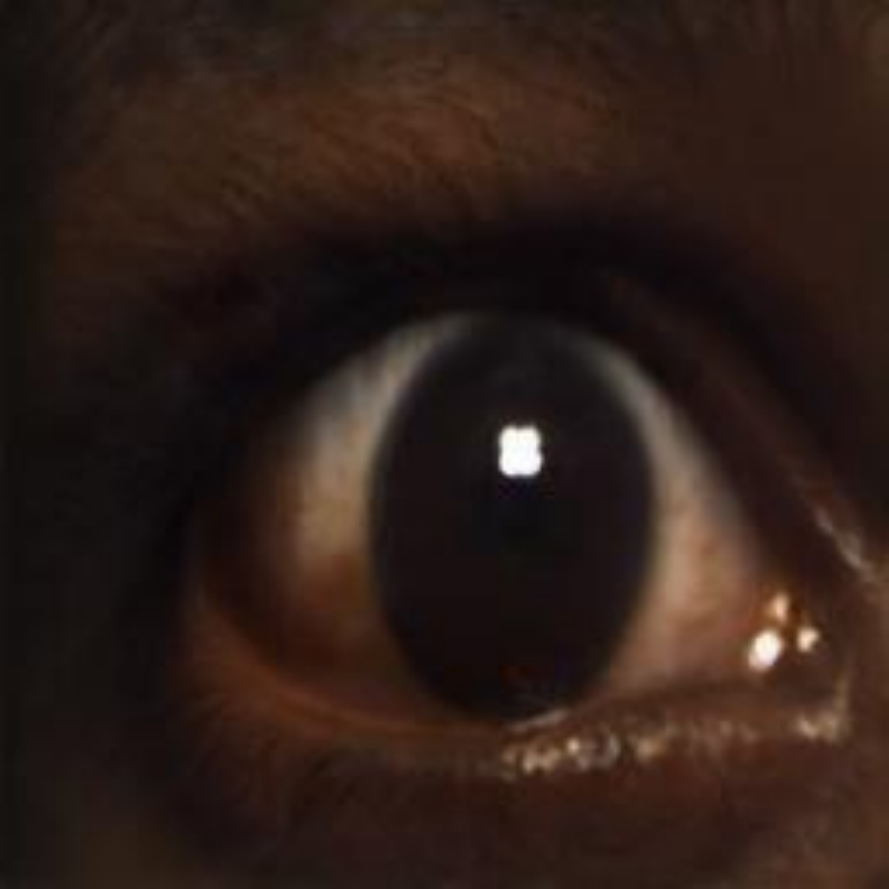} &
    
    \includegraphics[width=0.2\textwidth]{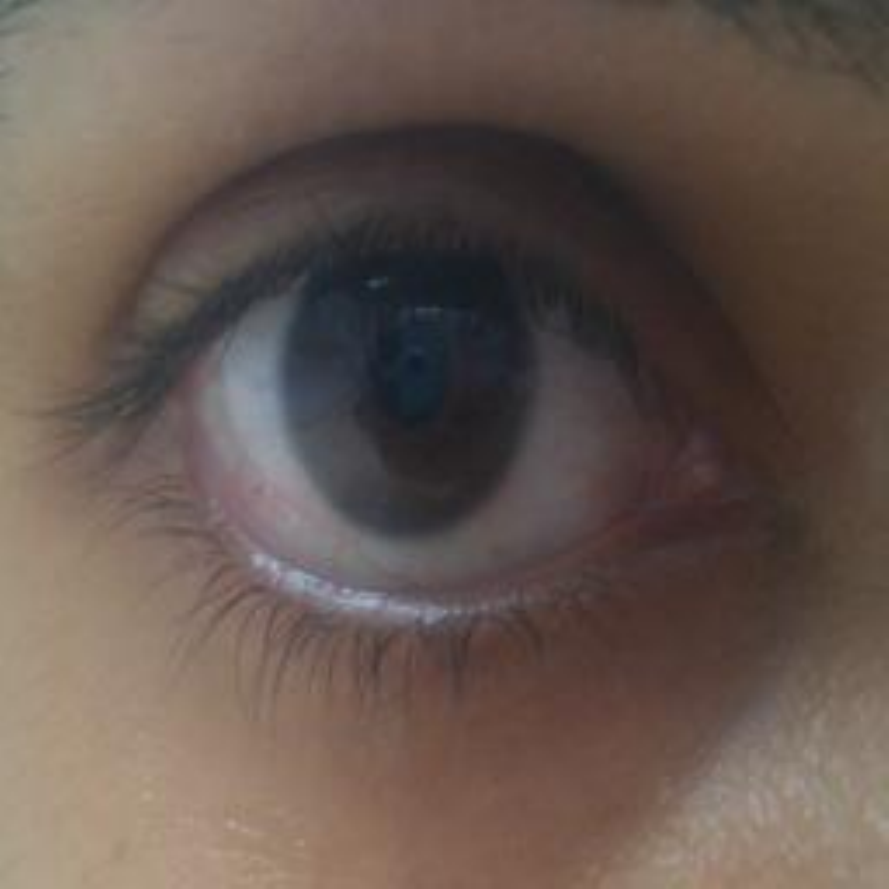}
    \includegraphics[width=0.2\textwidth]{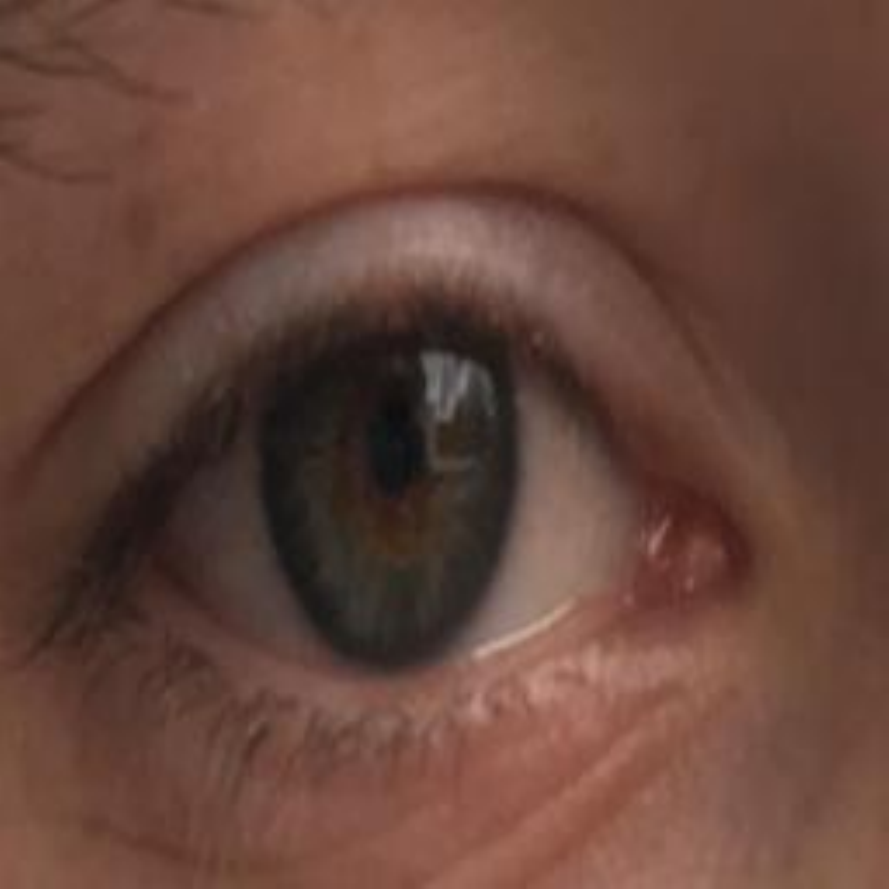}
    \includegraphics[width=0.2\textwidth]{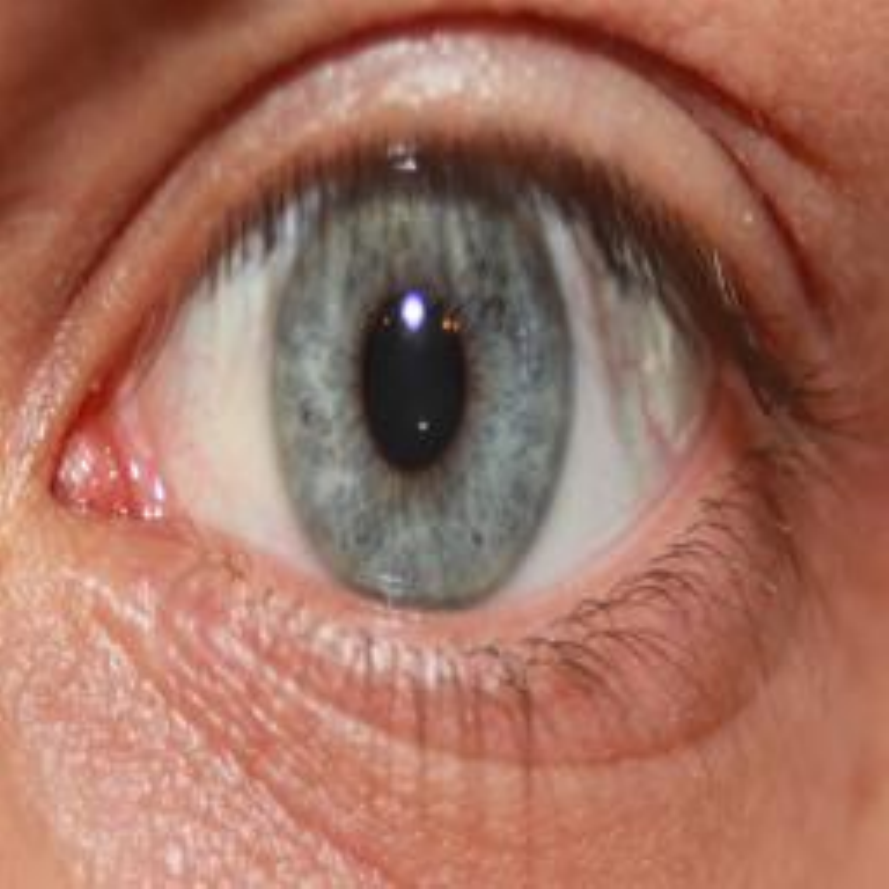}
     \\
    & &\vspace{-7mm}\\
    \rotatebox{90}{\hspace{7mm} \Large Seg. masks} &
    \includegraphics[width=0.2\textwidth]%,trim = 0 85mm 85mm 0,clip]
    {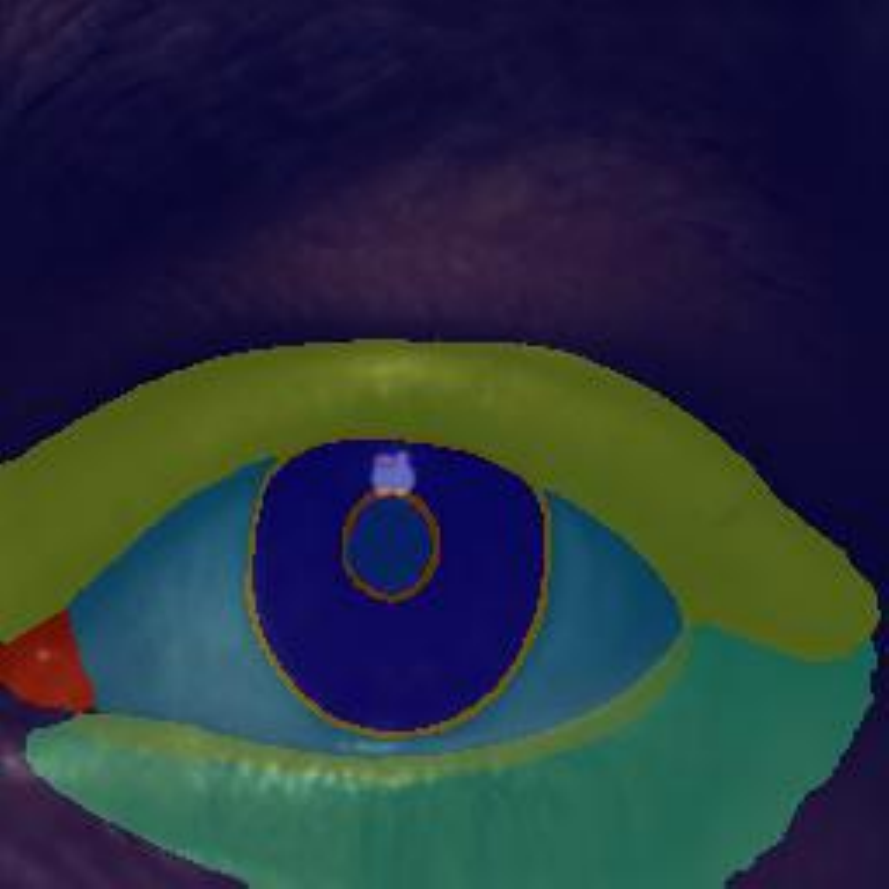} 
    \includegraphics[width=0.2\textwidth]{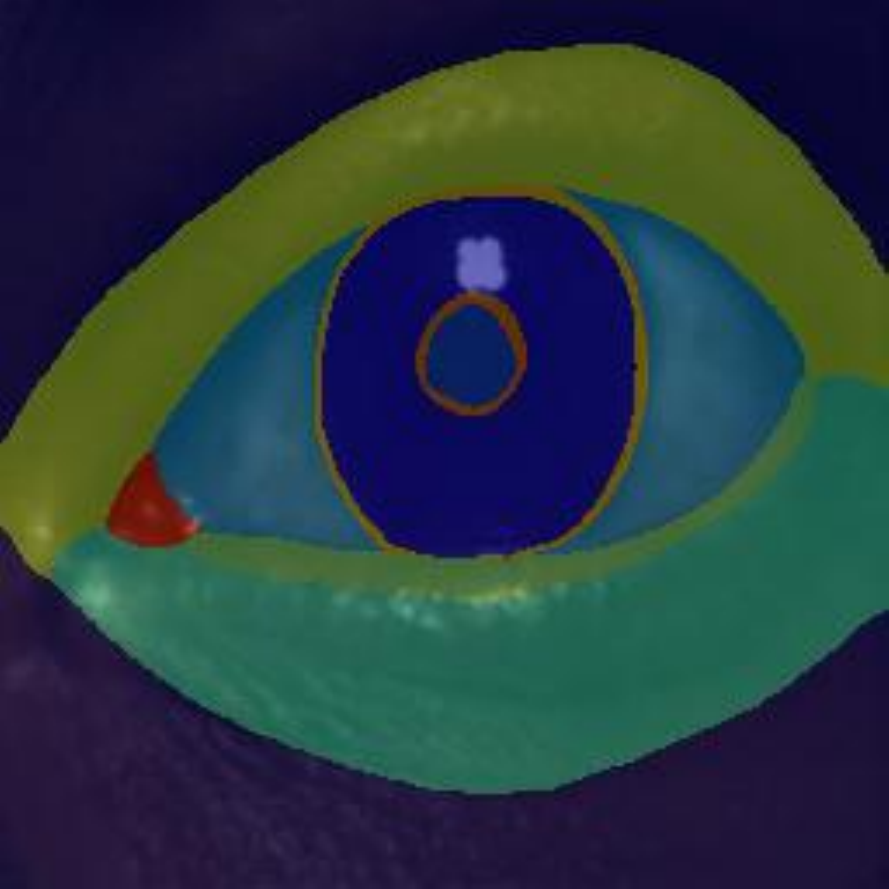} 
    \includegraphics[width=0.2\textwidth]{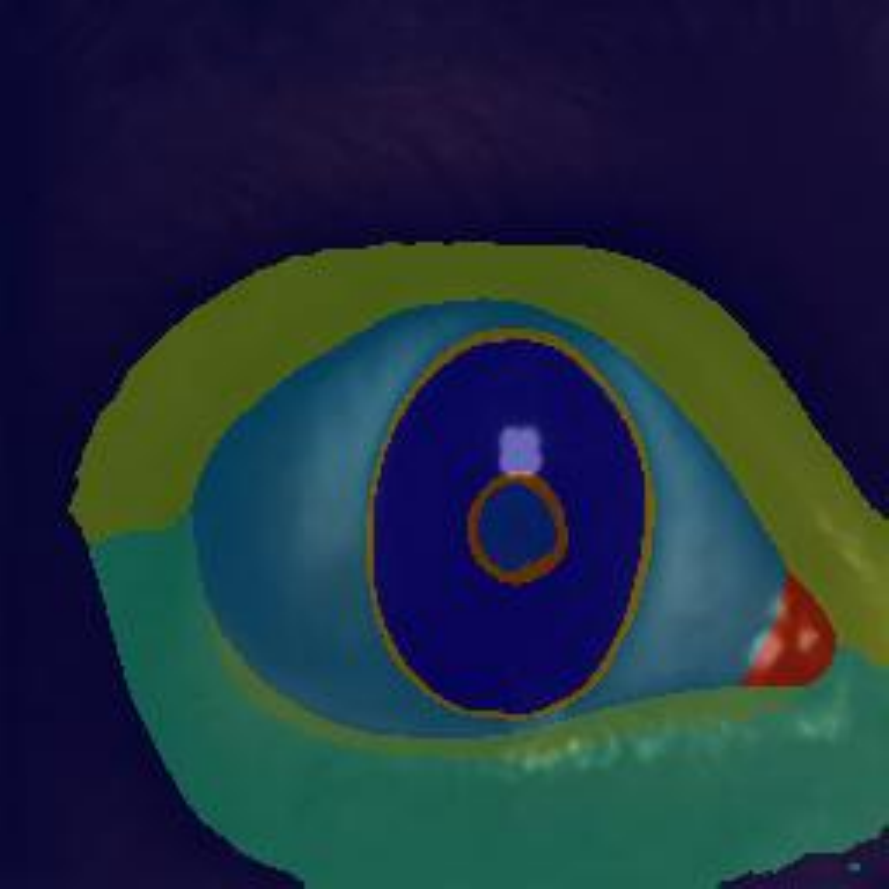}&
    
    \includegraphics[width=0.2\textwidth]{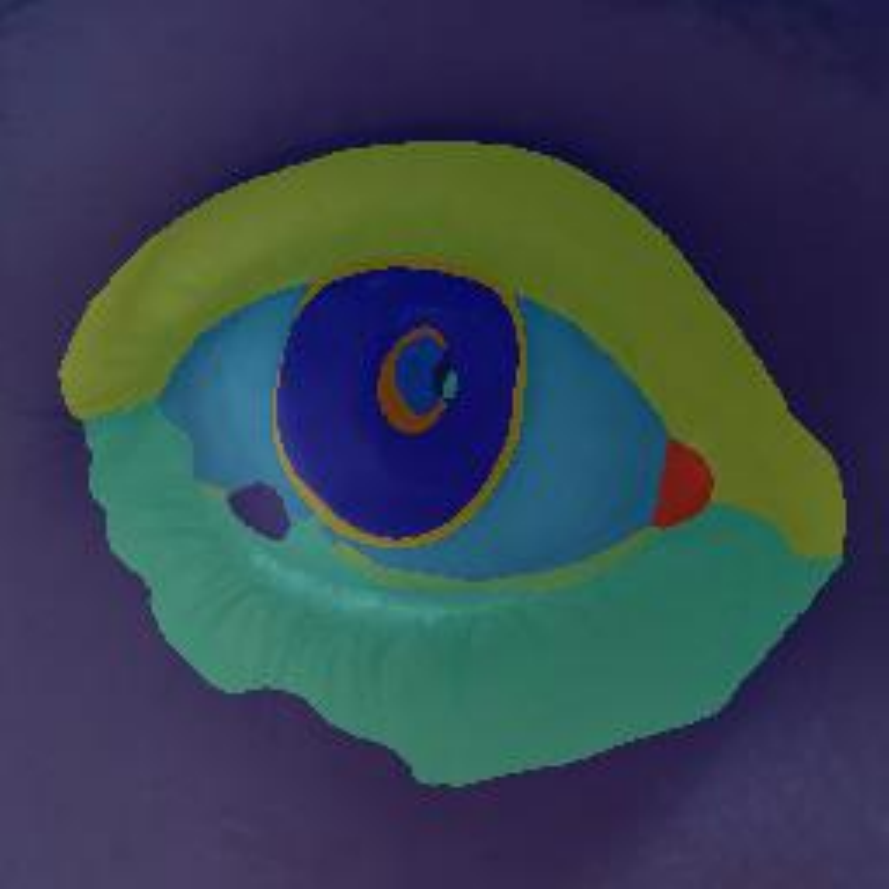}
    \includegraphics[width=0.2\textwidth]{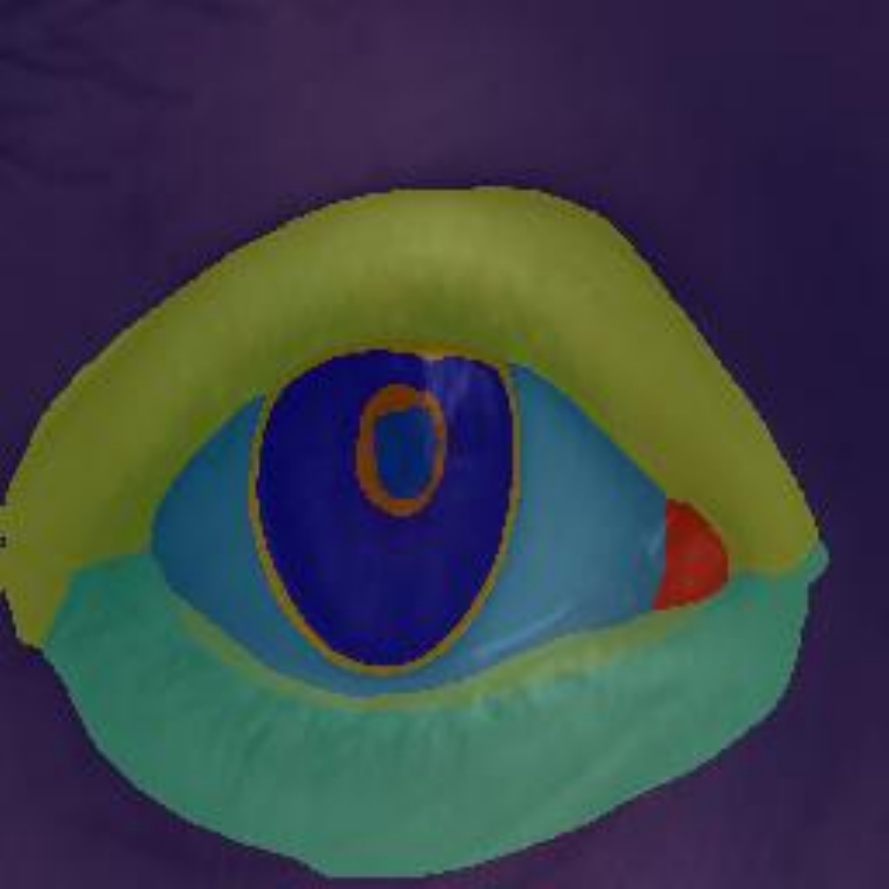}
    \includegraphics[width=0.2\textwidth]{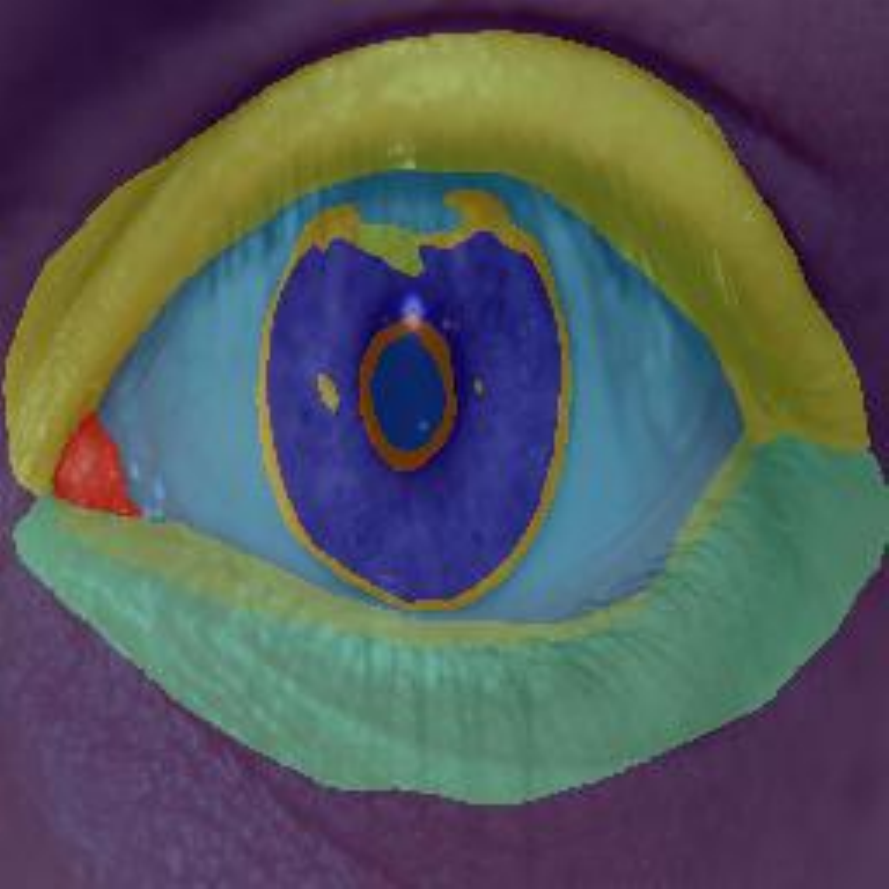}
    %& \multicolumn{2}{c}{ Original images}&\multicolumn{2}{c}{ Synthesized images}\\ 
\end{tabular}}
\end{center}\vspace{-5mm}
\caption{\textbf{Fine-grained segmentation examples.} On the left are training images and corresponding $10$ class masks generated by BiOcularGAN (pupil, pupil boundary, iris, iris boundary, sclera, upper eyelid, lower eyelid, inner lower eyelid, lacrimal caruncle and background). The right part shows sample results generated %by the trained segmentation model 
for the test images. \vspace{-2mm}% from the SMD, MOBIUS and SBVPI datasets.
}
\label{tab:multiclass_annotations}
\end{figure}

\textbf{Fine-grained Segmentation.} Because only a few manual annotations are needed to produce large amounts of training data for learning segmentation models, we manually annotate $2$ images with a $10$-class markup as shown on the left side of Figure \ref{tab:multiclass_annotations}. We then train a segmentation model (i.e., U-Net) with the dataset generated by BiOcularGAN using this fine-grained markup. The right part of Figure \ref{tab:multiclass_annotations} shows some qualitative segmentation results generated with images from the SMD, MOBIUS and SBVPI datasets. Note that despite the fact the BiOcularGAN framework relied only on the DB-StyleGAN2-P model (that generates ocular images of mostly Asian subjects)  and was learned with only $2$ manually annotated images, the trained segmentation model still perform reasonably well on images from all three test datasets.

%\iffalse

\textbf{Real-world Time Requirements.} The training of the data annotation procedure takes around $13$ minutes on PolyU and $11$ minutes on CrossEyed using $8$ annotated images per dataset with our hardware setup. At run-time, a single segmentation mask is produced in $77.8$ ms on average for an $256\times 256$ image produced by DB-StyleGAN2. % -- estimated over $100$ runs.%      {\color{blue}{Bova prestavila sem}}. 

\section{Conclusion}\label{Sec:Conclusion}

In this paper, we presented BiOcularGAN, a framework for generating synthetic datasets of  ocular images with corresponding ground truth segmentation masks. At the heart of the framework is a novel generative model, i.e., the dual-branch StyleGAN2 (DB-StyleGAN2), capable of generating photorealistic aligned bimodal (VIS and NIR) ocular images. Using the proposed BiOcularGAN framework, we showed that it is possible to generate large and representative synthetic datasets that can be used to train competitive segmentation models that generalize well across a diverse set of ocular images. As part of our future work, we plan to further explore the DB-StyleGAN2 models for cross-modal recognition tasks and investigate image editing possibilities within the DB-StyleGAN2 latent space. % and study GAN inversion solution that allow embedding real images into the DB-StyleGAN2 latent space.      

\section*{Acknowledgements}\label{Sec:ack}

Supported in parts by the Slovenian Research Agency ARRS through the Research Programmes P2-0250(B) "Metrology and Biometric System" and P2--0214 (A) “Computer Vision”, the ARRS Project J2-2501(A) "DeepBeauty" and the ARRS junior researcher program.

{\small
\bibliographystyle{ieee}
\bibliography{main}
}

\if\showAppendix1
\clearpage

%*************************************************
% Additional material for the appendix or supplementary file
%
%
%
%??***********************************************
\appendix{}

{\begin{table*}[t]
\resizebox{\textwidth}{!}{%
\begin{tabular}{cc|ccc|ccc|ccc}
\toprule
        
    \multirow{3}{*}{\bf{Seg. Model}}& \multirow{3}{*}{\bf{Labels from}} &  \multicolumn{9}{c}{\bf{Trained on PolyU (DB-StyleGAN2-P)}} \\\cline{3-11}
    & & \multicolumn{3}{c}{\bf{SMD$^{\dagger}$}} & \multicolumn{3}{c}{\bf{MOBIUS$^{\dagger,\ddagger}$}} & \multicolumn{3}{c}{\bf{SBVPI$^{\dagger,\ddagger}$}}  \\\cline{3-11}
    %\midrule
    & & \bf{IoU $\uparrow$} & \bf{$F_{1}$ $\uparrow$} & \bf{Pixel error $\downarrow$ [\%]} & \bf{IoU $\uparrow$} & \bf{$F_{1}$ $\uparrow$} & \bf{Pixel error $\downarrow$ [\%]} & \bf{IoU $\uparrow$} & \bf{$F_{1}$ $\uparrow$} & \bf{Pixel error $\downarrow$ [\%]} \\ 
    \midrule
    \multirow{3}{*}{U-Net} & $2$ annotations &  
        $0.686 \pm 0.102$  &  $0.767 \pm 0.116$  &  $0.046 \pm 0.022$ & 
        $0.554 \pm 0.177$  &  $0.651 \pm 0.198$  &  $0.091 \pm 0.055$ &  
        $0.782 \pm 0.046$  &  $0.866 \pm 0.038$  &  $0.047 \pm 0.015$  \\
    
    & $4$ annotations &  
       $0.696 \pm 0.091$  &  $0.781 \pm 0.099$  &  $0.046 \pm 0.021$ &  
       $0.564 \pm 0.166$  &  $0.659 \pm 0.181$  &  $0.087 \pm 0.051$ & 
       $0.789 \pm 0.042$  &  $0.871 \pm 0.035$  &  $0.046 \pm 0.014$  \\
        
     & $8$ annotations &  
       $0.772 \pm 0.081$  &  $0.853 \pm 0.070$  &  $0.041 \pm 0.025$ &  
       $0.584 \pm 0.173$  &  $0.674 \pm 0.187$  &  $0.082 \pm 0.051$ & 
       $0.818 \pm 0.052$  &  $0.891 \pm 0.041$  &  $0.040 \pm 0.015$  \\
    
\bottomrule
\multicolumn{11}{l}{$^\dagger$Cross-dataset experiments; $^\ddagger$Cross-ethnicity experiments}
\end{tabular}
}
%\resizebox{\textwidth}{!}{%
\begin{minipage}[b]{1\textwidth}
\vspace{2mm}
\captionof{table}{\textbf{Impact of the number of manually annotated images on segmentation performance.} The table shows segmentation results generated with a U-Net model learned based on training data produced with the BiOcularGAN framework, where the framework itself was learned with either $2$, $4$ or $8$ manually annotated images.  \label{tab:state_of_art_comparison_annotations}}
\end{minipage}
\end{table*}
\vspace{20mm}
}

%\end{table*}

\section{Appendix}

In this appendix, we present some additional results and discussions not included in the main part of the paper. Specifically, we $(i)$ analyze the impact the number of manually annotated images has on the quality of the training data generated by the BiOcularGAN framework, $(ii)$ show examples of qualitative segmentation results as a function of the number of manually annotated images used to learn the BiOcularGAN framework, $(iii)$ present a broader cross-section of qualitative results generated based on the fine-grained $10$-class markup, $(iv)$ report style-mixing experiments, %$(v)$ show additional visual/qualitative examples at higher resolutions ($512\times 512$), 
$(v)$ provide additional implementation details, and $(vi)$ some final discussions.

\subsection{Impact of manual annotations} 

To get a better insight into the behavior of the BiOcularGAN framework, we investigate in this section how the number of manually annotated images affects the performance of the segmentation models trained with the synthetic training data produced by the BiOcularGAN framework. For this experiment, we train a U-Net segmentation model with training data generated by BiOcularGAN  and the DB-StyleGAN2-P model. To learn the segmentation-mask generation procedure, we use either $2$, $4$ or $8$ manually annotated images, where the annotations again consist of four classes (iris, pupil, sclera and background). Similarly, as in the main part of the paper, we again use the (frontal gaze) VIS images from SMD, MOBIUS and SBVPI for testing.

\textbf{Quantitative Results.} From Table \ref{tab:state_of_art_comparison_annotations} we observe that (as expected) the segmentation performance increases with the number of manually annotated images across all test datasets and with respect to all performance measures reported. If we focus on the $F_1$ score, for example, we see an increase from $0.767$ when using $2$ annotated images to $0.853$ when using $8$ on the SMD dataset. 
Similarly, the $F_1$ results are also improved on the MOBIUS and SBVPI dataset, where an increase from $0.651$ and $0.866$ (with $2$ annotated images) to $0.674$ and $0.891$ (with $8$ annotated images) is seen, respectively. Nonetheless, even with only $2$ manually annotated images,  BiOcularGAN is still able to produce training data of reasonable quality for learning the segmentation model. Thus, if a higher level of granularity is needed in the segmentation masks, a suitable trade-off can be selected between the labor-intensive manual annotation process and the desired segmentation performance.\\ 

%\vspace{6.9cm}

\begin{figure}[!ht] 
%\vspace{4.2cm}
\begin{center}
\resizebox{\columnwidth}{!}{%
\begin{tabular}{ccc}
%\toprule
    %\midrule
    \rotatebox{90}{\hspace{2mm} \large $2$ annotations} & 
    \includegraphics[width=0.16\textwidth]%,trim = 0 85mm 85mm 0,clip]
    {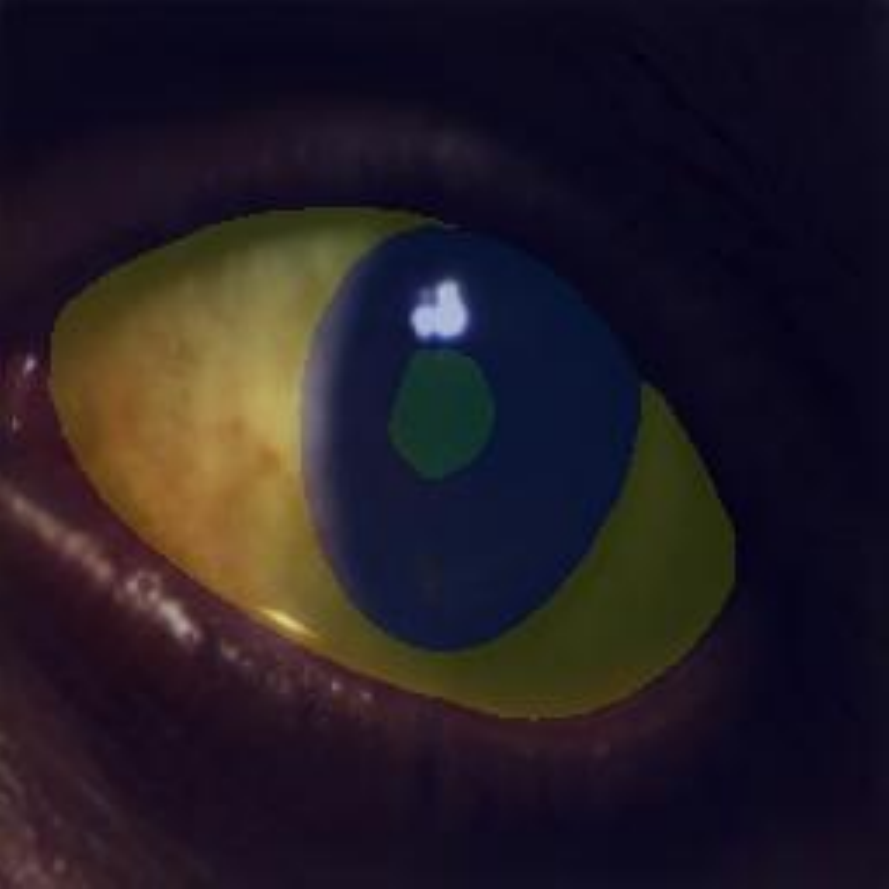} 
    \includegraphics[width=0.16\textwidth]{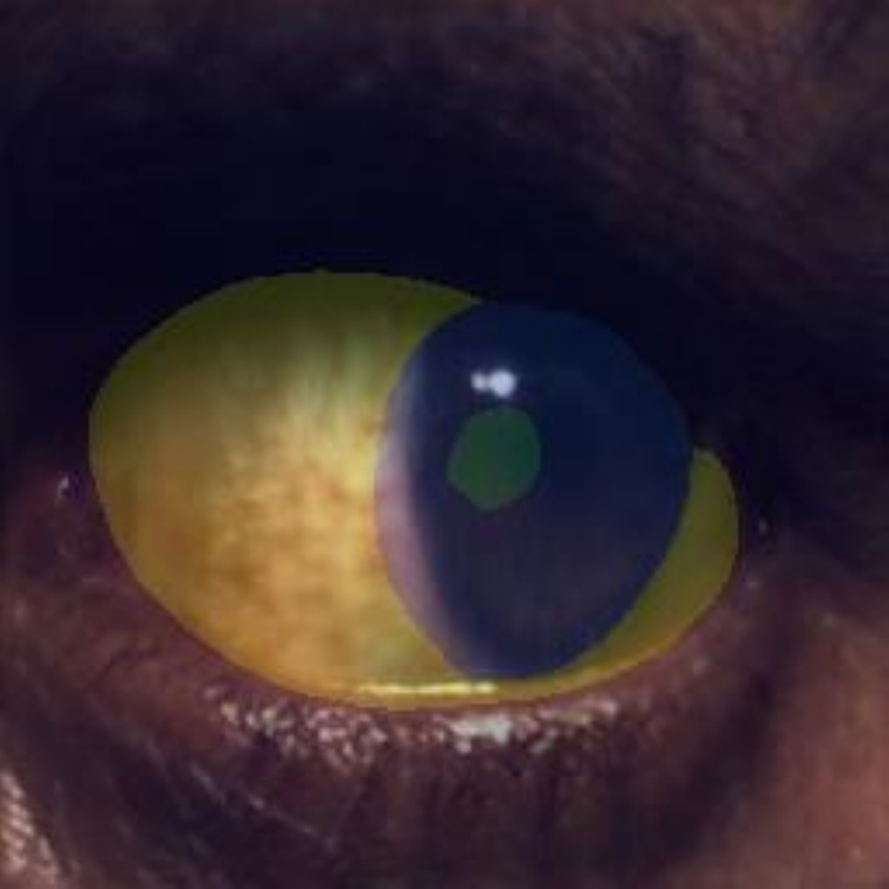}  
    \includegraphics[width=0.16\textwidth]{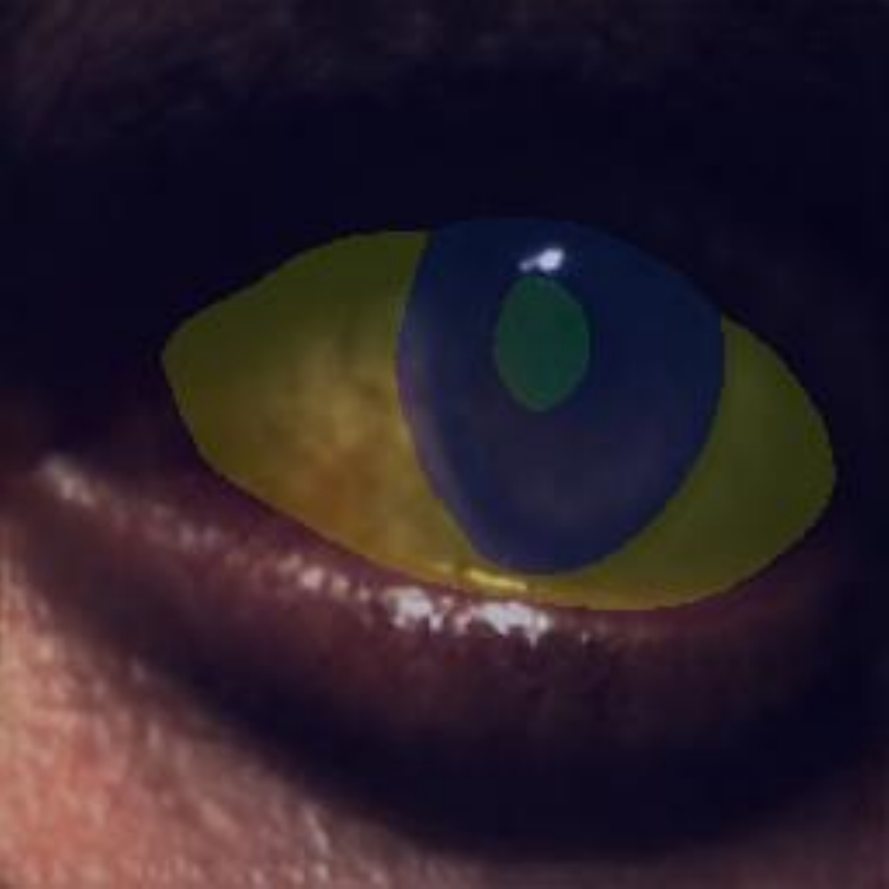} 
    &
    \includegraphics[width=0.16\textwidth]%,trim = 0 85mm 85mm 0,clip]
    {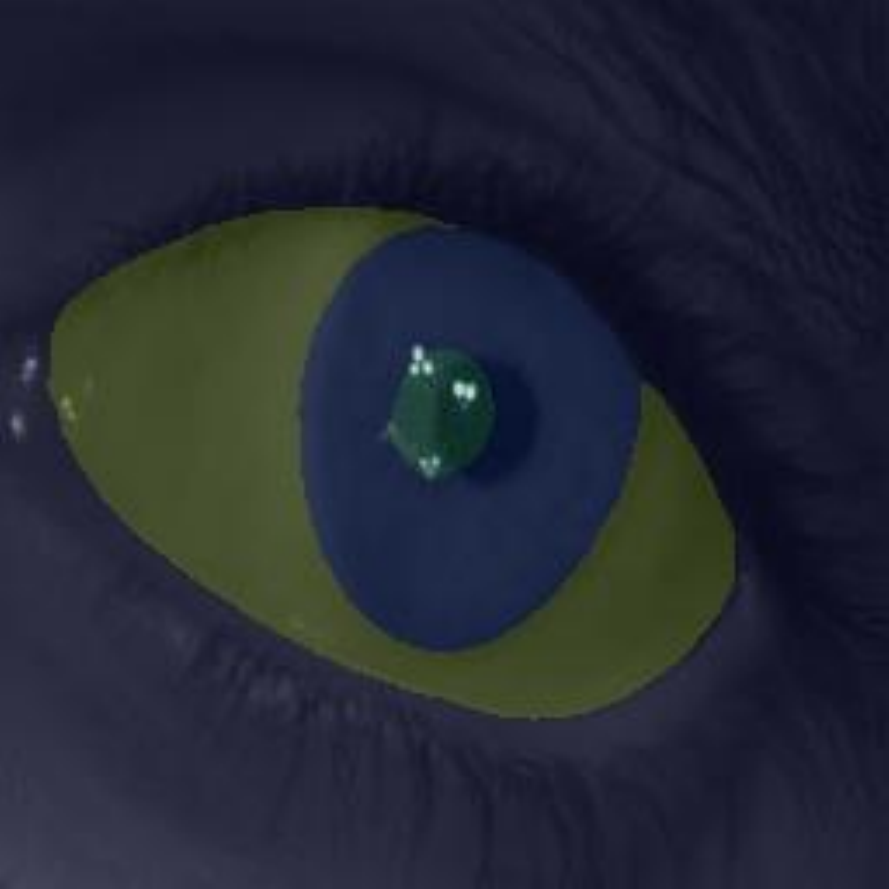} 
    \includegraphics[width=0.16\textwidth]{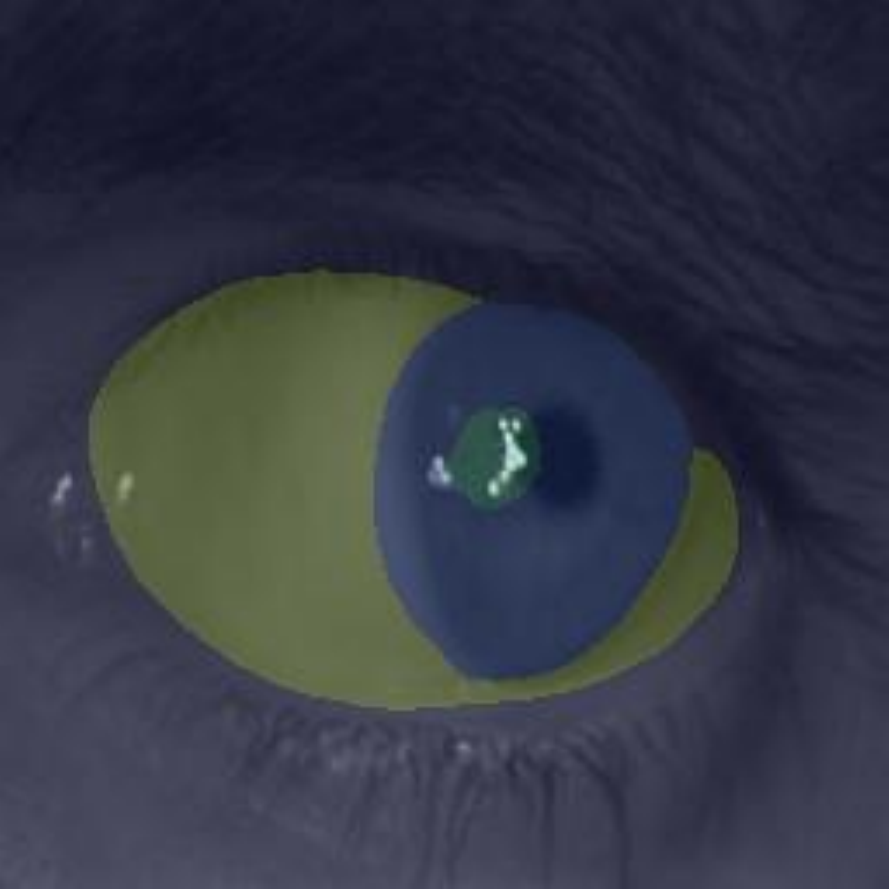} 
    \includegraphics[width=0.16\textwidth]{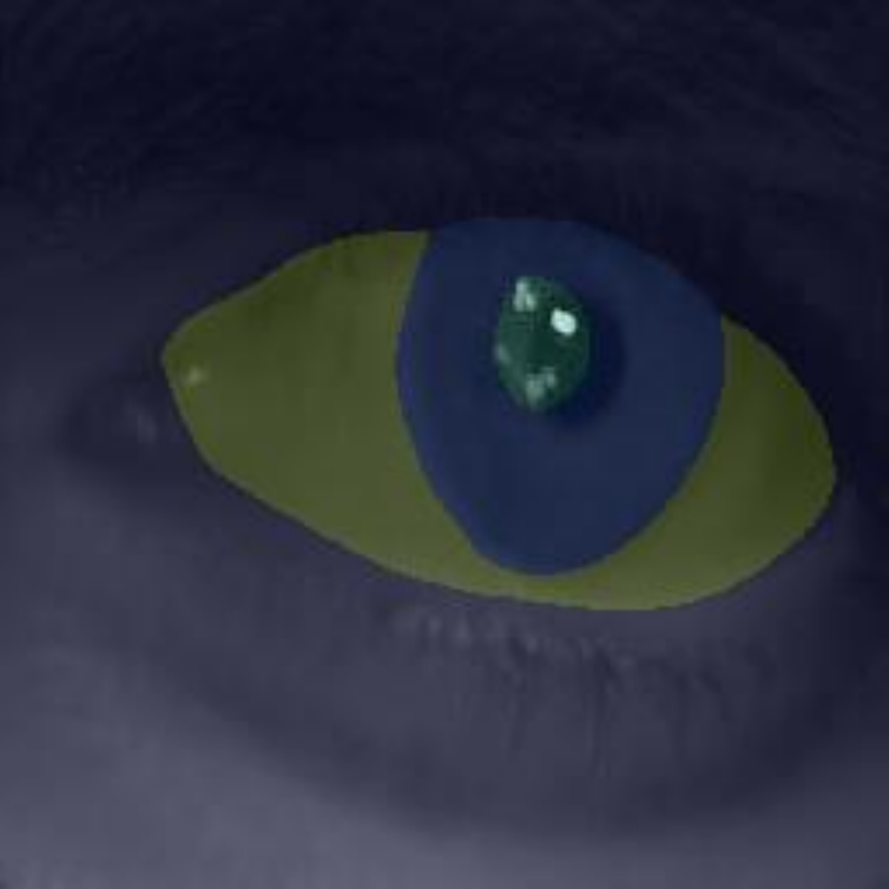} 
    \\
    \rotatebox{90}{\hspace{2mm} \large $4$ annotations} & 
    \includegraphics[width=0.16\textwidth]{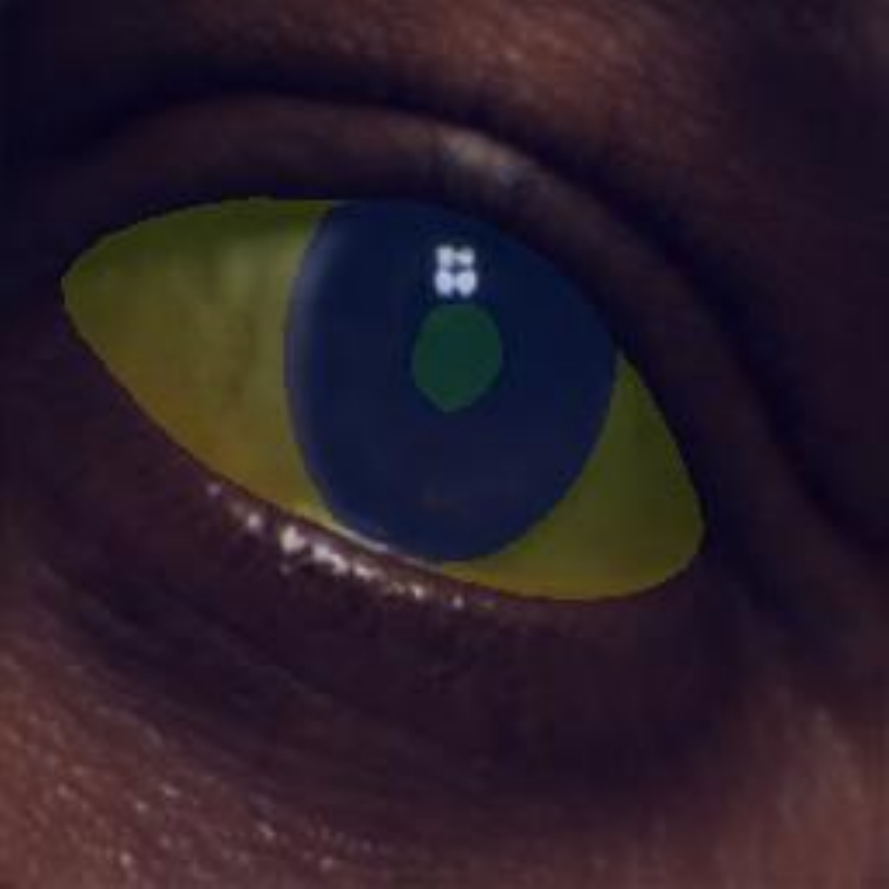}
    \includegraphics[width=0.16\textwidth]{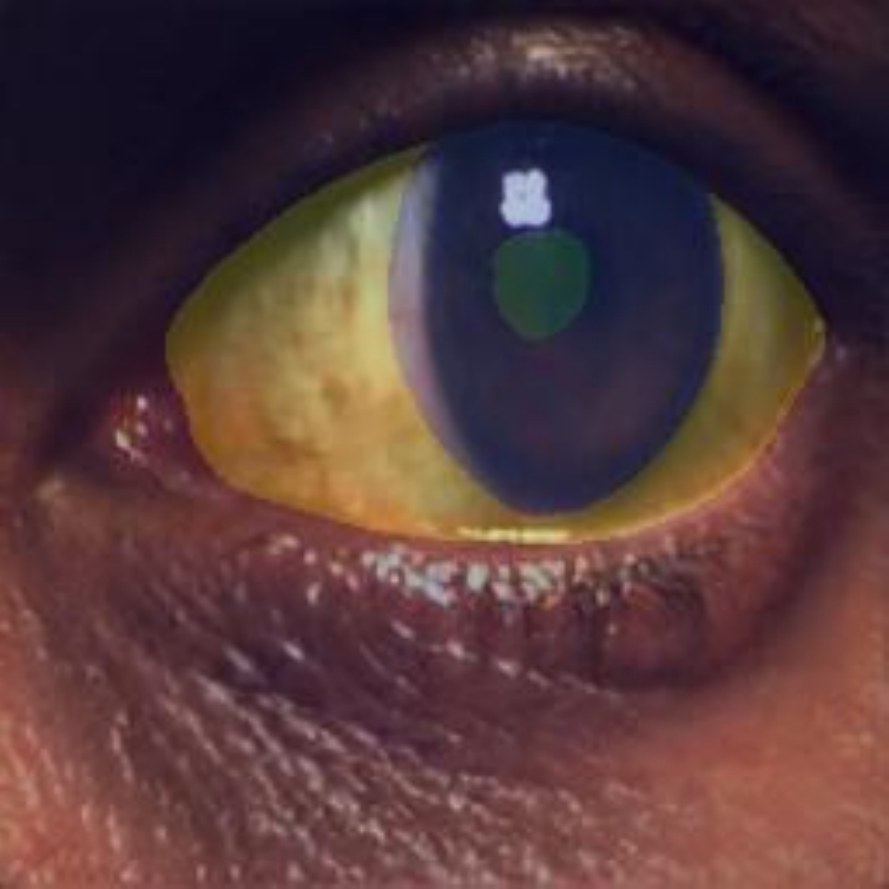} 
    \includegraphics[width=0.16\textwidth]{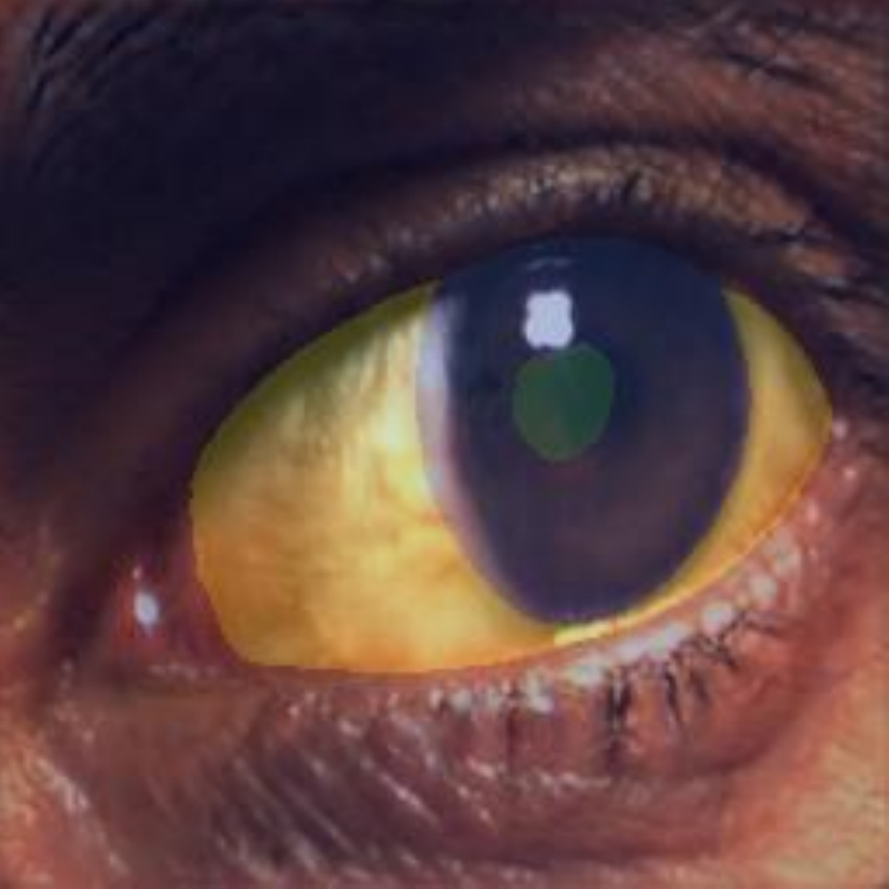} 
    &
    \includegraphics[width=0.16\textwidth]{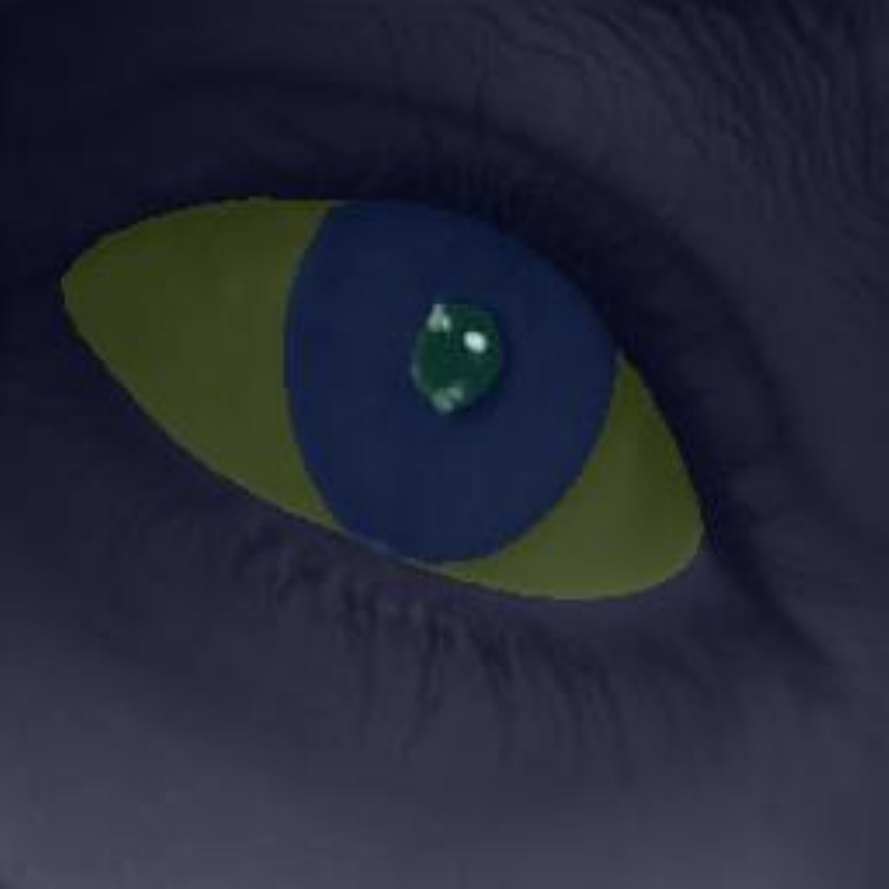} 
    \includegraphics[width=0.16\textwidth]{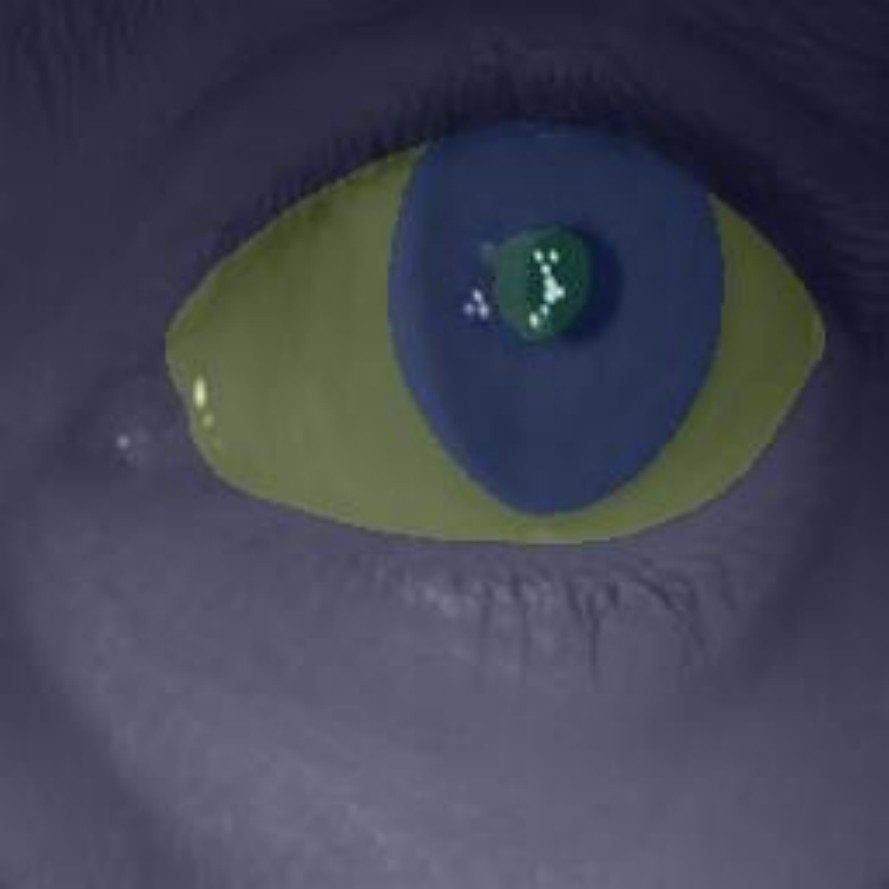} 
    \includegraphics[width=0.16\textwidth]{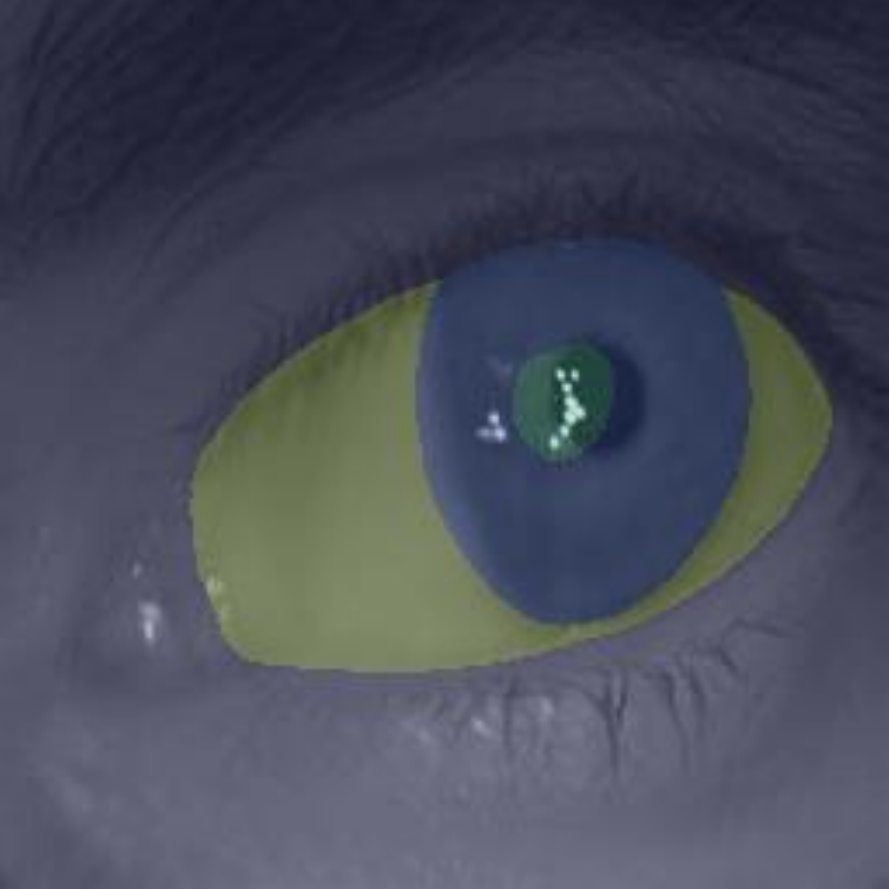} 
    
    \\
    \rotatebox{90}{\hspace{2mm} \large $8$ annotations} & 
    \includegraphics[width=0.16\textwidth]{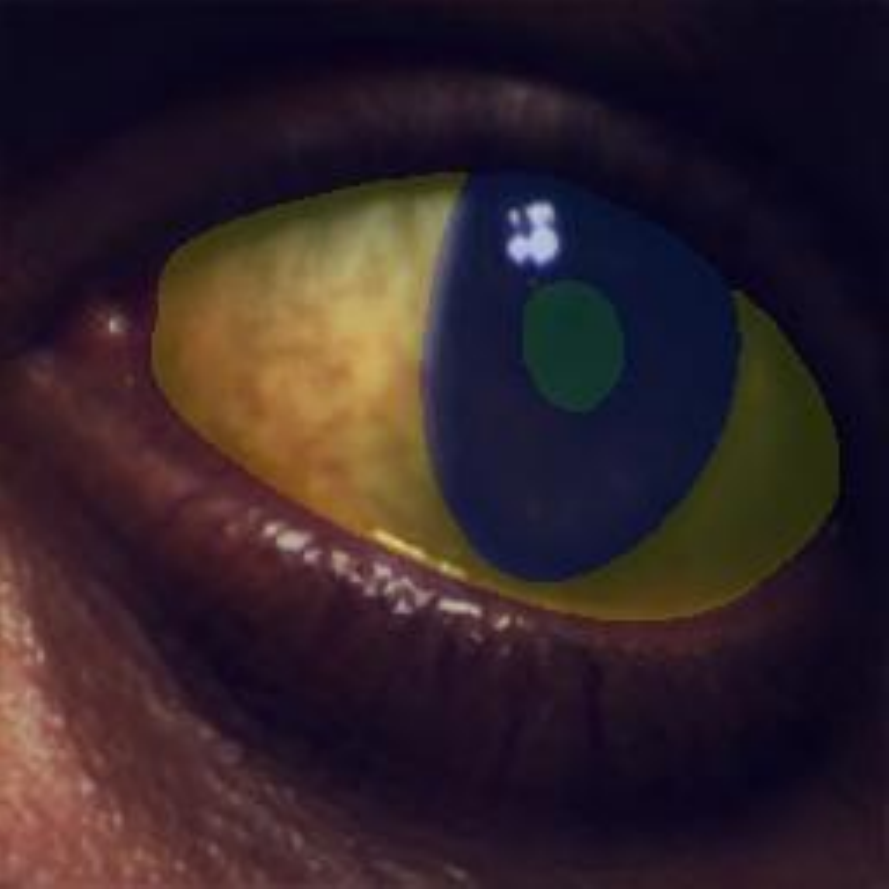}
    \includegraphics[width=0.16\textwidth]{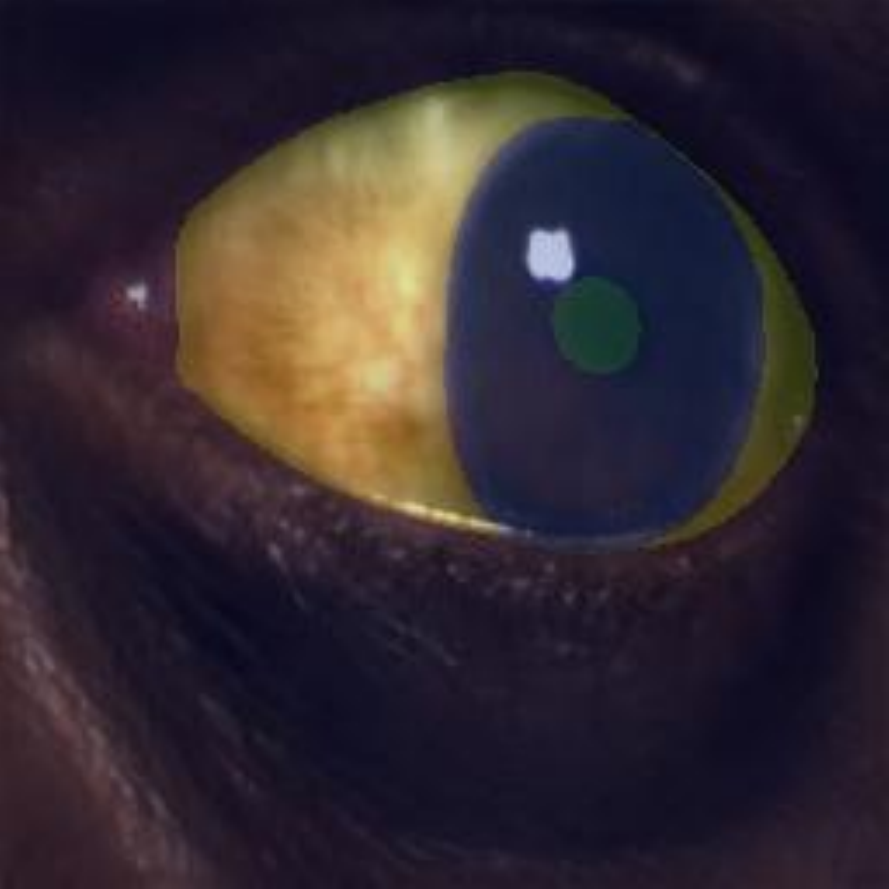}
    \includegraphics[width=0.16\textwidth]{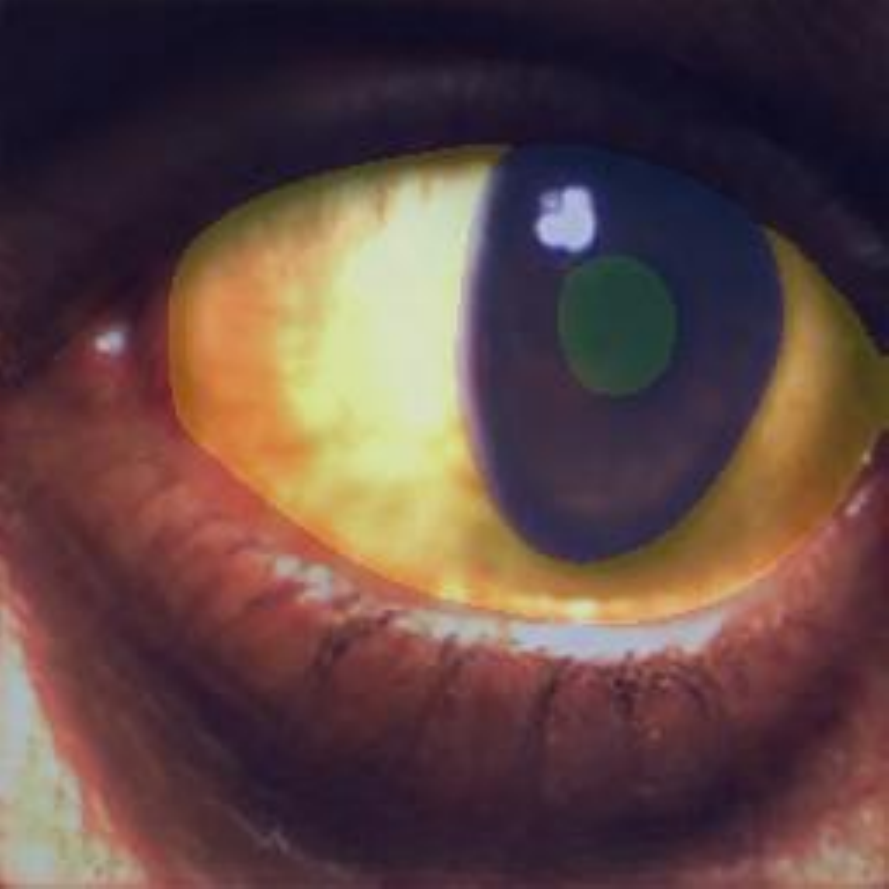}
    &
    \includegraphics[width=0.16\textwidth]{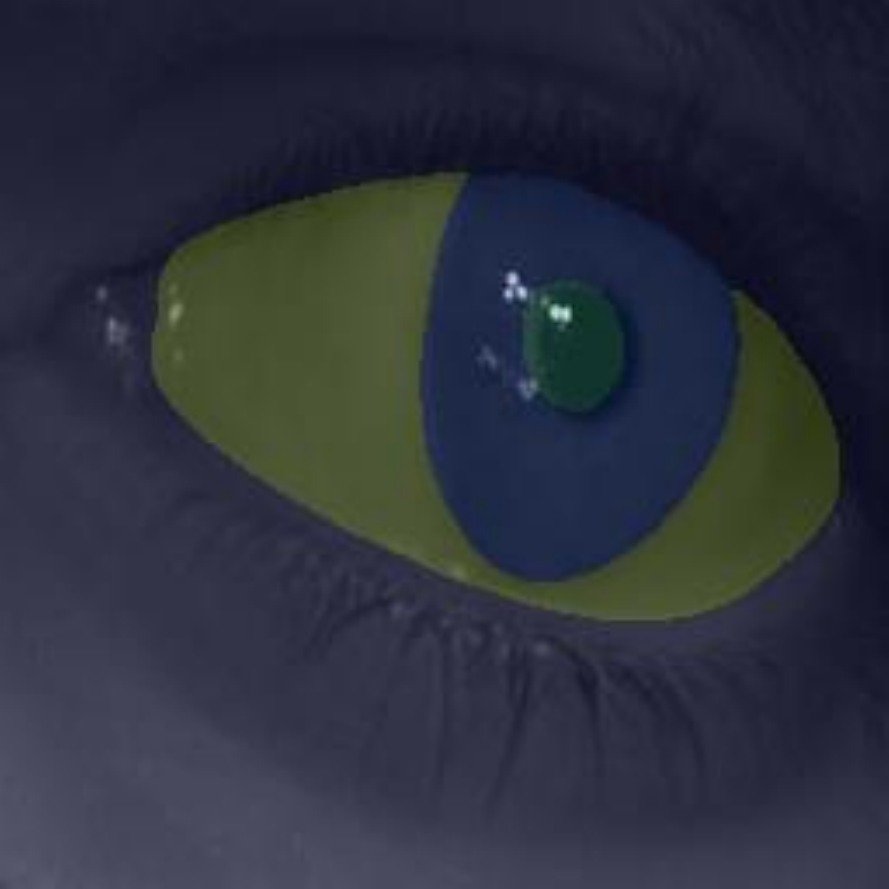}
    \includegraphics[width=0.16\textwidth]{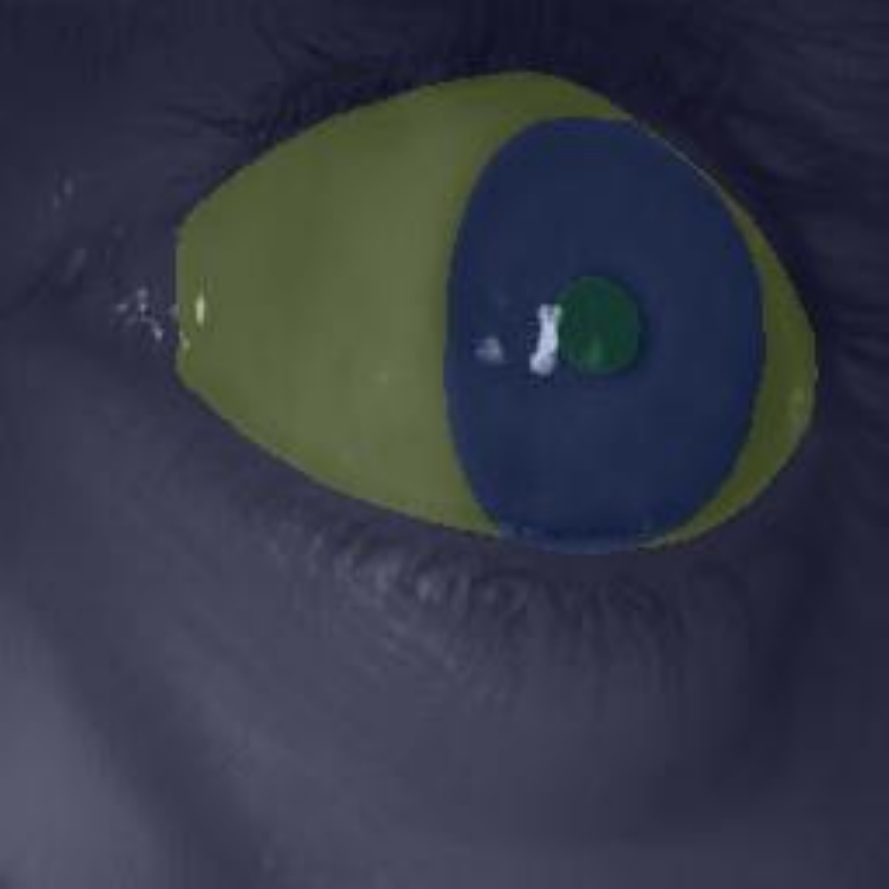}
    \includegraphics[width=0.16\textwidth]{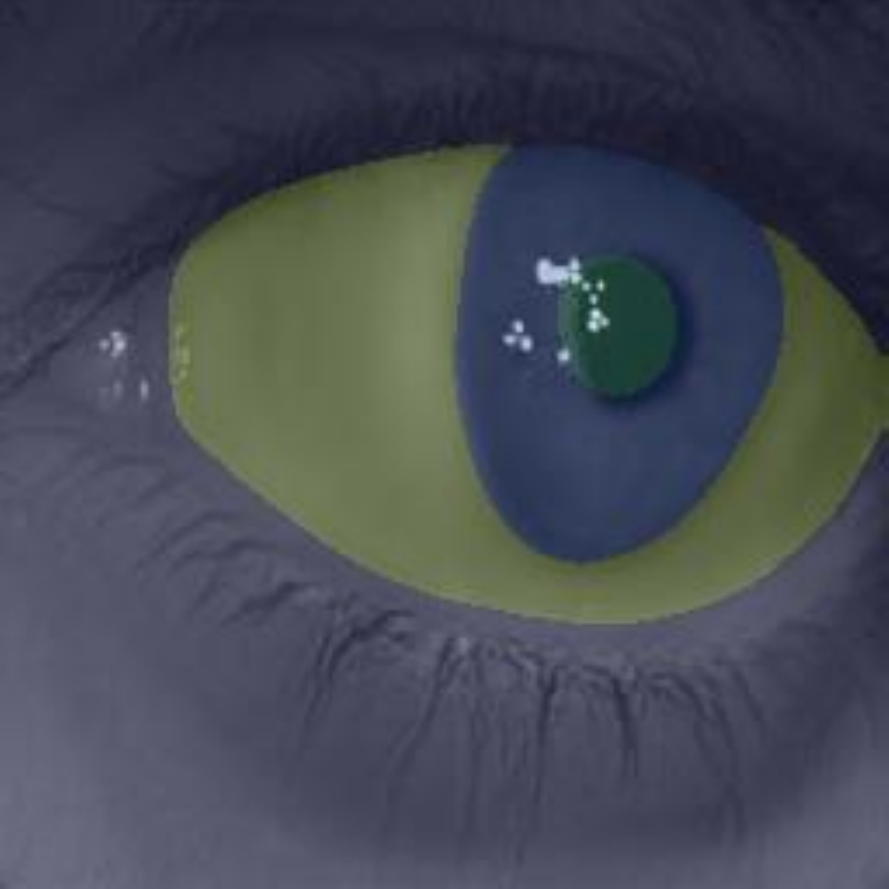}
    %\\
    %\hline 
    %& &&\vspace{-3mm}\\
    %\rotatebox{90}{\hspace{10mm} \large Ground truth} & 
    %\includegraphics[width=0.23\textwidth]{figures/segmentation_mask_comparison/example_GT_merge_PolyU_NIR_RGB__SMD_256_64.pdf} &
    %\includegraphics[width=0.23\textwidth]{figures/segmentation_mask_comparison/example_GT_merge_PolyU_NIR_RGB__MOBIUS_256_center_3.pdf} &
    %\includegraphics[width=0.23\textwidth]{figures/segmentation_mask_comparison/example_GT_merge_PolyU_NIR_RGB__SBVPI_sclera_center_56.pdf}
    %\\ 
    
    %& \multicolumn{2}{c}{ Original images}&\multicolumn{2}{c}{ Synthesized images}\\ 
\end{tabular}}
\end{center}\vspace{-4mm}
\caption{\textbf{Example training data generated by BiOcularGAN.} The figure shows a comparison in the quality of segmentation masks generated as a function of the number of manually annotated images used to learn the mask-generation procedure of BiOcularGAN. % the The results were generated with two U-Net models, trained on artificial data generated by the DatasetGAN and BiOcularGAN frameworks, learned with the DB-StyleGAN2-P model. Note that BiOcularGAN better captures the semantic content of the images and therefore 
%leads to %higher quality training data and consequently 
%better segmentation results. % {\color{blue}  Observation: U-Net on BiOcularGAN has more rounded and detailed masks... U-Net on DatasetGAN has issues with the blue iris in SBVPI and the pupil in MOBIUS. Probable reason: the GAN was trained on PolyU (only has brown irises and its pupils are often hard to distinguish). .... Sem imel razlago o latent vektorjih in informaciji, ki jo imajo na podlagi VIS in NIR slik, vendar po pregledu training primerov ni ravno smiselna...
%Thus the latent vectors probably did not include enough information regarding this split, when trained on only VIS images. However, in NIR images these regions can be easily separated (distinguished), thus it stands to reason that the DB-StyleGAN2 could encode more semantic information into its latent vectors than the normal StyleGAN2 (since the DB version uses shared latent maps to generate VIS and NIR images). This in turn helps the ensemble MLP classifiers generate better training examples. 
%}
%\color{red}{ Mogoče moramo primerjati artificial training podatke}
}
\label{fig:annotation_number_training}\vspace{-2mm}
\end{figure}
\begin{figure}[!ht] 
\begin{center}
\resizebox{\columnwidth}{!}{%
\begin{tabular}{cccc}
%\toprule
    %\midrule
   &  \Large SMD & \Large MOBIUS & \Large SBVPI\\
    \rotatebox{90}{\hspace{2mm} \large $2$ annotations} & 
    \includegraphics[width=0.16\textwidth]{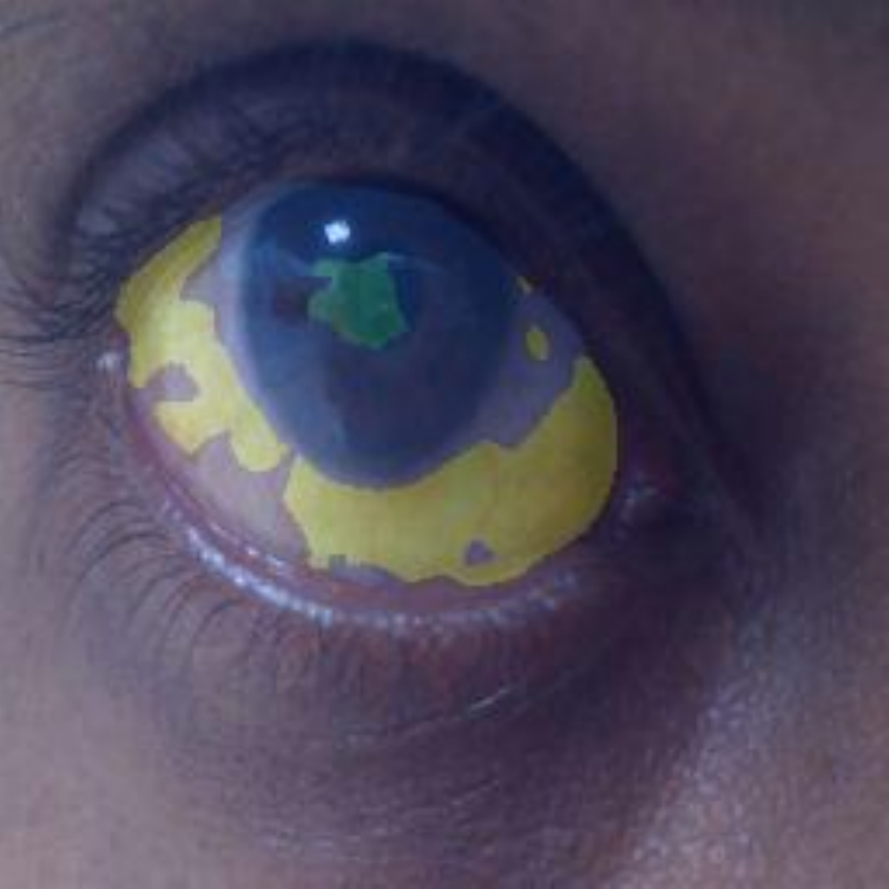} 
    \includegraphics[width=0.16\textwidth]{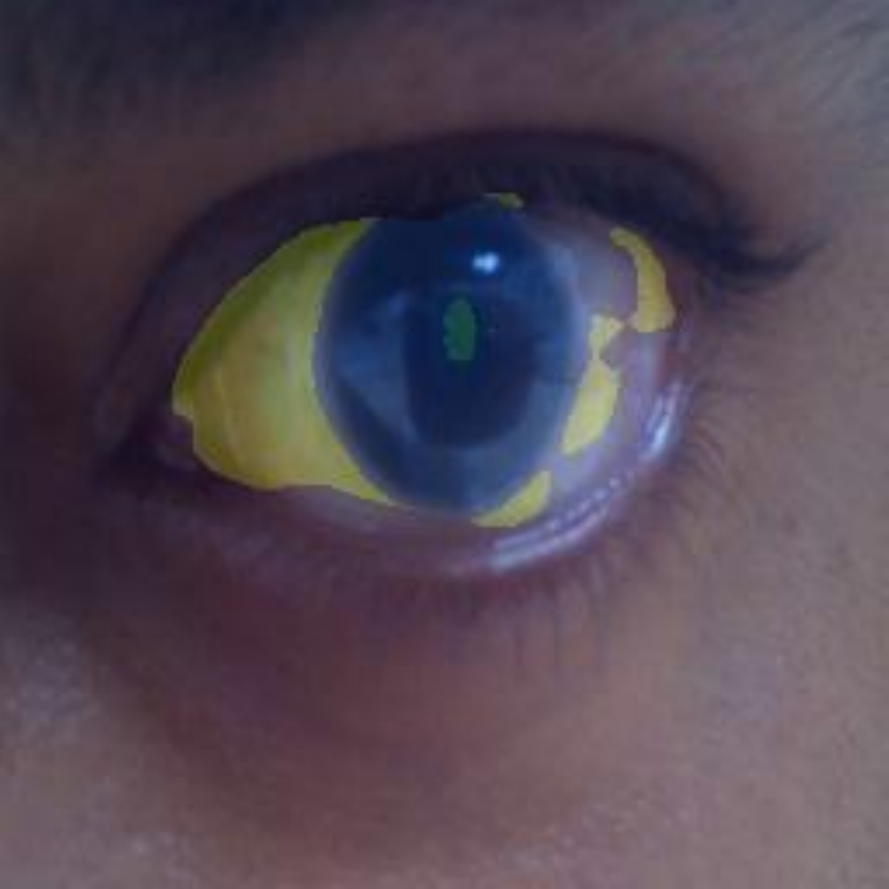} 
    &
    \includegraphics[width=0.16\textwidth]{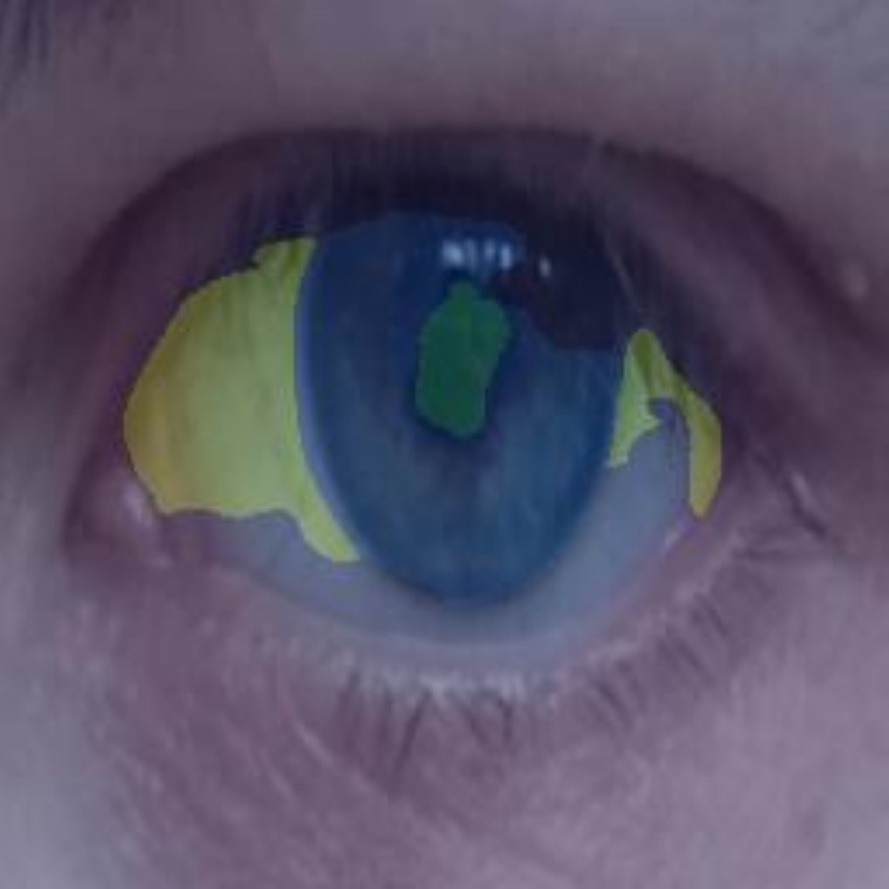}  
    \includegraphics[width=0.16\textwidth]{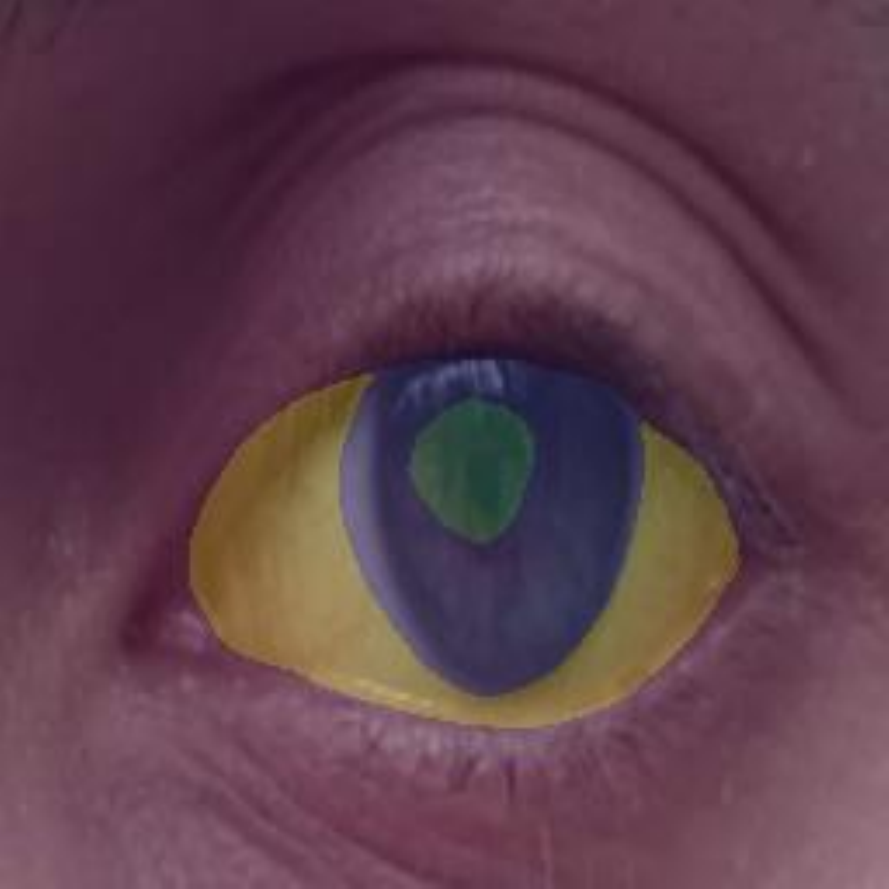}  
    &
    \includegraphics[width=0.16\textwidth]{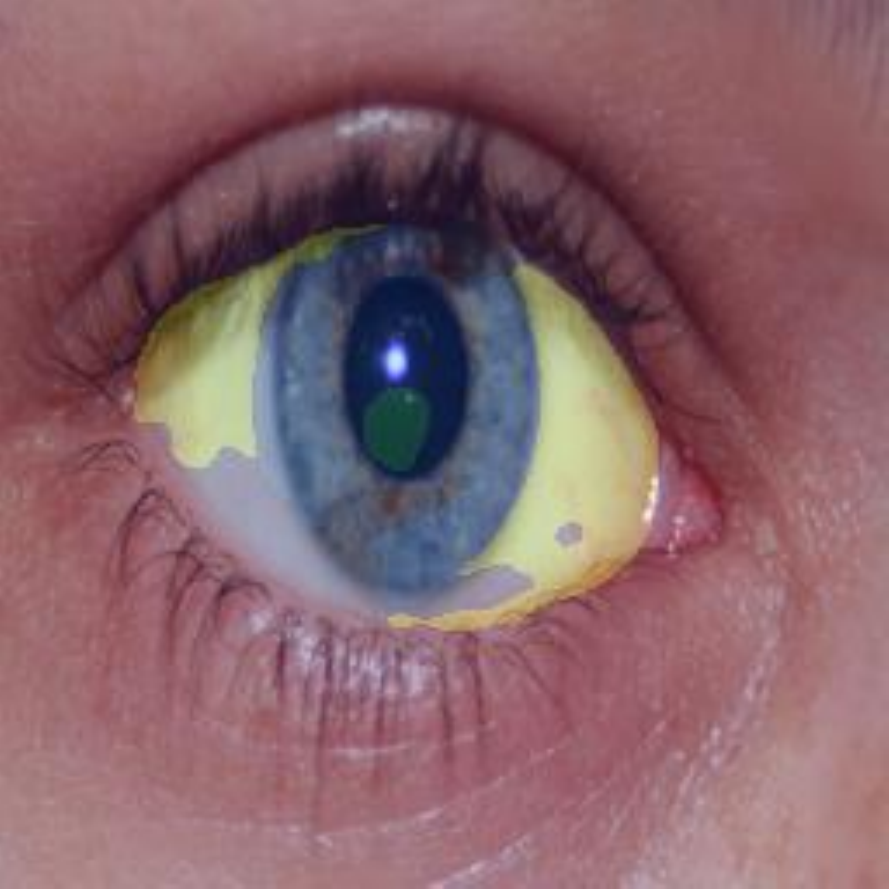}  
    \includegraphics[width=0.16\textwidth]{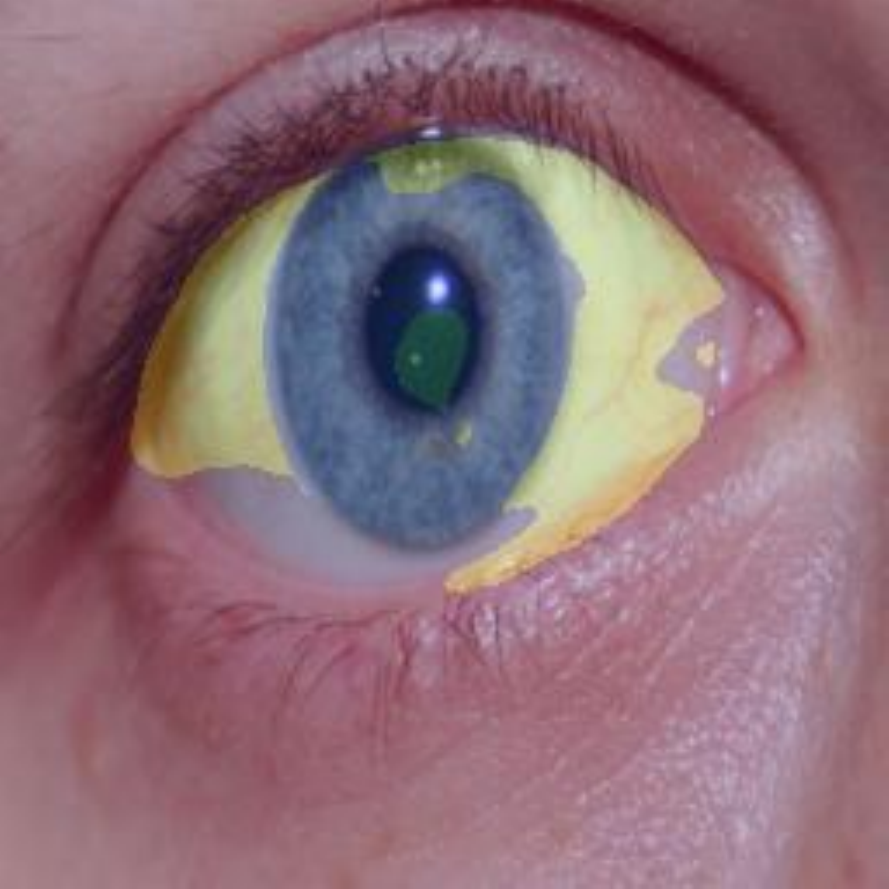} 
    \\
    \rotatebox{90}{\hspace{2mm} \large $4$ annotations} & 
    \includegraphics[width=0.16\textwidth]{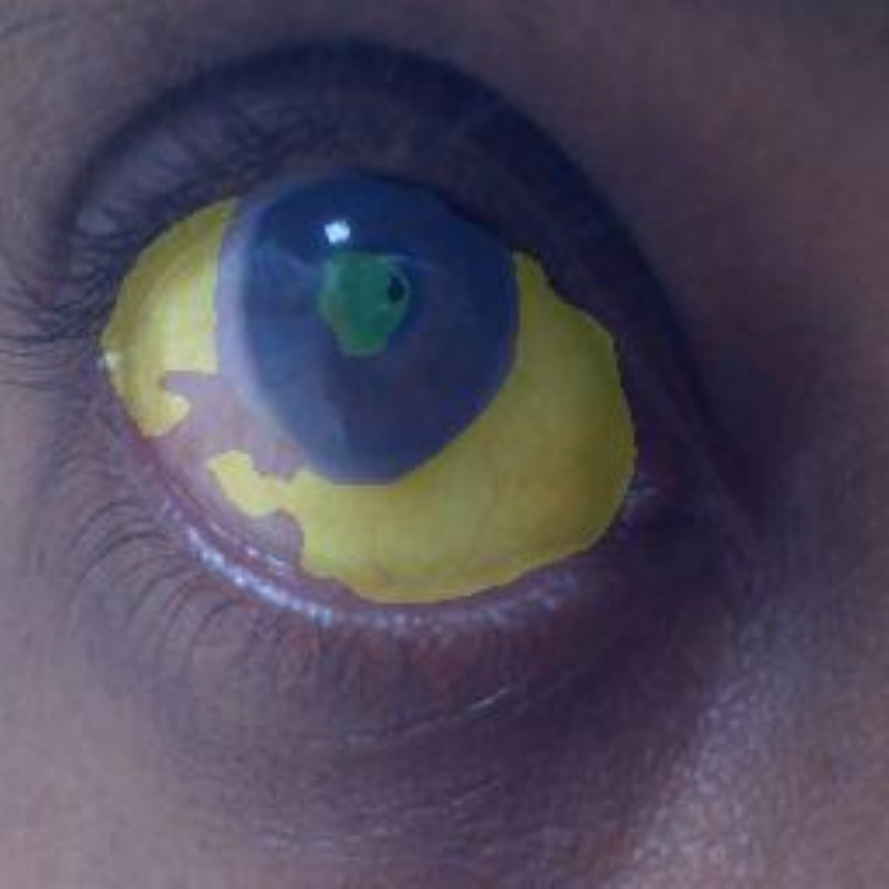}  
    \includegraphics[width=0.16\textwidth]{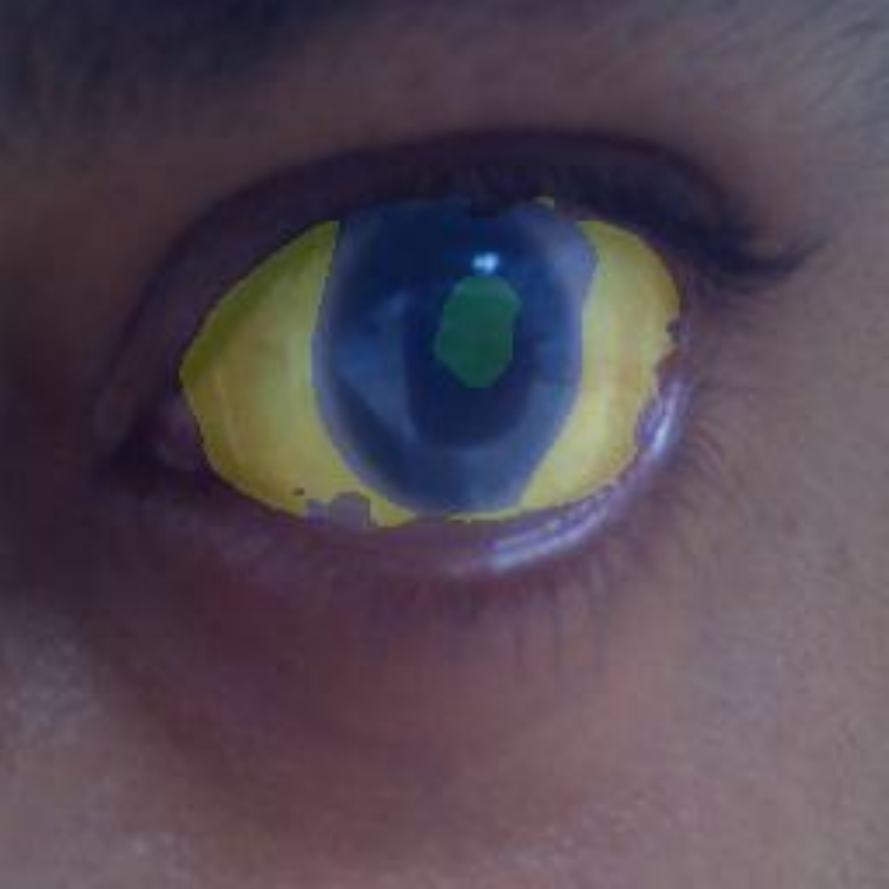}  
    &
    \includegraphics[width=0.16\textwidth]{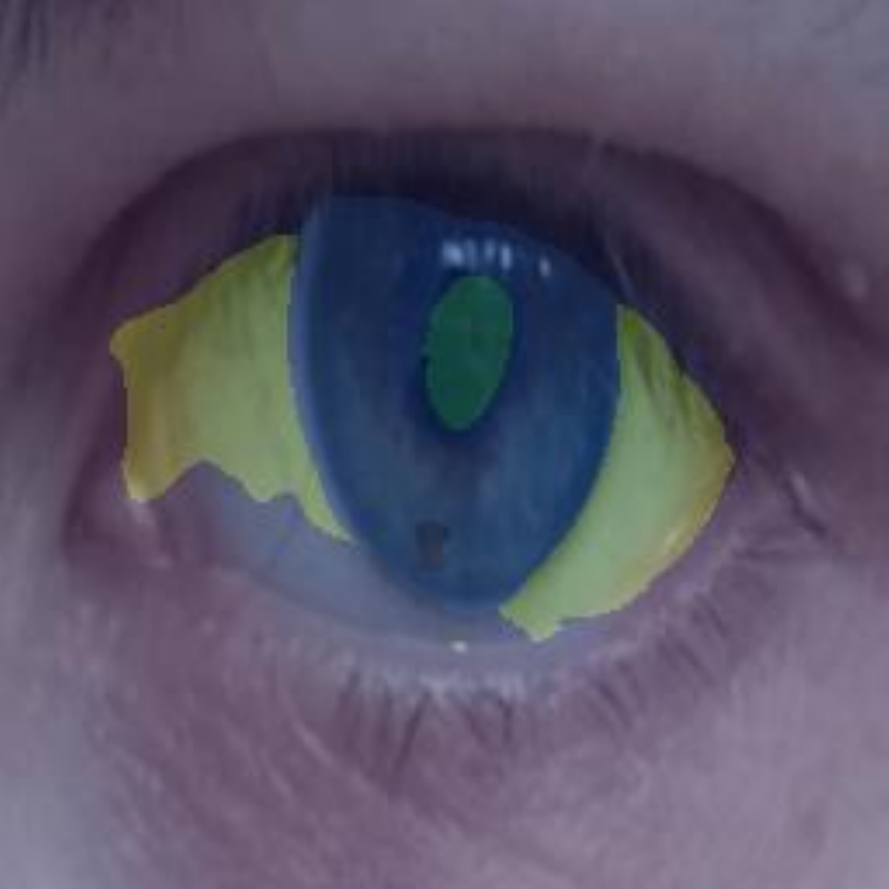}   
    \includegraphics[width=0.16\textwidth]{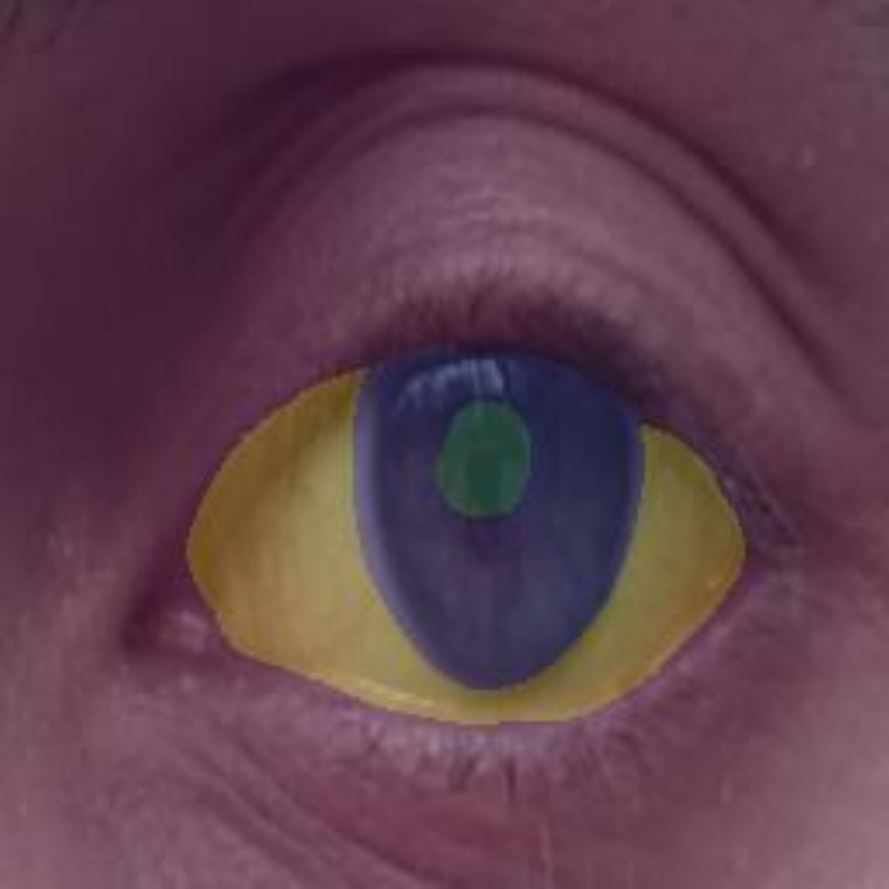} 
    &
    \includegraphics[width=0.16\textwidth]{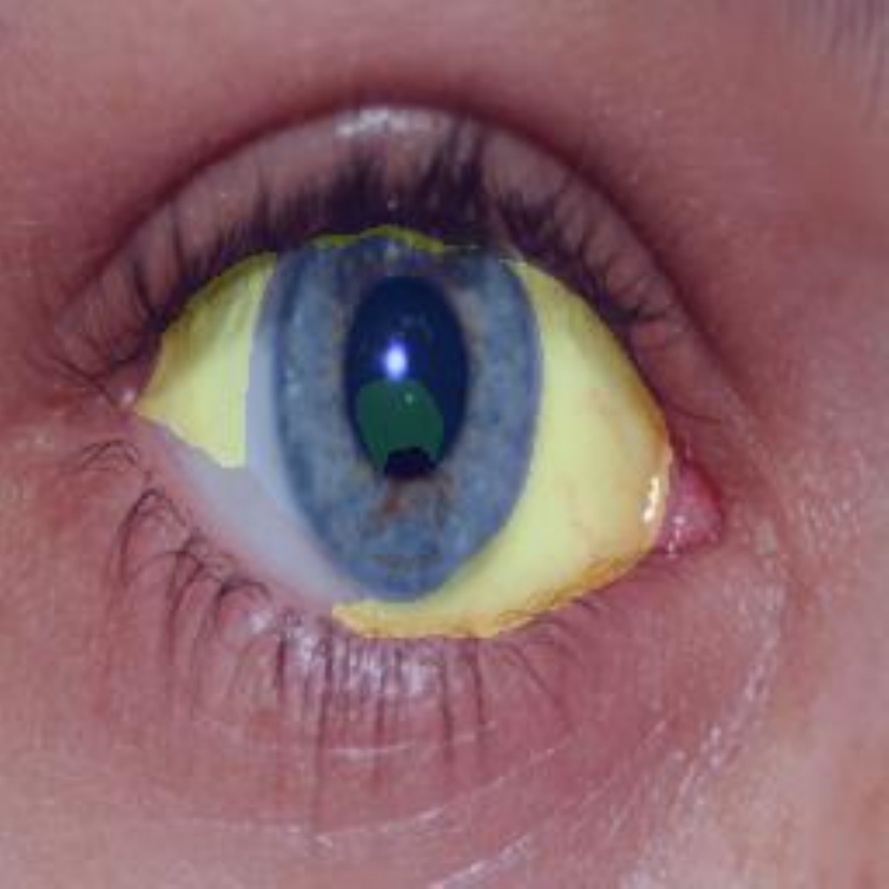}  
    \includegraphics[width=0.16\textwidth]{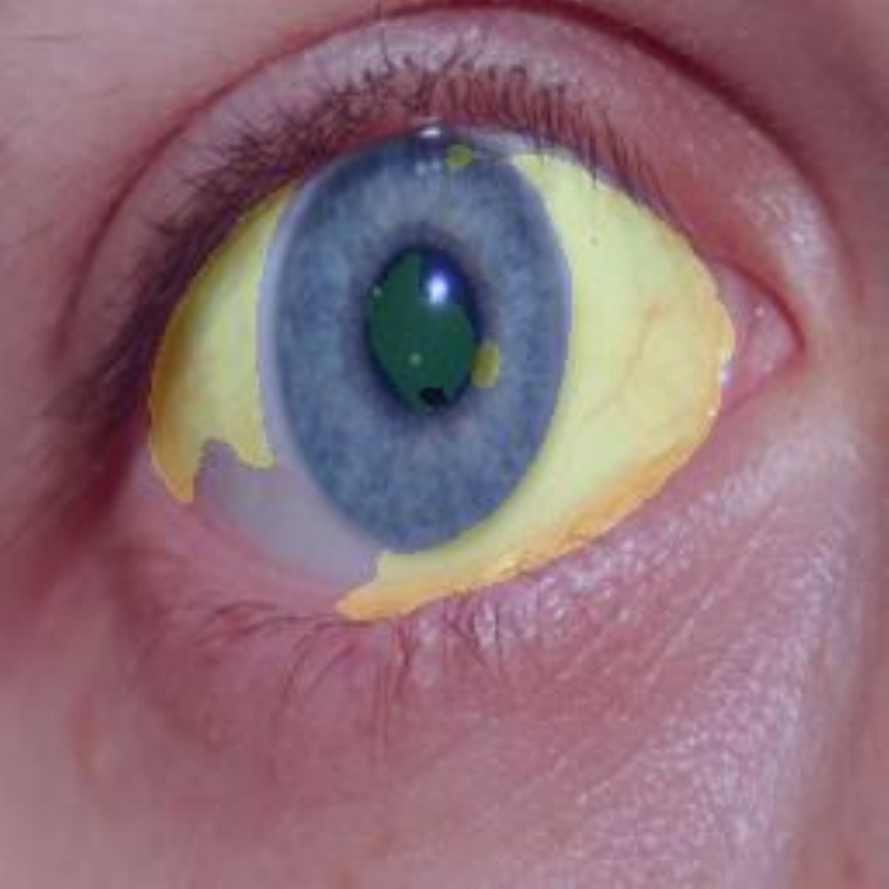} 
    \\
    \rotatebox{90}{\hspace{2mm} \large $8$ annotations} & 
    \includegraphics[width=0.16\textwidth]{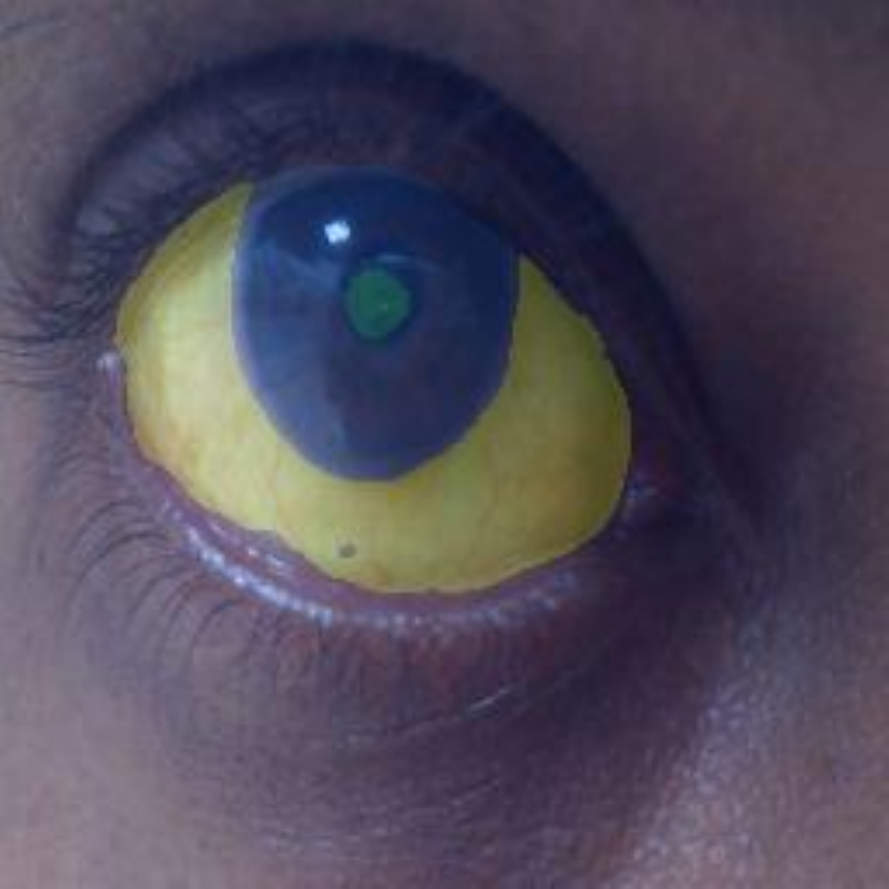} 
    \includegraphics[width=0.16\textwidth]{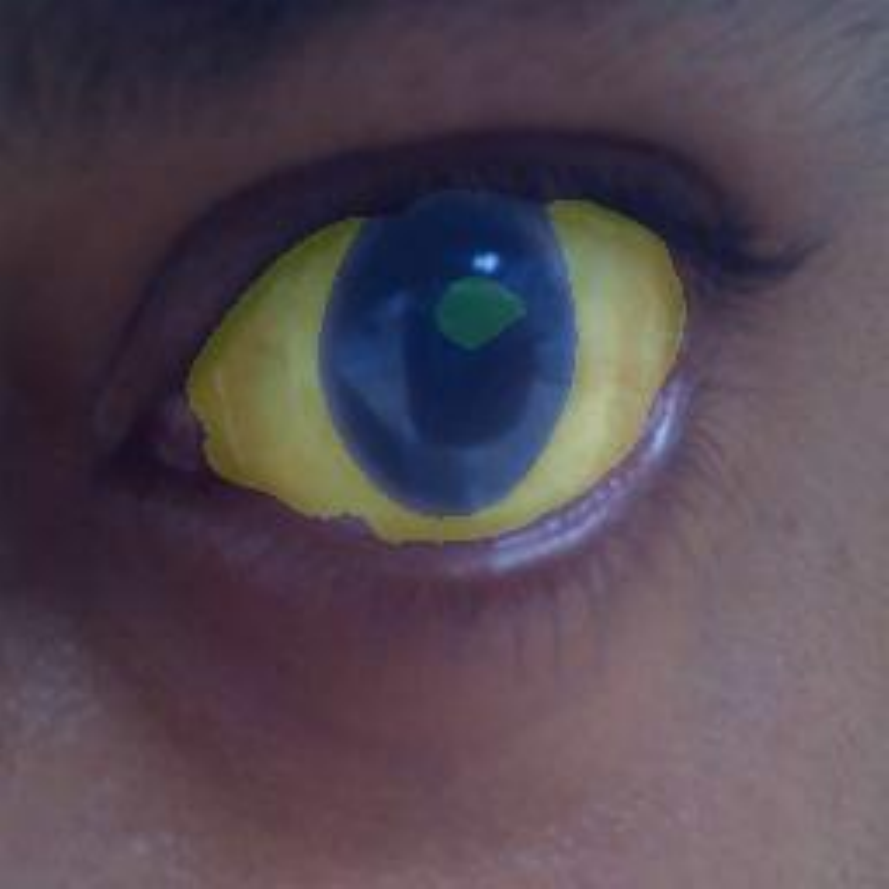} 
    &
    \includegraphics[width=0.16\textwidth]{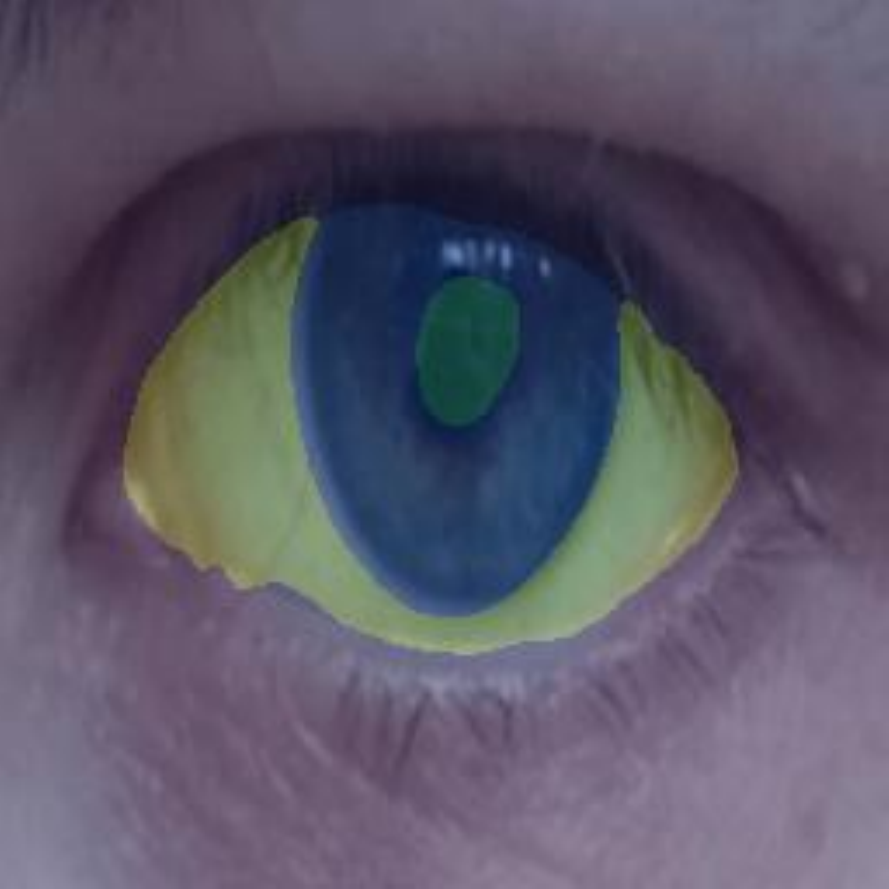}  
    \includegraphics[width=0.16\textwidth]{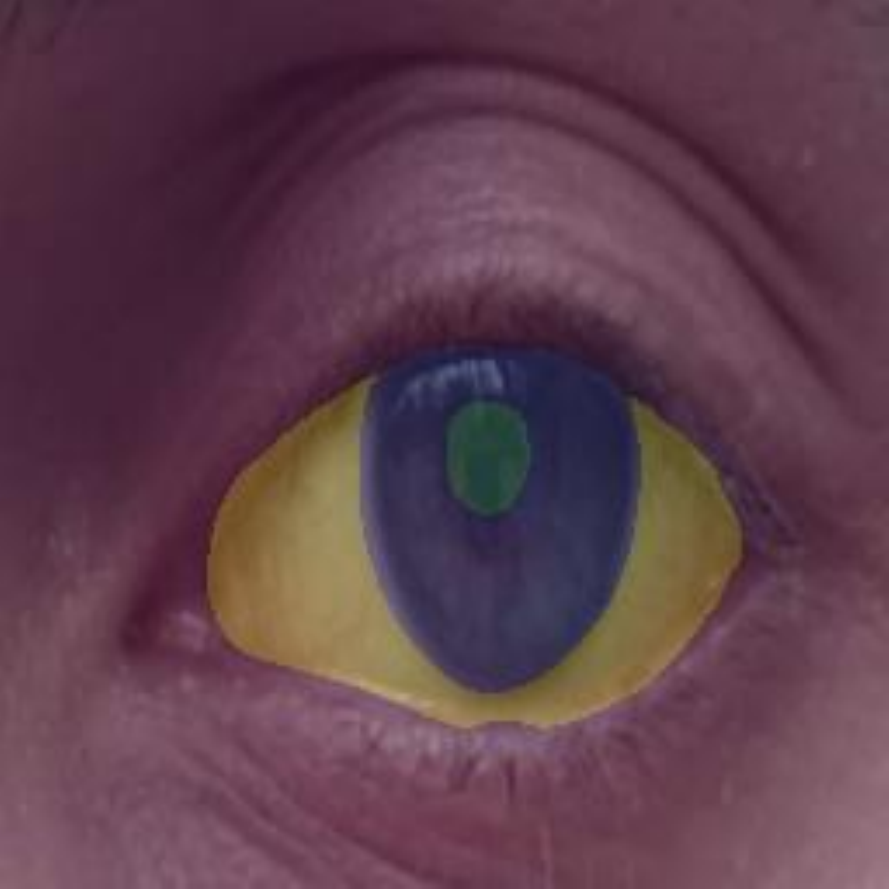} 
    &
    \includegraphics[width=0.16\textwidth]{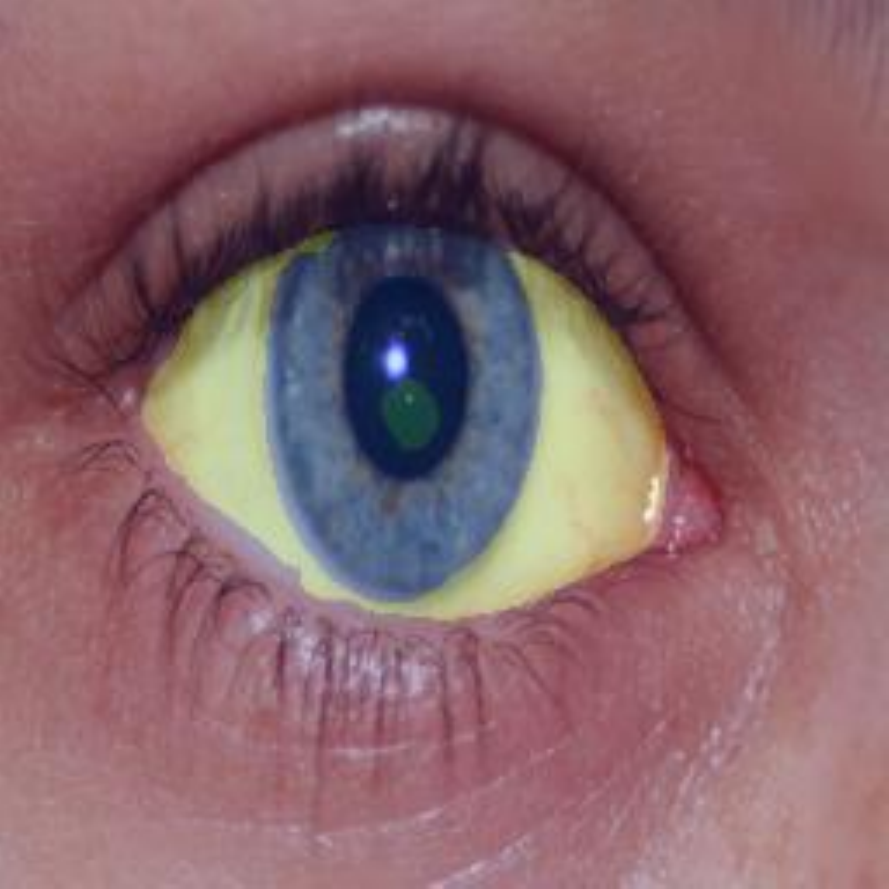}  
    \includegraphics[width=0.16\textwidth]{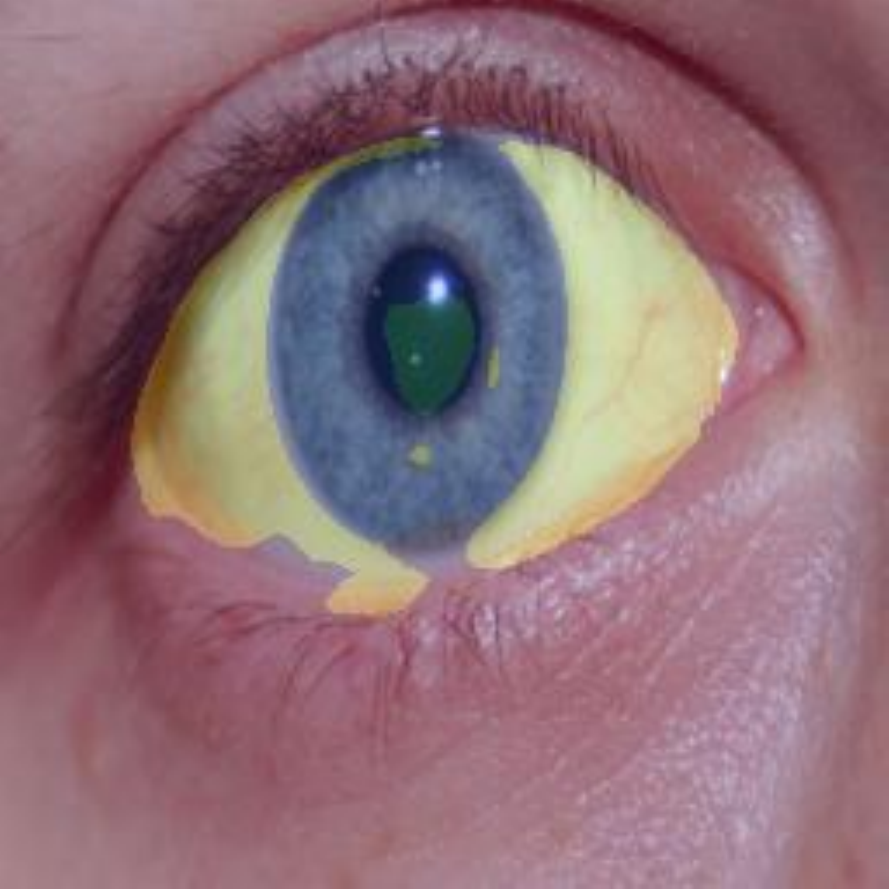} 
    \\
    %\\
    %\hline 
    %& &&\vspace{-3mm}\\
    %\rotatebox{90}{\hspace{10mm} \large Ground truth} & 
    %\includegraphics[width=0.23\textwidth]{figures/segmentation_mask_comparison/example_GT_merge_PolyU_NIR_RGB__SMD_256_64.pdf} &
    %\includegraphics[width=0.23\textwidth]{figures/segmentation_mask_comparison/example_GT_merge_PolyU_NIR_RGB__MOBIUS_256_center_3.pdf} &
    %\includegraphics[width=0.23\textwidth]{figures/segmentation_mask_comparison/example_GT_merge_PolyU_NIR_RGB__SBVPI_sclera_center_56.pdf}
    %\\ 
    
    %& \multicolumn{2}{c}{ Original images}&\multicolumn{2}{c}{ Synthesized images}\\ 
\end{tabular}}
\end{center}\vspace{-4mm}
\caption{\textbf{Example segmentation results as a function of the number of images used for training BiOcularGAN.} Note how the quality of the segmentations increases when more images are used to learn the mask-generation procedure of BiOcularGAN (see results down the columns).
}
\label{fig:annotation_number_testing}
\end{figure}

\begin{figure*}[t!] 
\begin{center}
\resizebox{\textwidth}{!}{%
\begin{tabular}{cc}
    \rotatebox{90}{\hspace{5mm}  SBVPI} &
    \includegraphics[width=0.12\textwidth]{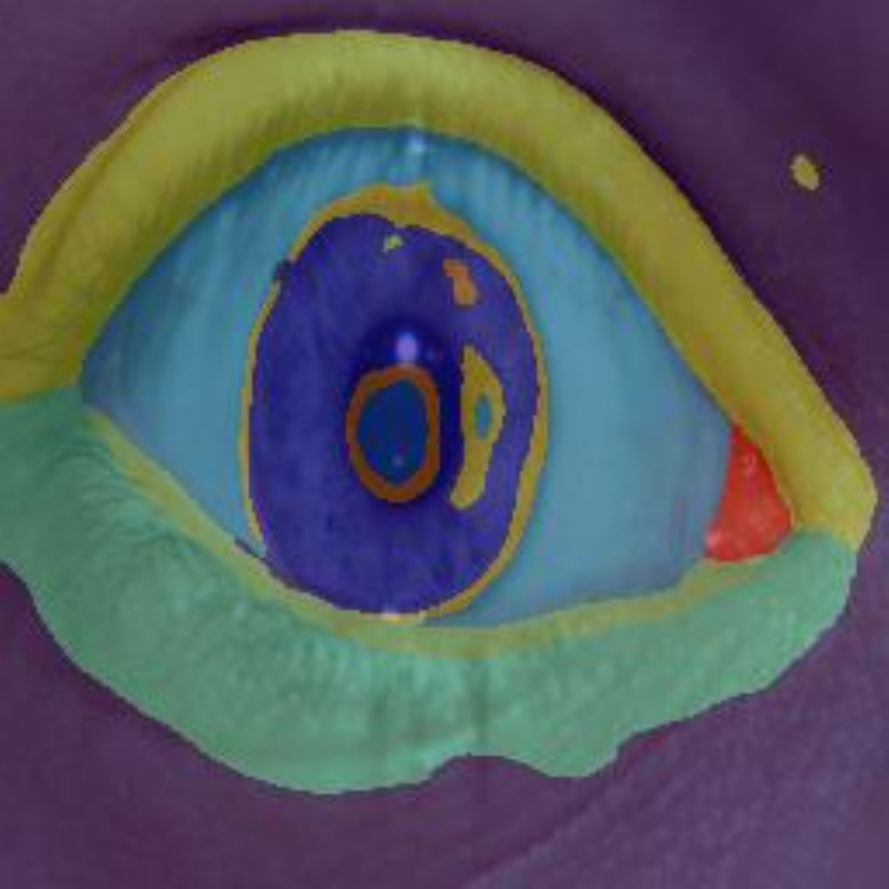} 
    \includegraphics[width=0.12\textwidth]{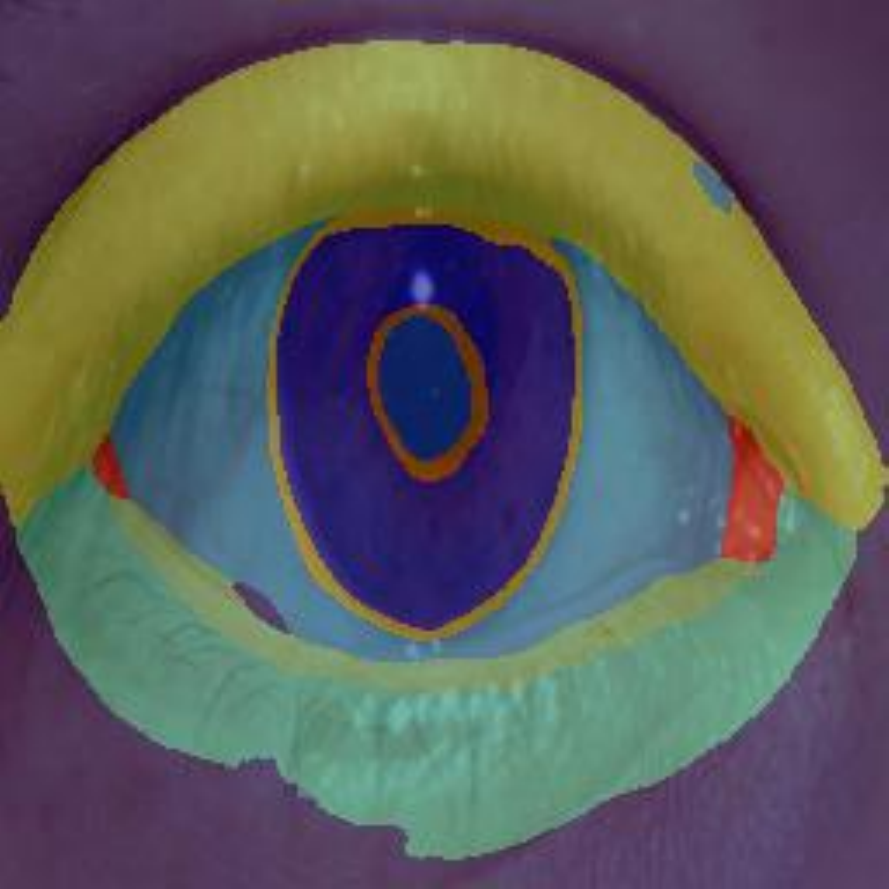} 
    \includegraphics[width=0.12\textwidth]{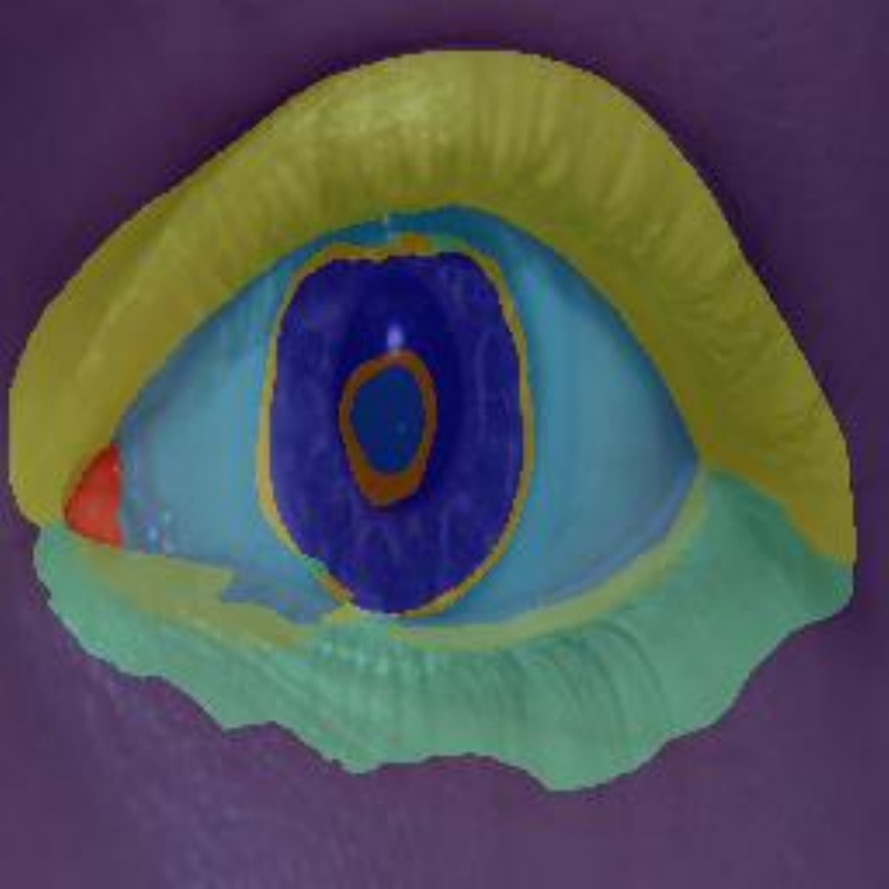} 
    \includegraphics[width=0.12\textwidth]{figures/multiclass_segmentation/test/SBVPI/example_merge_PolyU_NIR_RGB_multi_class__SBVPI_sclera_center_45.pdf} 
    \includegraphics[width=0.12\textwidth]{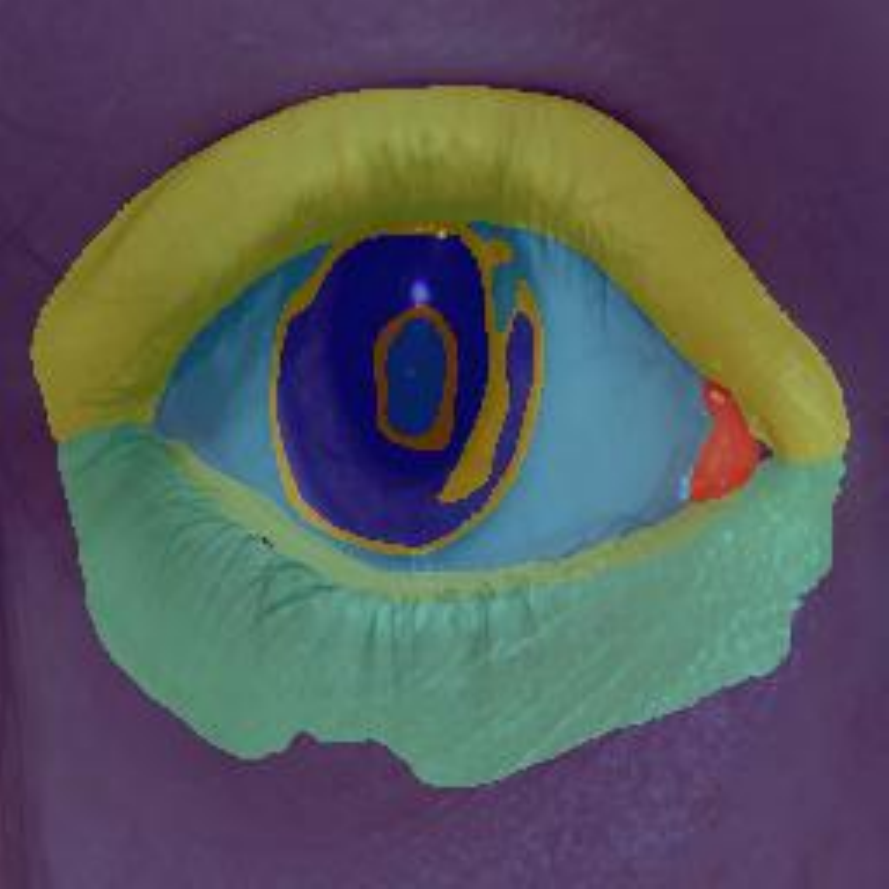} 
    \includegraphics[width=0.12\textwidth]{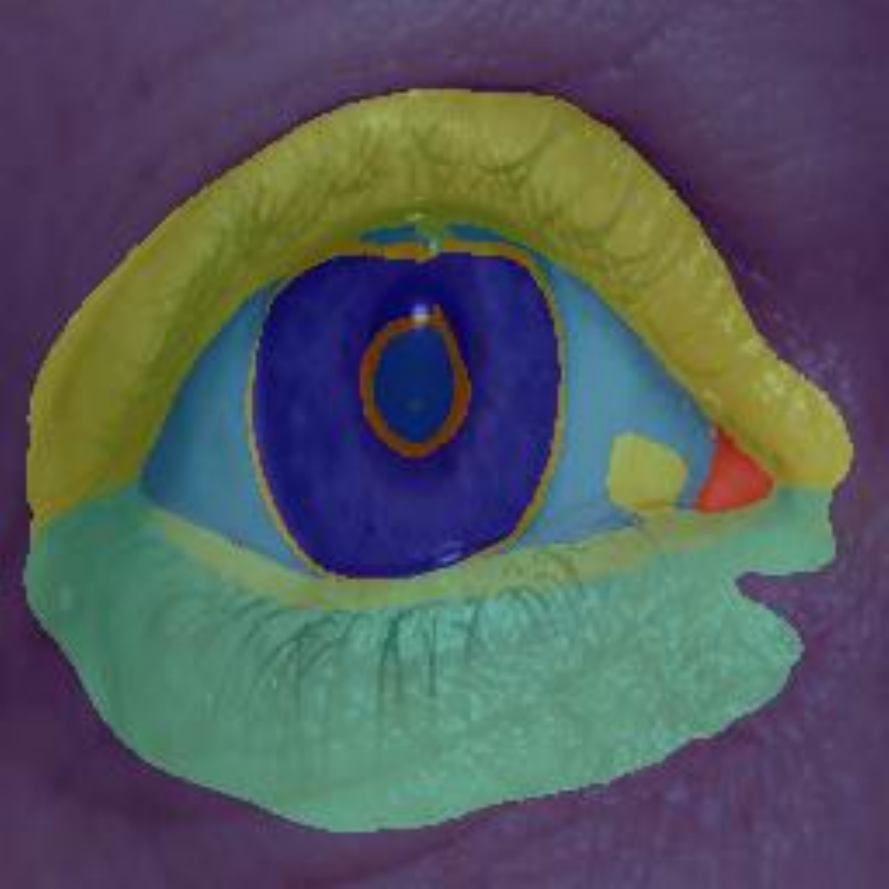} 
    \includegraphics[width=0.12\textwidth]{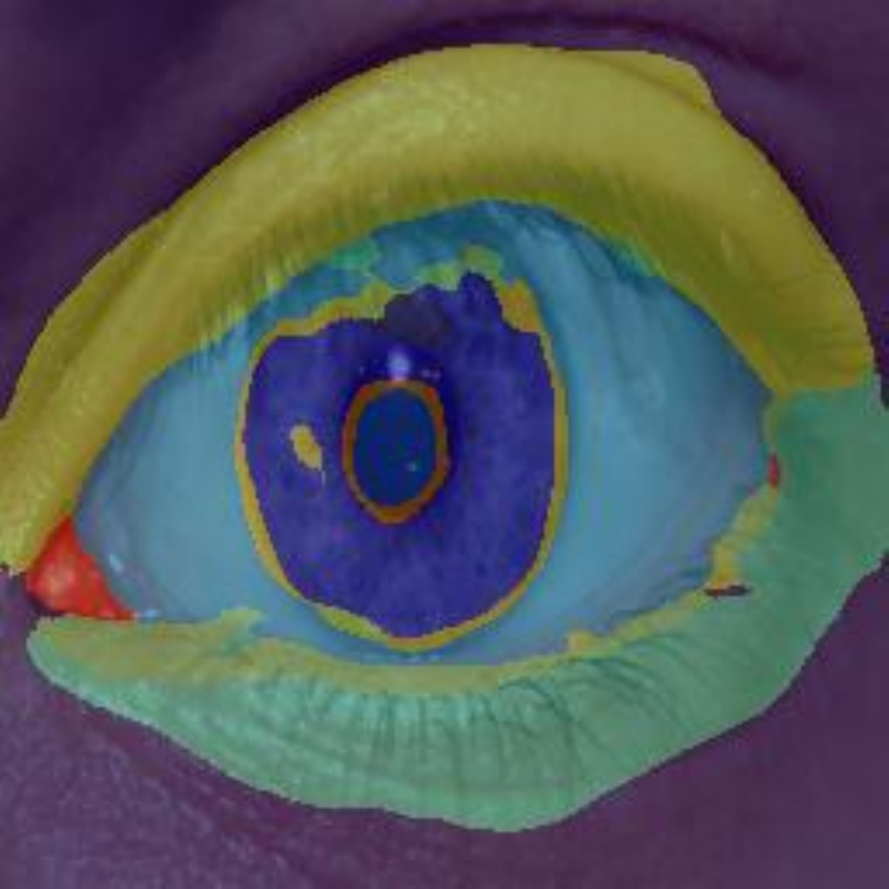}  
    \includegraphics[width=0.12\textwidth]{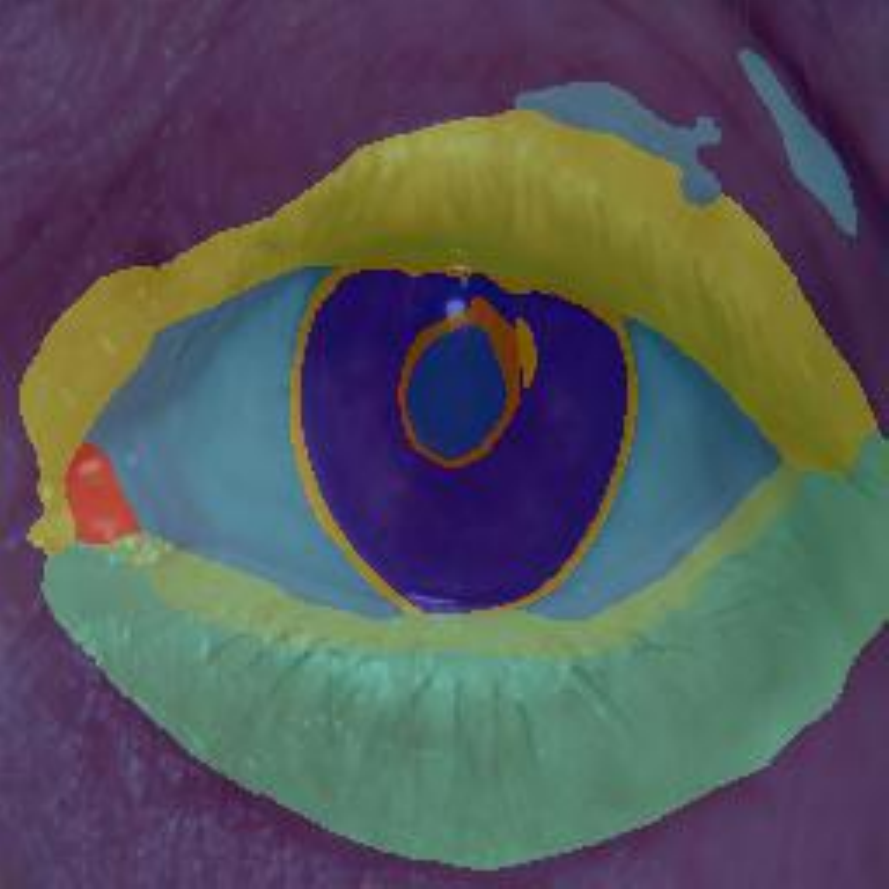}  
     \\
    \rotatebox{90}{\hspace{3mm}  MOBIUS} &
    \includegraphics[width=0.12\textwidth]{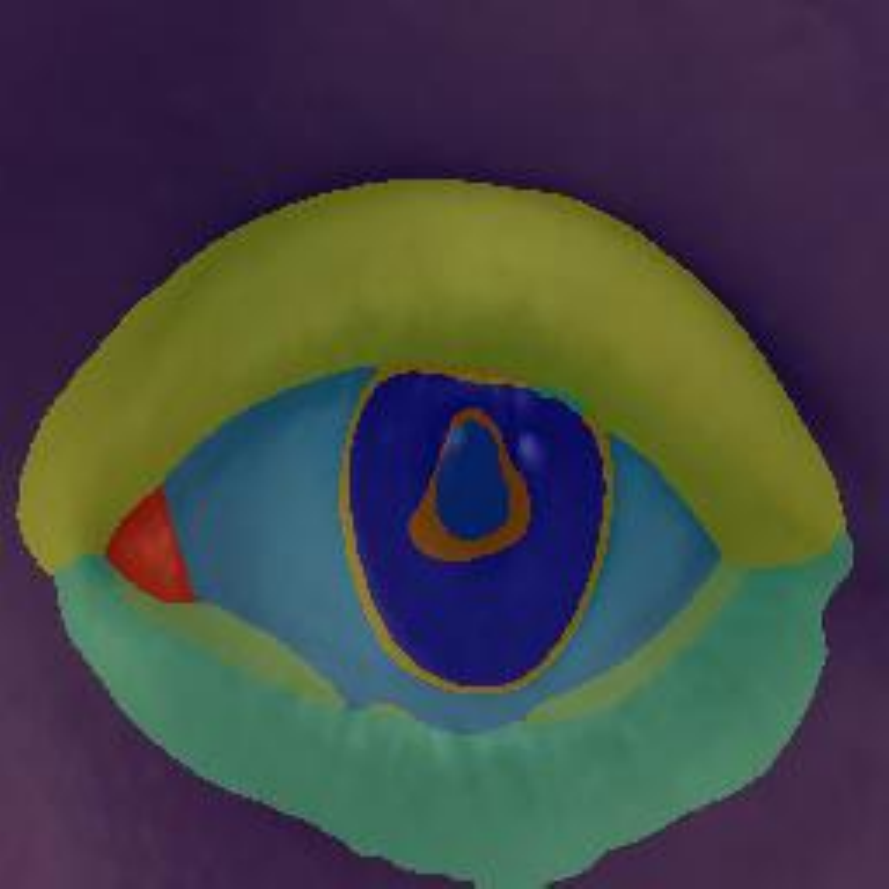}
    \includegraphics[width=0.12\textwidth]{figures/multiclass_segmentation/test/MOBIUS/example_merge_PolyU_NIR_RGB_multi_class__MOBIUS_256_center_20.pdf}
    \includegraphics[width=0.12\textwidth]{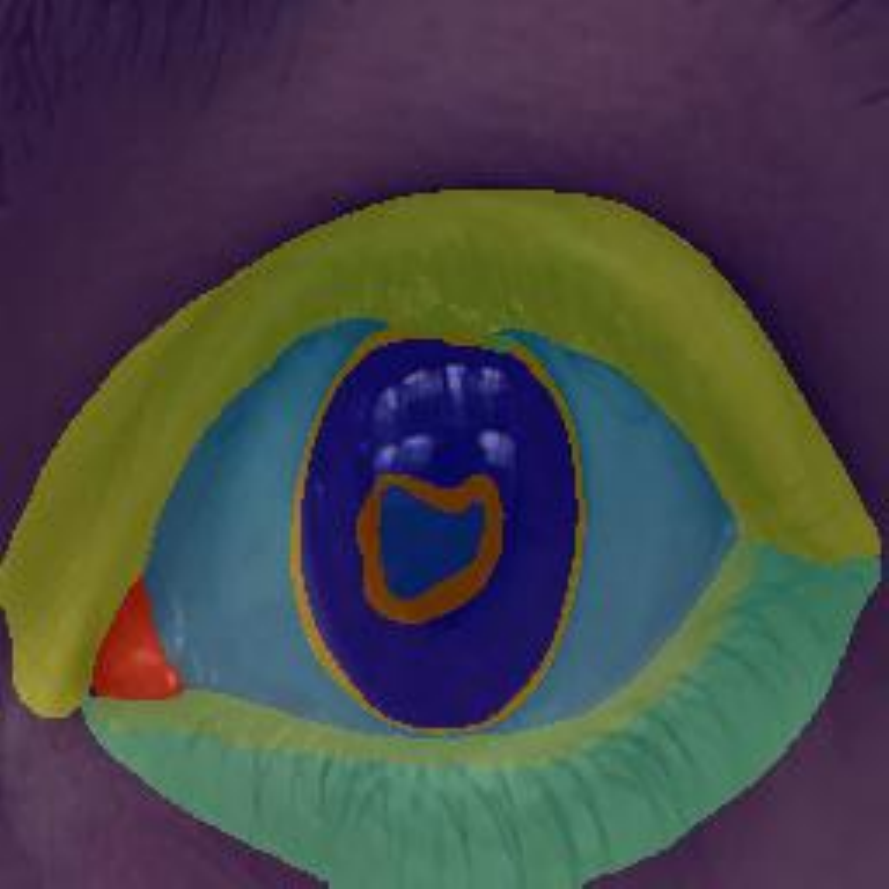}
    \includegraphics[width=0.12\textwidth]{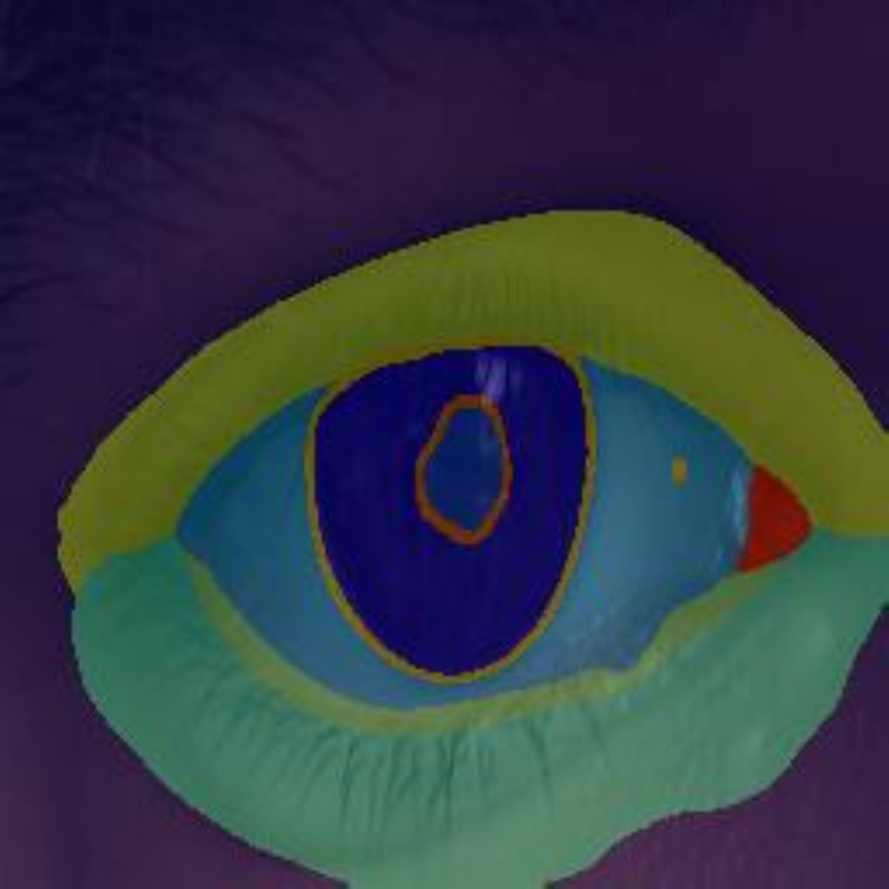} 
    \includegraphics[width=0.12\textwidth]{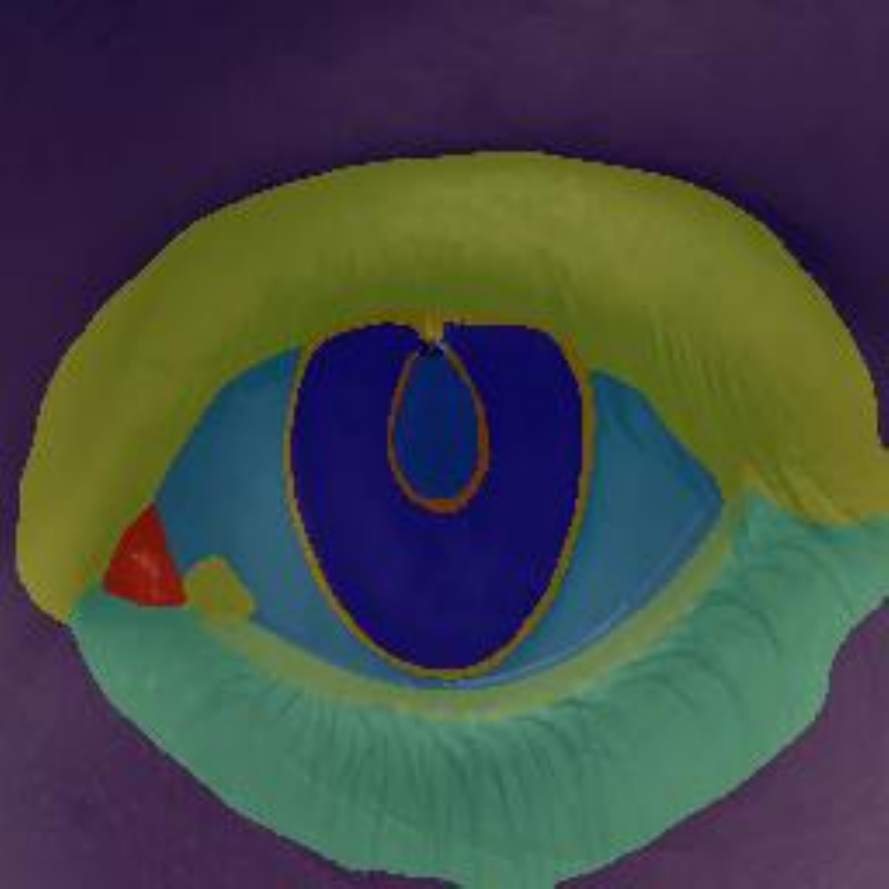}
    \includegraphics[width=0.12\textwidth]{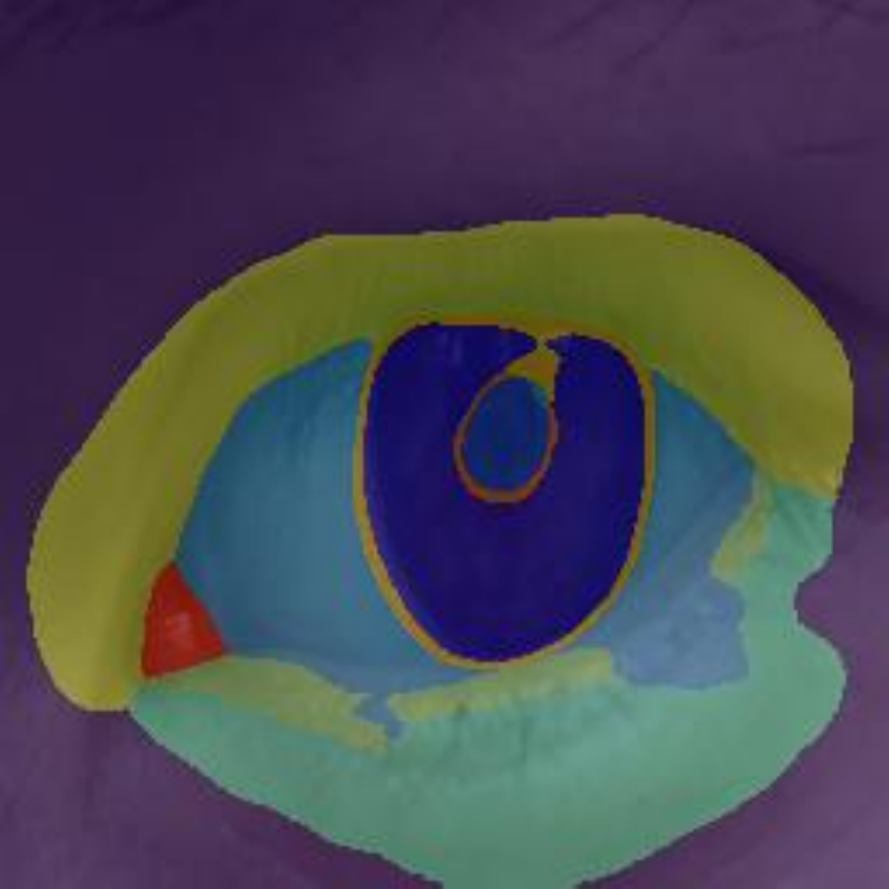}
    \includegraphics[width=0.12\textwidth]{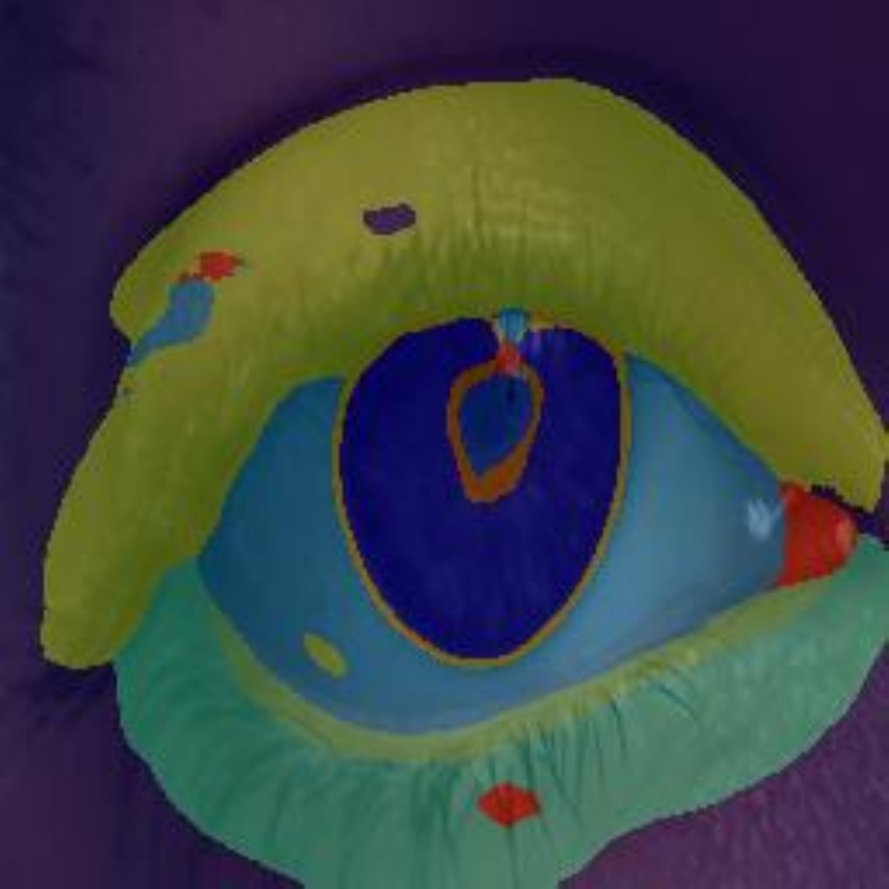}
    \includegraphics[width=0.12\textwidth]{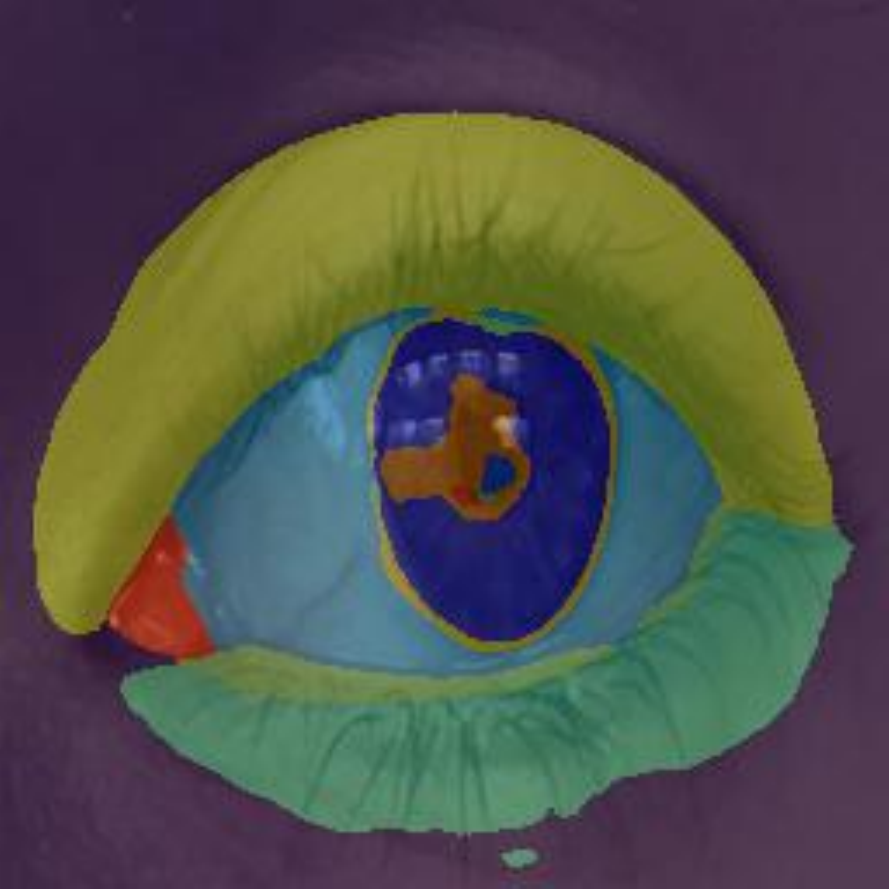}
    \\
    \rotatebox{90}{\hspace{6mm}  SMD} &
    \includegraphics[width=0.12\textwidth]{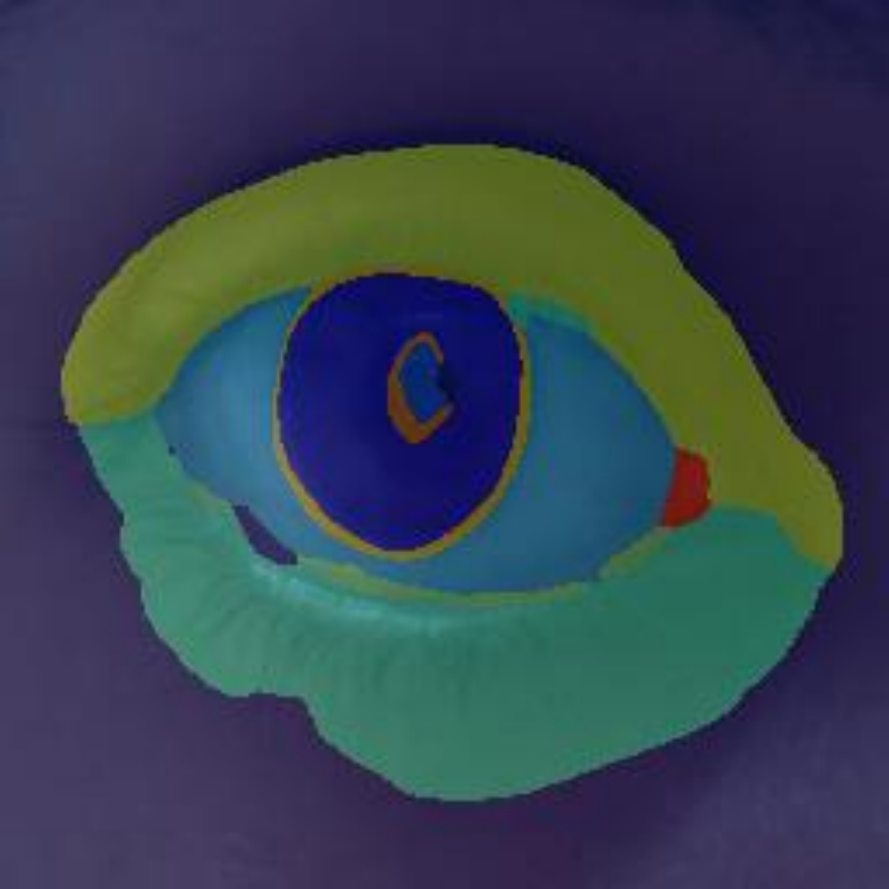}
    \includegraphics[width=0.12\textwidth]{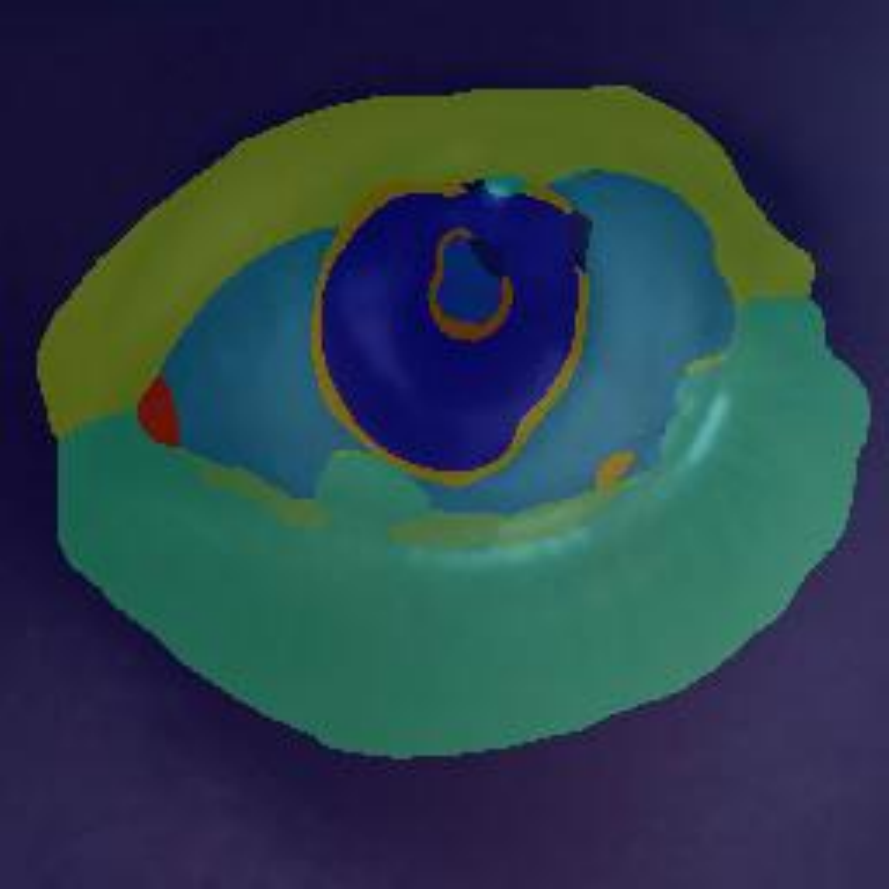}
    \includegraphics[width=0.12\textwidth]{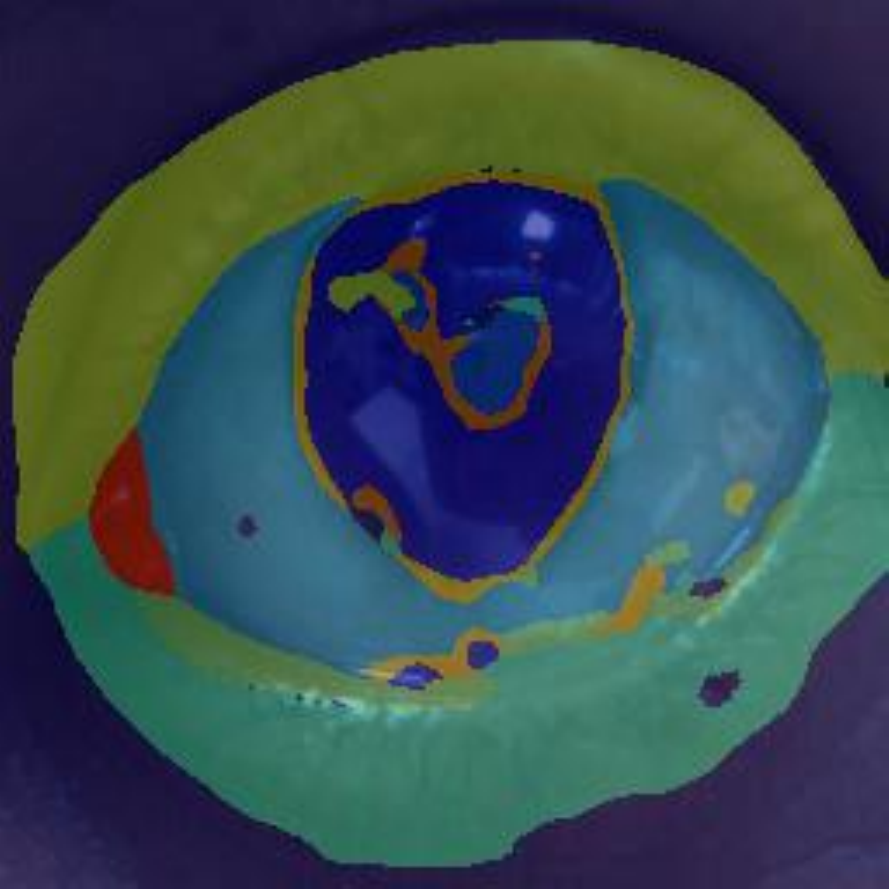}
    \includegraphics[width=0.12\textwidth]{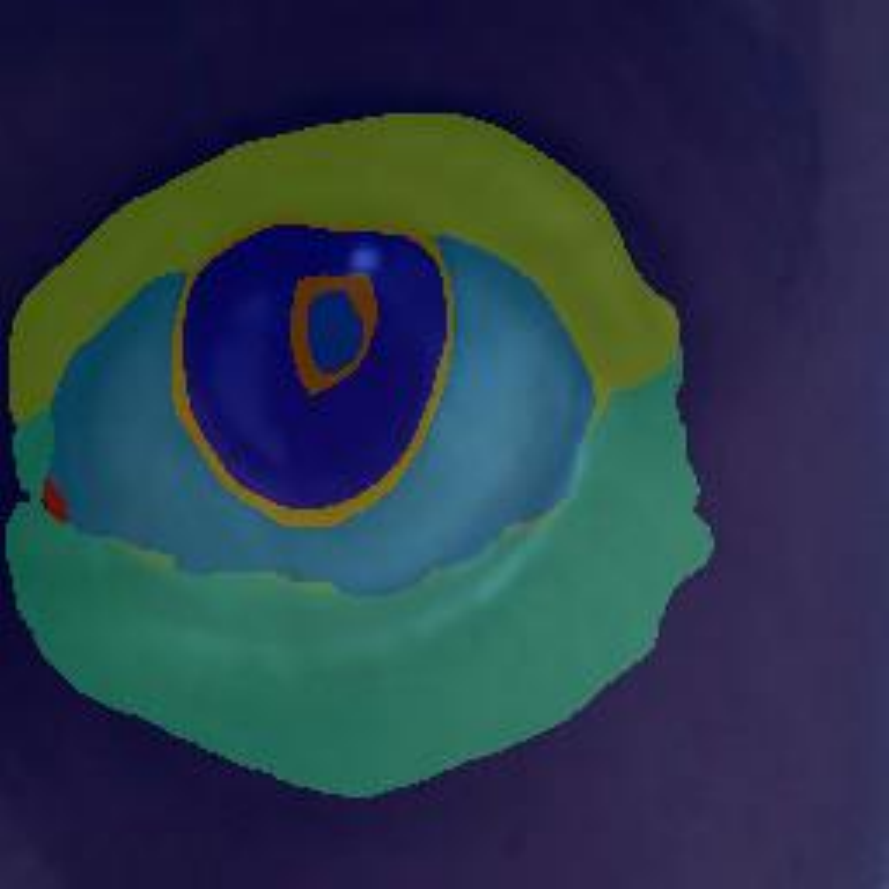}
    \includegraphics[width=0.12\textwidth]{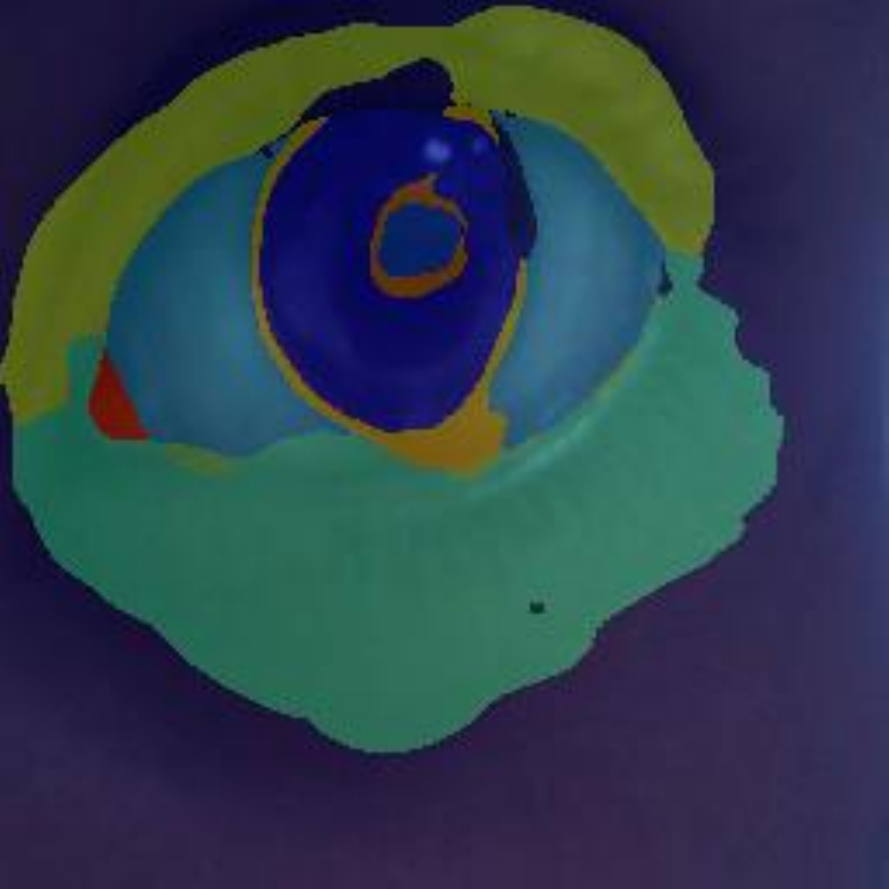}
    \includegraphics[width=0.12\textwidth]{figures/multiclass_segmentation/test/SMD/example_merge_PolyU_NIR_RGB_multi_class__SMD_256_60.pdf}
    \includegraphics[width=0.12\textwidth]{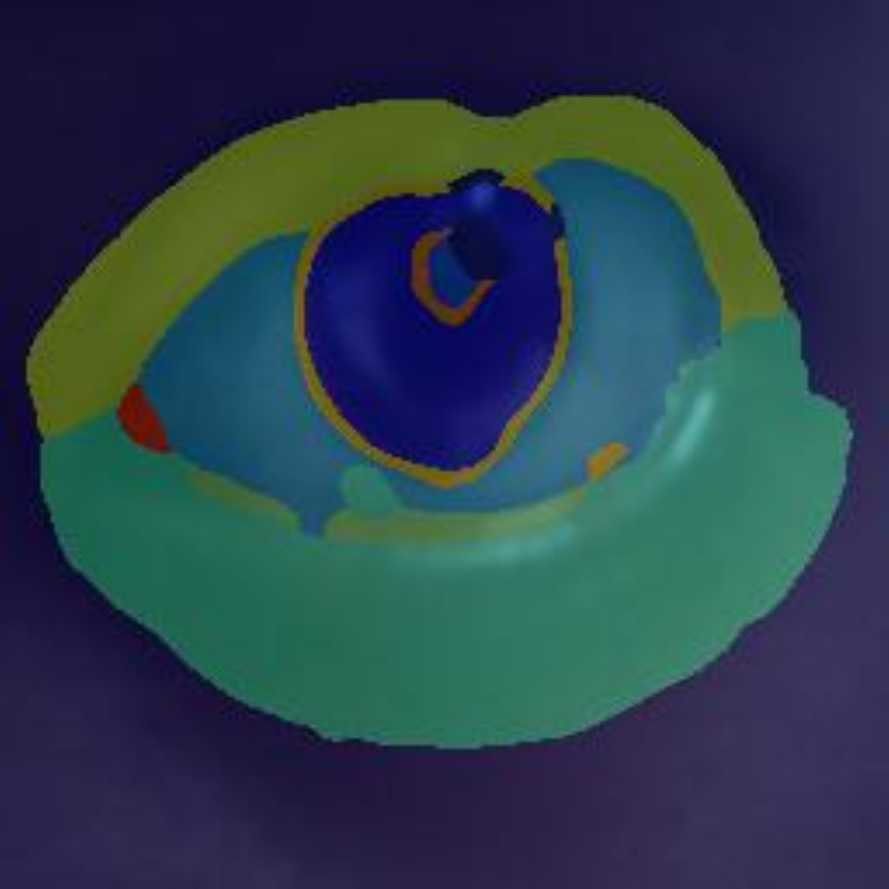}
    \includegraphics[width=0.12\textwidth]{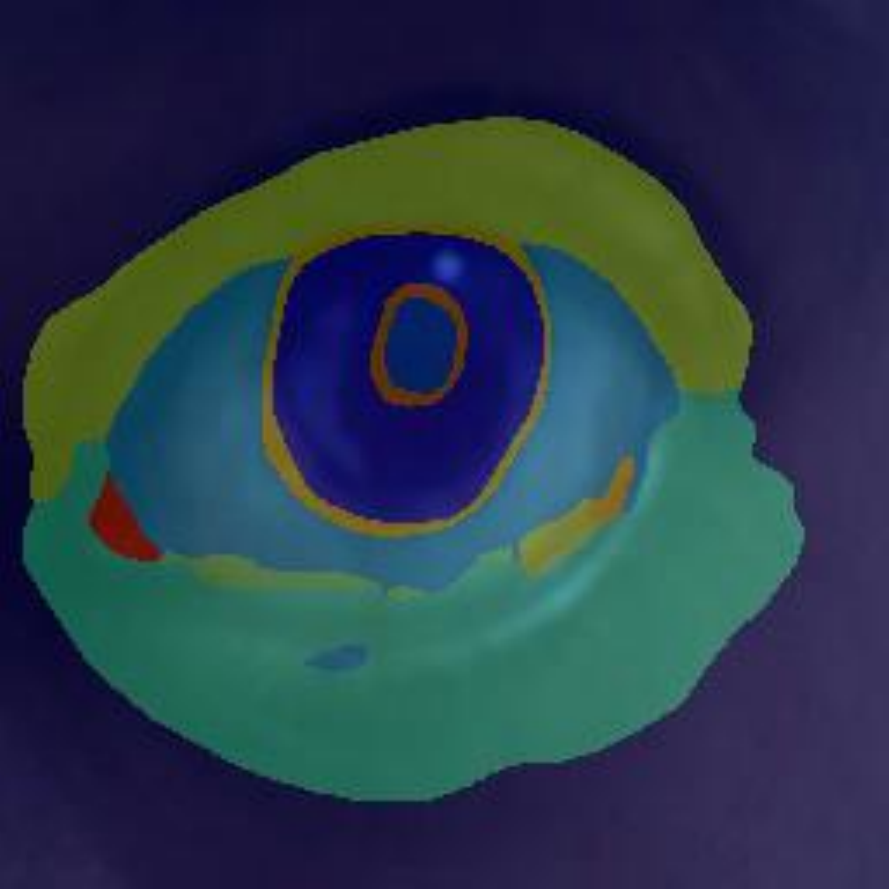} 
    %& \multicolumn{2}{c}{ Original images}&\multicolumn{2}{c}{ Synthesized images}\\ 
    
    \\
    \rotatebox{90}{\hspace{1.5mm}  CASIA-Iris} &
    \includegraphics[width=0.12\textwidth]{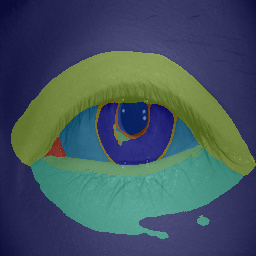}
    \includegraphics[width=0.12\textwidth]{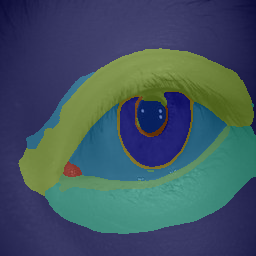}
    \includegraphics[width=0.12\textwidth]{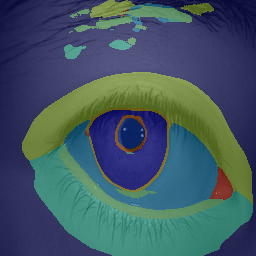}
    \includegraphics[width=0.12\textwidth]{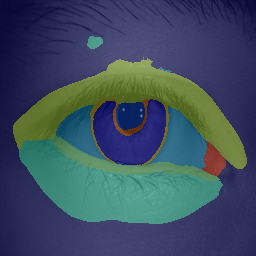}
    \includegraphics[width=0.12\textwidth]{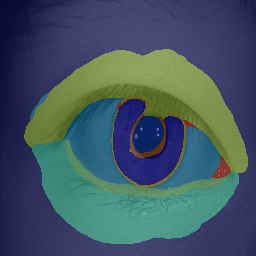}
    \includegraphics[width=0.12\textwidth]{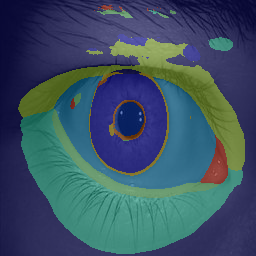}
    \includegraphics[width=0.12\textwidth]{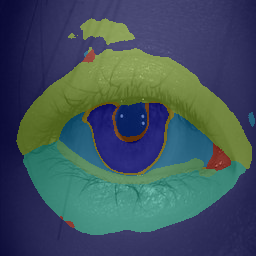}
    \includegraphics[width=0.12\textwidth]{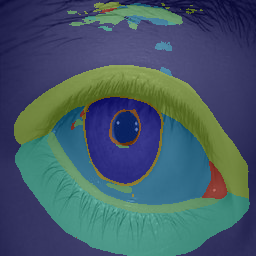} 
    
\end{tabular}}
\end{center}\vspace{-4mm}
\caption{\textbf{Example (fine-grained) segmentation results on four test datasets.} The test datasets include VIS images (SMD, MOBIUS, SBVPI), but also NIR data (CASIA-Iris V4). The presented examples were generated with a U-Net model trained with synthetic data produced by the BiOcularGAN framework. The mask-generation procedure of BiOcularGAN was learned with $2$ manually annotated images that included $10$ classes, i.e., the pupil, the pupil boundary, the iris, the iris boundary, the sclera, the upper eyelid, the lower eyelid, the inner part of the lower eyelid, the lacrimal caruncle and the background. The figure is best viewed electronically and zoomed-in for details.}
\label{tab:multiclass_annotations_additional}
\end{figure*}

\textbf{Qualitative Results.} To put the quantitative results reported in Table~\ref{tab:state_of_art_comparison_annotations} into perspective, we show in Figure \ref{fig:annotation_number_training} visual examples of the training data produced by BiOcularGAN with respect to the number of manually annotated images used.
% Note that we only present the generated segmentation mask overlaid over the VIS images (and not also the corresponding NIR images) to keep the presentation uncluttered. 
Note how the quality of the automatically generated reference annotations gets refined with a larger number of manually annotated images. This quality increase is then also reflected in the performance of the trained segmentation models, as seen by the sample segmentation results in Figure \ref{fig:annotation_number_testing}. Here, U-Net models were used again to generate the sample results, as they ensured somewhat better performance than the DeepLab-V3 competitors in the experiments presented in the main part of the paper.

\subsection{Fine-grained segmentation} 

In the main part of the paper, we showed examples of segmentation results generated based on a detailed $10$-class markup that included the pupil, the boundary of the pupil, the iris and its boundary, the sclera, the upper eyelid, the lower eyelid, the inner part of the lower eyelid, the lacrimal caruncle and the background. Because the number of examples presented was limited due to space constraints, we show in Figure \ref{tab:multiclass_annotations_additional} a broader cross-section of visual results from all three previously mentioned test datasets, i.e., SMD, MOBIUS and SBVPI. 
The results were again generated with a U-Net model trained using the synthetic data produced by BiOcularGAN, where the mask generation procedure was learned with only $2$ manually annotated images. 
Furthermore, we also include segmentation results on NIR images of the CASIA-Iris V4 dataset, which are obtained in a similar fashion, with a U-Net model trained on the NIR counterpart of the synthetic data produced by BiOcularGAN. 

As can be seen from the presented examples, we are able to learn  well-performing segmentation models, capable of locating a large amount of semantic classes in highly diverse ocular images in both the VIS and NIR domain, despite training the models on synthetic data only. Note, for example, that the ocular data produced by the DB-StyleGAN2-P model corresponds largely to subject of Asian origin. While this is well-matched by the SMD dataset, images in MOBIUS and SBVPI come exclusively from Caucasian subjects. Nonetheless, the training data produced by the BiOcularGAN framework contains sufficiently rich information to allow the trained segmentation model to also generalize reasonably well to the two Caucasian datasets. 

Additionally, our results show that the NIR data generated by BiOcularGAN can also be used, in conjunction with the generated ground truth semantic masks, to train deep models for fine-grained segmentation of NIR images. Differently from other VIS spectrum results, segmentation errors are mostly present in the periocular region, especially near the eyebrows. This is most likely caused by the similarity between eyebrows and eyelashes, and the clear difference between eyebrows and the rest of the periocular region, which is substantially more evident in the NIR data. This also explains why similar errors are not present on VIS images.  
% Similar errors are not present in other VIS spectrum results, because the eyelashes present in the PolyU VIS training data are not as distinct as in the NIR data. 

% which did not receive its own class, but are clearly different from the rest of the periocular region. 

% CITE CASIA: 
% Institute of Automation, Chinese Academy of Science: CASIA v1.0 Iris Image Database
% http://www.nlpr.ia.ac.cn/english/irds/irisdatabase.html
% Accessed 7.8.2022

%%%%%%%%%%%%%%%%%%%%%%%%%%%%%%%%%%
\subsection{Style mixing experiments}

To further evaluate the proposed BiOcularGAN framework, we explore its capabilities for generating data with desired characteristics. 
For this, we rely on the Style mixing procedure~\cite{stylegan2_karras2020analyzing}, which entails the use of two separate intermediate latent codes $w$ to determine the style of a single generated image. By switching the $w$ input at a certain point during the image synthesis process, we are able to merge the styles of two different images. 
Here, the style inputs corresponding to the lower resolution layers in the DB-StlyeGAN2 model dictate the high-level features of the image and vice versa. By analyzing the  style mixing results, we are able to assess how entangled the various features are in the intermediate space $\mathcal{W}$. This is important, as style mixing can serve as an additional mechanism for data augmentation that can ensure a higher level of diversity in the synthetic data, while also ensuring control over the data characteristics.  

% \begin{figure}[!t] 
% \begin{center}
% \vspace{-2mm}
% \resizebox{\columnwidth}{!}{%
% \begin{tabular}{cc}
%     & \hspace{7mm}  Source $X$ \\  
%     \rotatebox{90}{\hspace{28mm}   Source $Y$} 
%     & \hspace{-2mm}\includegraphics[width=\columnwidth] {figures/Style_mixing/q75_grid_test_0-4.jpg} 
% \end{tabular}}
% \end{center}\vspace{-4mm}
% \caption{\textbf{Style mixing results of DB-StyleGAN2-CE.} Latent codes of Source $X$ images determine the shape and $Y$ codes determine the texture.
% }
% \label{fig:style_mixing}\vspace{-2mm}
% \end{figure}

\begin{figure}[!t]
\centering
\resizebox{\columnwidth}{!}{%
\begin{tabular}{cc|c} %@{\hskip 2.5mm}
    & & \huge Source $X$ \\ [1.5mm]
    & & \includegraphics[]{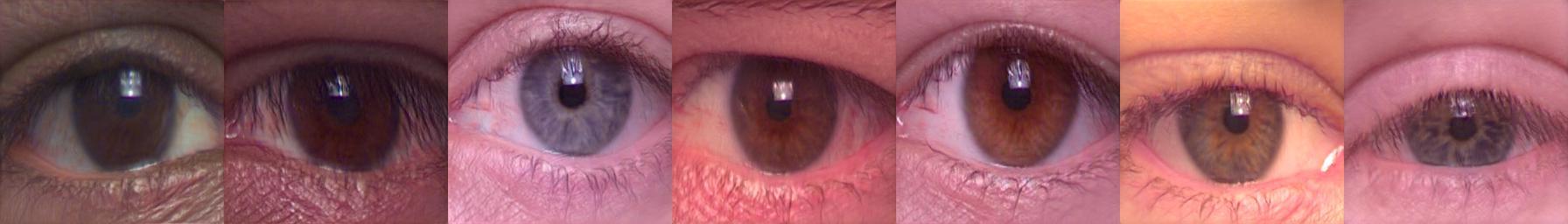} \\ [1mm]
    \hline 
    & & \vspace{-1.7mm}\\
    \rotatebox{90}{\hspace{64mm}  \huge Source $Y$} & \includegraphics[]{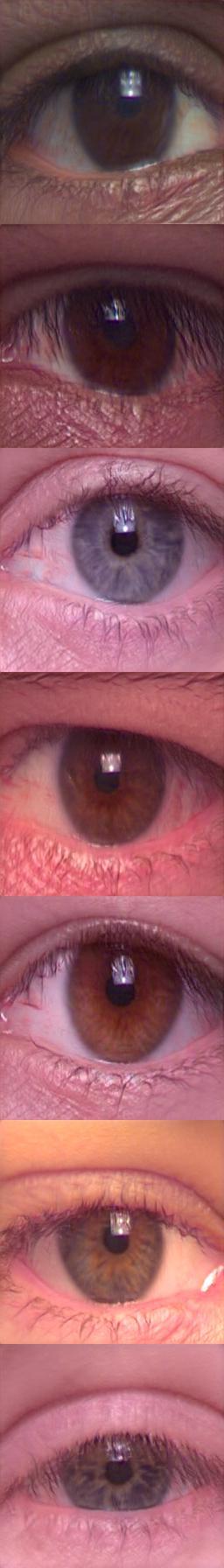} & \includegraphics[]{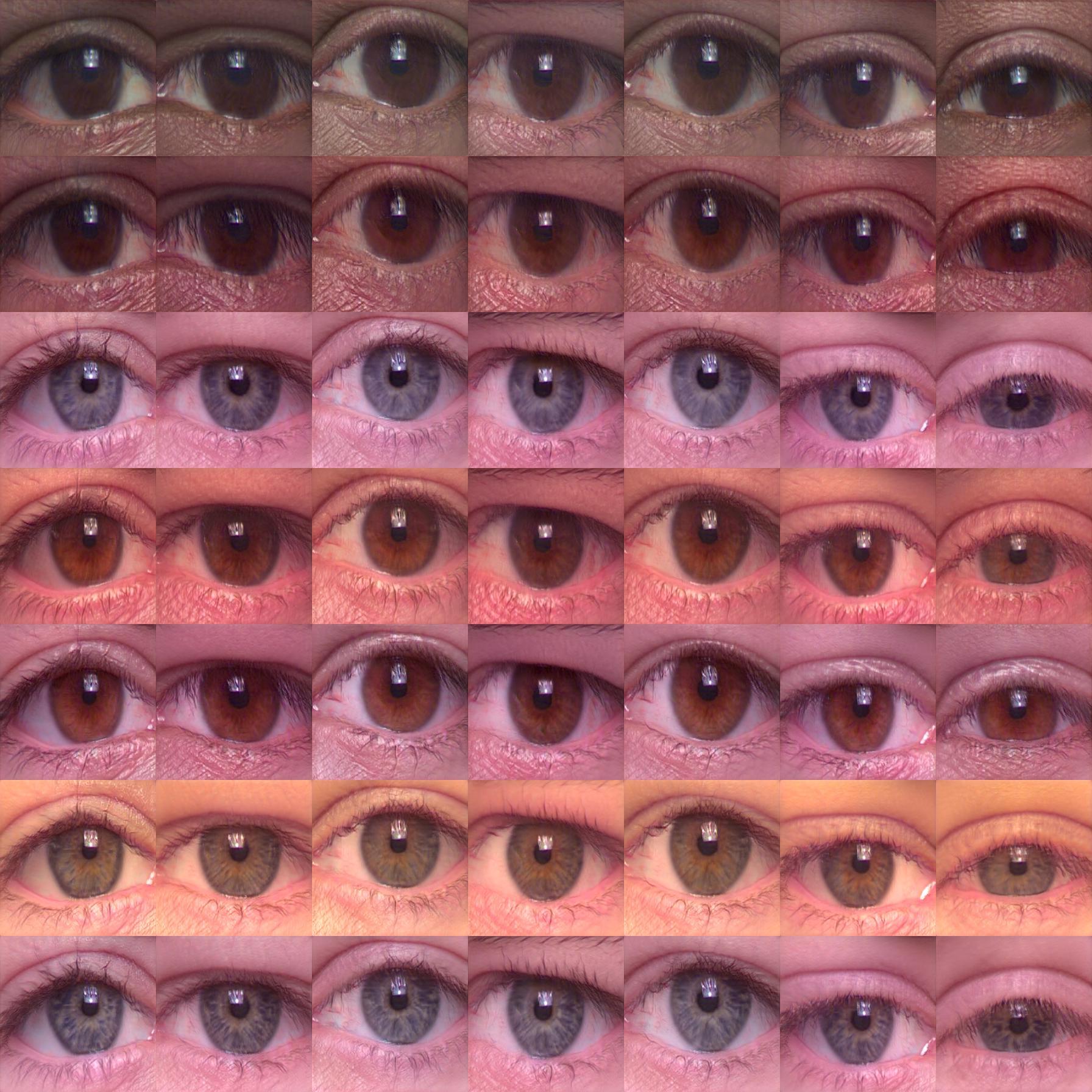}
\end{tabular}
}
\caption{{Style mixing results of DB-StyleGAN2-CE.}
Latent codes of source $X$ images determine the shape and
$Y$ codes determine the texture.
% \vspace{-mm}
}
\label{fig:style_mixing}
\end{figure}

Figure~\ref{fig:style_mixing} depicts style mixing results of $7$ different latent codes, corresponding to the images in the first column and row. Source $X$ latent codes are used as style inputs up to, and including, the $16\times16$ resolution layer, after which source $Y$ codes are used. The results show that it is possible to control the overall shapes present in the generated images with source $X$ codes. This includes features such as the position, size and shape of the various eye regions and the eyelids. Meanwhile, source $Y$ codes determine the color and texture of the iris and the skin. 

The presented results showcase that low-level and high-level features are fairly disentangled in the intermediate latent space of the DB-StyleGAN2 model. This property can be exploited to generate synthetic ocular images with a desired shape and texture, which could be utilized to address the problem of underrepresented samples in real-world or synthetic ocular datasets, and thus balance the distribution of data characteristics.

%%%%%%%%%%%%%%%%%%%%%%%%%%%%%%%%%%
\iffalse
\subsection{Higher-resolution synthesis}

\begin{figure}[b!] 
\begin{center}
\resizebox{0.9\columnwidth}{!}{%
\begin{tabular}{c}
    \includegraphics[width=0.12\textwidth]{figures/512_samples/image_127.jpg} 
    \includegraphics[width=0.12\textwidth]{figures/512_samples/image_141.jpg}  
    \includegraphics[width=0.12\textwidth]{figures/512_samples/image_156.jpg}   
    \includegraphics[width=0.12\textwidth]{figures/512_samples/image_217.jpg} 
    \includegraphics[width=0.12\textwidth]{figures/512_samples/image_344.jpg}  
    \\
    \includegraphics[width=0.12\textwidth]{figures/512_samples/image_127_NIR.jpg} 
    \includegraphics[width=0.12\textwidth]{figures/512_samples/image_141_NIR.jpg}  
    \includegraphics[width=0.12\textwidth]{figures/512_samples/image_156_NIR.jpg} 
    \includegraphics[width=0.12\textwidth]{figures/512_samples/image_217_NIR.jpg} 
    \includegraphics[width=0.12\textwidth]{figures/512_samples/image_344_NIR.jpg}  
\end{tabular}}
\end{center}\vspace{-4mm}
\caption{\textbf{TODO 512 resolution} Iskreno raje ne bi vkljucil teh rezultatov, ker so slabši od rezultatov resolucije 256 }
\label{tab:higher_resolution_512}
\end{figure}

TODO
\fi

%%%%%%%%%%%%%%%%%%%%%%%%%%%%%%%%%%
\subsection{Additional implementation details}

\textbf{Mapping network.} The mapping network of the DB-StyleGAN2 follows the design from \cite{stylegan_1_karras2019style} and consists of $8$ fully-connected layers. As input it takes a $512$-dimensional randomly sampled latent vector $\mathbf{z}\in\mathcal{Z}$, and converts it into a $512$-dimensional intermediate latent vector $\mathbf{w}\in\mathcal{W}$. The network shares training parameters with the generator.

\textbf{Segmentation models.} All segmentation experiments in the main part of the paper as well as the appendix were conducted with the DeepLab-V3 \cite{florian2017rethinking} and U-Net \cite{ronneberger2015u} segmentation models. For all experiments, the two models were trained using the Adam optimizer \cite{kingma2014adam} with a learning rate of $10^{-4}$ and a batch size of $8$. To guide learning, the cross-entropy loss function was used. During training, the learning rate was decreased by a factor of $10$, if the validation loss did not improve for $5$ consecutive epochs. The training procedure was stopped once the validation loss did not improve for $10$ epochs in a row.

\subsection{Discussion}

The proposed BiOcularGAN framework allows for high-quality bimodal ocular image generation, as we have demonstrated throughout the paper. While we mostly focused on images of $256\times256$ pixels in size, which was sufficient for the segmentation experiments presented, the progressive structure of the DB-StyleGAN2 model also allows for the generation of larger images, e.g., $512\times 512$.  It is also important to note that the overall characteristics (e.g., resolution, appearance, diversity, or quality -- as defined, for instance, by ISO/IEC 29794-6) of the generated data are inherited from the characteristics of the training data, in our case from the PolyU and the CrossEyed datasets. %. In our experiments, we showed results for two DB-StyleGAN2 models -- trained on PolyU and CrossedEye, where the models were shown to be capable of generating synthetic images with matching characteristics to the training data.   
%or its binary equivalent when only two classes were present 

%{\color{blue}{ (Table \ref{tab:state_of_art_comparison_annotations}) na eni bazi, z enim segmentavcijskim modelom naredimo test, če bi vse skup učili z 2 , 4 in 8 slikami. Lahko graf damo noter ne tabelo.}}

%In Table \ref{tab:state_of_art_comparison_annotations} ... rahlo čudni rezultati .. bom poskusil še z drugačnim izborom anotacijskih slik ali drugim datasetom

%%%%%%%%%%%%%%%%%%%%%%%%%%%%%%
%\begin{table*}[!bht!] %  use * at the table* for double column

%%%%%%%%%%%%%%%%%%%%%%%%%%%%%%%%%%%%%%%%%%%%%%%
%
%   Synthetic Dataset Size - to bi blo sicer ok za pokazati, ni pa sploh kritično in pustiva, če rezultati ne podpirajo naše rdeče niti.
%
%%%%%%%%%%%%%%%%%%%%%%%%%%%%%%%%%%%%%%%%%%%%%%%

%\textbf{Synthetic Dataset Size.} 
%{\color{blue}{TODO (Table \ref{tab:state_of_art_comparison_10k}) prikaz uspešnosti segmentacijskih model treniranih na 5k in 10k slikah}} Does it overfit? Does more data help? Does it generalize to other datasets better?

\fi

\end{document}